\documentclass{article}
\usepackage[preprint,nonatbib]{neurips_2020}

\usepackage[utf8]{inputenc} % allow utf-8 input
\usepackage[T1]{fontenc}    % use 8-bit T1 fonts
\usepackage{hyperref}       % hyperlinks
\usepackage{url}            % simple URL typesetting
\usepackage{booktabs}       % professional-quality tables
\usepackage{amsfonts}       % blackboard math symbols
\usepackage{nicefrac}       % compact symbols for 1/2, etc.
\usepackage{microtype}      % microtypography

\usepackage{amsthm}
\usepackage{amssymb}
\usepackage{amsmath}
\usepackage{graphicx}
\usepackage{subfigure}
\usepackage{hyperref}
\usepackage{color}
\usepackage{array}

\usepackage[normalem]{ulem}
\useunder{\uline}{\ul}{}

\theoremstyle{definition}
\newtheorem{theorem}{Theorem}[section]
\newtheorem{assumption}[theorem]{Assumption}

\newtheorem{lemma}[theorem]{Lemma}

\theoremstyle{remark}

\newtheorem{remark}{Remark}[section]

\def\E{\mathbb{E}}

\DeclareRobustCommand{\stirling}{\genfrac\{\}{0pt}{}}

\makeatletter
\def\maketag@@@#1{\hbox{\m@th\normalfont\normalsize#1}}
\makeatother

\newcolumntype{P}[1]{>{\centering\arraybackslash}p{#1}}

\title{Support Estimation \\with Sampling Artifacts and Errors}

\author{%
  Eli Chien \\
  ECE\\
  UIUC\\
  \texttt{ichien3@illinois.edu} \\
  \And
   Olgica Milenkovic \\
   ECE \\
   UIUC \\
   \texttt{milenkov@illinois.edu} \\
   \And
   Angelia Nedich \\
   School of Electrical, Computer and Energy Engineering \\
   Arizona State University \\
   \texttt{Angelia.Nedich@asu.edu} \\
}

\begin{document}

\maketitle

\begin{abstract}
The problem of estimating the support of a distribution is of great importance in many areas of machine learning, computer science, physics and biology.
Most of the existing work in this domain has focused on settings that assume perfectly accurate sampling approaches, which is seldom true in practical data science. Here we introduce the first known approach to support estimation in the presence of sampling artifacts and errors where each sample is assumed to arise from a Poisson repeat channel which simultaneously captures repetitions and deletions of samples. The proposed estimator is based on regularized weighted Chebyshev approximations, with weights governed by evaluations of so-called Touchard (Bell) polynomials. The supports in the presence of sampling artifacts are calculated using discretized semi-infite programming methods. The estimation approach is tested on synthetic and textual data, as well as on GISAID data collected to address a new problem in computational biology: mutational support estimation in genes of the SARS-Cov-2 virus. In the later setting, the Poisson channel captures the fact that many individuals are tested multiple times for the presence of viral RNA, thereby leading to repeated samples, while other individual's results are not recorded due to test errors. For all experiments performed, we observed significant improvements of our integrated methods compared to those obtained through adequate modifications of state-of-the-art noiseless support estimation methods. Our code will be released upon acceptance.
% We introduce a new method for estimating the support size of an unknown distribution which provably matches the performance bounds of the state-of-the-art techniques in the area and outperforms them in practice. In particular, we present both theoretical and computer simulation results that illustrate the utility and performance improvements of our method. The theoretical analysis relies on introducing a new weighted Chebyshev polynomial approximation method, jointly optimizing the bias and variance components of the risk, and combining the weighted minmax polynomial approximation method with discretized semi-infinite programming solvers. Such a setting allows for casting the estimation problem as a linear program (LP) with a small number of variables and constraints that may be solved as efficiently as the original Chebyshev approximation problem. Our technique is tested on synthetic data, textual data (Shakespeare's plays) and used to address an important problem in computational biology - estimating the number of bacterial genera in the human gut. On synthetic datasets, for practically relevant sample sizes, we observe significant improvements in the value of the worst-case risk compared to existing methods. The same is true of the text data. For the bioinformatics application, using metagenomic data from the NIH Human Gut and the American Gut Microbiome Projects, we generate a list of frequencies of bacterial taxa that allows us to estimate the number of bacterial genera to $\sim2300$.
\end{abstract}

% \fi
% \IEEEpeerreviewmaketitle

\section{Introduction}
Estimating the support size of a discrete distribution is an important theoretical and data processing problem~\cite{fisher1943relation,efron1976estimating}. In computer science, this task frequently arises in large-scale database mining and network monitoring where the objective is to estimate the types of database entries or IP addresses from a limited number of observations~\cite{raskhodnikova2009strong,bar2002counting,charikar2000towards}. In machine learning, support estimation is used to bound the number of clusters in clustering problems encountered in semi-supervised or active learning~\cite{chien2018query,ashtiani2016clustering,chien2018hs,chien2019active}. In life sciences, support estimation arises when estimating population sizes or increases in population sizes~\cite{good1956number}.
%In computational genomics and life sciences, estimating the support of a distribution is equivalent to estimating the number of species or cell types in a biosystem~\cite{nelson2012abundance,tennessen2012evolution,keinan2012recent}.
%The difficulty of this problem depends on how relatively ``simple'' is the underlying distribution in comparison to the sample size. For example, if we have $1000$ independent samples drawn from a roughly uniform distribution supported on $100$ domain elements, then counting the number of observed distinct elements will likely be an accurate support estimation. On the other hand, if the underlying distribution has support size $1000$ or even slightly larger, then the naive counting approach will give us a poor estimation. In this case, can one still estimate the support accurately?\\
%Many real world tasks face this challenge. For example, a major task for ecologists is to estimate the number of species in a population of animals from limited observations \cite{fisher1943relation}. Linguists are interested in estimating the vocabulary size of an author based on his/her complete works \cite{efron1976estimating}. For biologists, they are interested in the study of genetic mutations across a population where there are many ``rare events'' \cite{nelson2012abundance,tennessen2012evolution,keinan2012recent}.
The most challenging practical support estimation issues are encountered in the ``small sample set'' regime in which one has only a limited number of observations for a distribution with a large support. In such a setting, classical maximum likelihood frequency techniques are known to perform poorly~\cite{orlitsky2003always}. It is for this sampling regime that the estimation problem has received significant attention from both the theoretical computer science and machine learning community, as well as researchers from various computational data processing areas~\cite{acharya2011competitive,paninski2003estimation,bunge1993estimating,bar2001sampling,charikar2000towards,batu2000testing,nelson2012abundance,keinan2012recent}.

By now, a number of efficient and near-optimal support estimation techniques has been reported in the literature~\cite{valiant2013estimating,wu2019chebyshev,acharya2017unified,wu2018sample,pavlichin2017approximate,han2018local,yi2018data}. All these methods traditionally use the assumption that the samples are observed without errors. In practice, sampling artifacts and noise are ubiquitous, especially when dealing with data acquired from biological and medical science experiments. As a motivating example, consider the problem of estimating the number of mutations in a viral genome (such as that of SARS-Cov-19) during the \emph{early stages of an outbreak}. Viral RNA/DNA is usually of length $\times10,000$ and testing is time consuming and expensive, and additionally hampered by privacy issues. Consequently, a small number of sequenced genomes may be available when trying to determine in a timely manner if the virus is mutating at a high rate and therefore potentially dangerous to the population (Note that large body of work reports mutational rates of viruses as indicators of their virulence and potential to cause epidemic and pandemic outbreaks~\cite{hoenen2015mutation,ribeiro2012quantifying}.). In this particular case, the actual alphabet size is known and equal to the length of the genome, but not all genomic sites are subject to mutations. Furthermore, sequencing errors introduce counting artifacts, and so do sampling biases which are caused by some individuals being tested multiple times (e.g., health workers~\cite{owidcoronavirus}) or not tested at all. To address these issues, we propose a novel noisy support estimation problem under the Poisson repeat channel~\cite{cheraghchi2019sharp} model. The Poisson repeat channel models both deletion and repetitions of particular mutational sites, and is adequate for capturing unknown sampling biases and sequencing error phenomena. To the best of our knowledge, this work is the first to consider the noisy support estimation problem where symbols are potentially repeated or missed. The only other line of work addressing a similar problem was reported in~\cite{farnoud2009small}, with a focus on Good-Turing distribution estimation in the presence of sample insertion errors.

We address the noisy small sample support estimation problem under the Poisson repeat channel by using novel \emph{regularized weighted} Chebyshev approximation techniques. Weighted polynomial approximation techniques are largely unknown in the machine learning community~\cite{lubinsky2007survey} since all previous approximation based methods proposed so far have only focused on unweighted and noiseless settings~\cite{wu2019chebyshev,wu2018sample,han2018local,yi2018data}. As will be shown in our subsequent analysis, exponential smoothing and ``noise-compensating'' weights play a major role in improving the performance of polynomial methods as well as making them computationally tractable. Within this framework, the Mhaskar-Saff theorem and extensions thereof presented in the work are of great importance~\cite{lubinsky2007survey,mhaskar1984extremal,mhaskar1985does}. In addition, our regularization term arises from consideration of the variance of the estimators, as the weighted Chebyshev approximation component only takes into account the bias. Hence, solving our regularized weighted Chebyshev approximation problem is equivalent to jointly optimizing the bias and variance of the estimator. In addition, we show that Touchard (Bell) polynomials~\cite{touchard1939cycles} naturally arise when incorporating the Poisson repeat channel into the model. To numerically solve the underlying optimization problem we use discretized semi-infinite programming (SIP) techniques. We prove that the solution of discretized SIP converges to the true unique optimal solution for the noisy support estimation problem. Through extensive experiments on both synthetic and real-world data, we show that our methods are able to accurately estimate the support size under the Poisson repeat channel.

\paragraph{Prior work}
Noiseless support estimation methods operating in the small sample regime can be roughly grouped into two categories~\cite{valiant2013estimating,wu2019chebyshev,acharya2017unified,wu2018sample,pavlichin2017approximate,han2018local,yi2018data}. The first line of works~\cite{valiant2013estimating,acharya2017unified,pavlichin2017approximate} makes use of the maximum likelihood principle. While~\cite{acharya2017unified} constructs estimators based on the Profile Maximum Likelihood (PML)~\cite{orlitsky2004modeling}, the work reported in~\cite{valiant2013estimating} focuses on Sequence Maximum Likelihood (SML) estimators~\cite{aldrich1997ra}. The main advantage of ML-based methods is that they easily generalize to many other estimation tasks. For example, the authors of~\cite{acharya2017unified} showed that a single method may be used for entropy estimation, support coverage and distance to uniformity analysis. However, most ML-based estimators require large computational resources~\cite{pavlichin2017approximate,wu2019chebyshev}. To address the computational issue, a sophisticated approximate PML technique that reduces the computational complexity of support estimation at the expense of some performance loss was proposed in~\cite{pavlichin2017approximate}.
On the other hand, the second line of works~\cite{wu2019chebyshev,wu2018sample,han2018local,yi2018data} formulates support estimation as an approximation problem. The underlying methods, which we henceforth refer to as approximation-based methods, design estimators by minimizing the worst case risk. In particular,~\cite{wu2019chebyshev} uses shifted and scaled Chebyshev polynomials of the first kind to construct efficient estimators. In contrast, the authors of~\cite{yi2018data} suggest disposing of minmax estimators and implementing a data amplification technique with analytical performance guarantees. The aforementioned estimator is based on polynomial smoothing~\cite{orlitsky2016optimal} related to approximation techniques. All described approximation-based estimators are computational efficient, with the exception of~\cite{han2018local}, as reported in~\cite{yi2018data}. Sampling artifacts lead to non-iid observations which are very different from Markovian models discussed in~\cite{han2018entropy}.

The paper is organized as follows. In Section~\ref{sec:Probform}, we introduce the relevant notation and the support estimation problem under the Poisson repeat channel. We also describe a class of estimators termed \emph{polynomial class estimators}. In Section ~\ref{sec_bias}, we outline our analysis of polynomial class estimators and describe our main results needed to overcome the technical challenges associated with regularized weighted minmax polynomial approximations. Section~\ref{sec:Simulations} is devoted to experimental verifications and testing, both on synthetically generated data and real-world data. Our real world data includes Shakespeare's plays, used to illustrate the performance of our methods, and a collection of $\sim4,100$ SARS-Cov-2 viral genomes retrieved from GISAID~\cite{elbe2017data,shu2017gisaid}.
%In the latter case, we compute the mutational support from data retrieved on a certain data and compare it to the actual counts obtained at a later date, using additional samples.
%\comment{I think we should just focus on SARS-Cov-2 data. As it's more interesting and relevant to our setting.}
% , metagenomic data samples from the NIH Human Microbiome~\cite{peterson2009nih} and American Gut Microbiome project~\cite{McDonalde00031-18}

\section{Problem formulation and a new class of polynomial estimators}\label{sec:Probform}

Let $P = (p_1,p_2,\ldots)$ be a discrete distribution over some countable alphabet and let $X_1,\ldots,X_n$ be i.i.d. samples drawn according to the distribution $P$. The problem of interest is to estimate the support size, defined as $S(P) = \sum_{i}\mathbf{1}_{\{p_i>0\}}$ where $\mathbf{1}_A$ stands for the indicator of the event $A$. When clear from the context we use $S$ instead of $S(P)$ to avoid notational clutter. We make the assumption that the minimum nonzero probability of the distribution $P$ is at least $1/k,$ for some $k\in \mathbb{R}^{+}$, i.e., $\inf\{p\in P\,| \, p>0\}\geq \frac{1}{k}$. Furthermore, we let $D_k$ denote the space of all probability distribution satisfying $\inf\{p\in P\, |\, p>0\}\geq \frac{1}{k}$. Clearly, $S \leq k$, $\forall P\in D_k$. A sufficient statistics for $X_1,\ldots,X_n$ is the empirical distribution (i.e., histogram) $N = (N_1,N_2,\ldots),$ where $N_i = \sum_{j=1}^{n}\mathbf{1}_{\{X_j=i\}}$.

\textbf{The Poisson repeat channel. }For each sample index $1 \leq i \leq n$, the Poisson repeat channel with parameter $\eta$ outputs $R_i$ copies of the sample $X_i$, where $R_i\sim Poisson(\eta)$ is a Poisson distributed random variable with mean $\eta$. For simplicity of analysis, the variables $R_i$ are assumed to be i.i.d. and the corresponding channel memoryless; the value $R_i = 0$ indicates that a sample input $X_i$ has been deleted.
We also define the empirical distribution of the output sequence of the Poisson channel as $N' = (N_1',N_2',...)$, where $N_i' = \sum_{j=1}^{n}R_j\mathbf{1}_{\{X_j=i\}}$. Throughout the paper, we assume that the parameter $\eta$ is known although it is possible to learn it simultaneously with the support.

The focal point of our analysis is to upper bound the minmax risk under normalized squared loss
\begin{equation} \label{eq:minimax}
    R^\ast(n,k) = \inf_{\hat{S}}\sup_{P\in D_k} \mathbb{E}\left[\left(\frac{\hat{S}(N)-S}{k}\right)^2\right]\;\; \text{ and } \;\;\inf_{\hat{S}}\sup_{P\in D_k} \mathbb{E}\left[\left(\frac{\hat{S}(N')-S}{k}\right)^2\right],
\end{equation}
which correspond to the case without and with Poisson repeat channel, respectively.

We focus on the case including the Poisson repeat channel, as a similar analysis can be easily performed for the case without Poisson repeats. We seek a support estimator $\hat{S}$ that minimizes
\begin{align*}
&\sup_{P\in D_k} \mathbb{E}\left[\left(\frac{\hat{S}(N')-S}{k}\right)^2\right]= \sup_{P\in D_k} \left[ \mathbb{E}^2\left(\frac{\hat{S}(N')-S}{k}\right)+var\left(\frac{\hat{S}(N')-S}{k}\right) \right]. \notag
\end{align*}
The first term within the supremum captures the expected bias of the estimator $\hat{S}$. The second term represents the variance of the estimator $\hat{S}$. Hence, ``good'' estimators are required to balance out the worst-case contributions of
the bias and variance.

We define a class of polynomial based estimators as follows. Given a parameter $L \in \mathbb{N}$, we say that an estimator $\hat{S}$ is a polynomial class estimator with the parameter $L$ (i.e.,  a $Poly(L)$ estimator) if it takes the form $\hat{S} = \sum_{i}g_L(N_i),$ where $g_L$ is defined as
\begin{equation}\label{eq:gL}
g_L(j) =
    \begin{cases}
          a_j j!+1,  &\mbox{if } j\leq L\\
          1,         &\mbox{ otherwise. }
    \end{cases}
\end{equation}
Here, $a_j \in \mathbb{R},$ and $a_0 = -1,$ since this choice ensures that $g_L(0) = 0$. One can associate an estimator $\hat{S}$ with its corresponding coefficients $\mathbf{a}$, and define a family of estimators
\begin{equation*}
    Poly(L) = \left\{\mathbf{a}\in \mathbb{R}^{L+1}|a_0 = -1\right\}.
\end{equation*}
%\textcolor{red}{Would it not make sense to have one coefficient without $j!$? Also, the name poly maybe misleading?}
% Clearly, $\tilde{S} \in Poly(L)$, with corresponding coefficients
% \begin{equation}\label{biaseq5}
%   \mathbf{\tilde{a}} = \argmin_{\mathbf{a}\in Poly(L)}\, \sup_{\lambda\in [l=\frac{n}{k},r=c_1\log k]} | \sum_{l=0}^{L} a_l \lambda_i^l  |.
% \end{equation}
Next, we show that the problem of minimizing worst-case risk within the class $Poly(L)$ can be cast as a regularized exponentially weighted Chebyshev approximation problem~\cite{lubinsky2007survey}.

\section{Estimator analysis}\label{sec_bias}
We start by analyzing the minmax risk under the Poisson channel model through Poissonization arguments. These assert that the number of samples drawn is Poisson distributed $N\sim Poisson(n)$ and that the counts $N_i\sim Poisson(\lambda_i)$ are independent, where $\lambda_i = np_i$. Poissonization was also used in~\cite{wu2019chebyshev,han2018local} to derive the following tight upper bound on the minmax risk.
\begin{lemma}[Lemma 1 in~\cite{wu2019chebyshev}]\label{lma:Poissonization}
Let $R_P^\ast(n,k)$ be the minmax risk under the Poissonized model. Then, for any $\beta > 1$ we have
\begin{align}
    & R^\ast(n,k) \leq \frac{R_P^\ast((1-\beta)n,k)}{1-\exp(-n\beta^2/2)}.
\end{align}
\end{lemma}
Note that it is straightforward to show that Lemma~\ref{lma:Poissonization} is also valid under the Poisson repeat channel. Let $\mathcal{L} = \{\ell|\lambda_{\ell}>0\}$ be the set of symbols with positive probability. A simple calculation reveals that for any $\hat{S}\in Poly(L)$, one has
% \small
\begin{align}\label{PRC:eq1}
     \mathbb{E}\bigg(\frac{\hat{S}(N')-S}{k}\bigg)^2 &= \frac{1}{k^2}\bigg\{\sum_{i\in \mathcal{L}}\mathbb{E}\bigg(g_L(N_i')-1\bigg)^2+\sum_{i\neq j\in \mathcal{L}}\mathbb{E}\bigg(g(N_i')-1\bigg)\mathbb{E}\bigg(g(N_i')-1\bigg)\bigg\} \nonumber \\
    & \leq \frac{1}{k^2}\bigg\{\sum_{i\in \mathcal{L}}\mathbb{E}\bigg(g_L(N_i')-1\bigg)^2+(S-1)\sum_{i\in \mathcal{L}}\bigg(\mathbb{E}\bigg(g(N_i')-1\bigg)\bigg)^2\bigg\},
\end{align}
\normalsize
where the inequality we use Cauchy-Bunyakowski-Schwarz inequality for the cross terms. Note that the first term captures the variance while the second term is related to the bias. An interesting observation is that the objective function above is symmetric in the parameters $\lambda_i$. The little-known problem of establishing when the optima of such constrained symmetric functions is achieved for the case that all parameters are equal has been studied in~\cite{waterhouse1983symmetric}.

We first analyze the bias term $\mathbb{E}\left(g(N_i')-1\right)$. By the definition of the Poisson repeat channel, we have $N_i'\sim Poisson(\eta N_i)$, which allows us to write $\mathbb{E}\left(g(N_i')-1\right)$ as
% \small
\begin{align}\label{PRC:eq2}
% = \mathbb{E}\bigg(a_{N_i'}(N_i')!\mathbf{1}[N_i'\leq L]\bigg)
    & \sum_{h = 0}^{\infty}\mathbb{P}(N_i = h)\sum_{l=0}^{L} \mathbb{P}(N_i'=l|N_i=h)\bigg(a_{l}l!\bigg) = \sum_{l=0}^{L}\sum_{h = 0}^{\infty} \frac{\lambda_i^h}{h!}e^{-\lambda_i}\times\frac{(\eta h)^l}{l!}e^{-\eta h}\times a_l l!\nonumber\\
     & = \sum_{l=0}^{L}e^{-\lambda_i}a_l \eta^l \sum_{h = 0}^{\infty}\frac{(\lambda_i e^{-\eta})^h}{h!}h^l\times \frac{e^{-\lambda_i e^{-\eta}}}{e^{-\lambda_i e^{-\eta}}} = e^{-\lambda_i(1-e^{-\eta})}\sum_{l=0}^{L}a_l \eta^l M_{N_i^\ast}^{(l)}(0),
\end{align}
\normalsize
where $M_{N_i^\ast}$ is the moment generating function (MGF) of the random variable $N_i^\ast \sim Poisson(\lambda_i e^{-\eta})$. We use $M_{N_i^\ast}^{(l)}(0)$ to denote the $l^{th}$ derivative of the MGF at $0$. It is worth noting that $M_{N_i^\ast}^{(l)}(0)$ has a closed form expression of the form of Touchard (Bell) polynomials~\cite{touchard1939cycles} in $\lambda_i e^{-\eta}$. More specifically, we have $M_{N_i^\ast}^{(l)}(0) = \sum_{r = 0}^{l} \stirling{l}{r}(\lambda_i e^{-\eta})^r,$
% \begin{align}
%     & M_{N_i^\ast}^{(l)}(0) = \sum_{r = 0}^{l} \stirling{l}{r}(\lambda_i e^{-\eta})^r,
% \end{align}
where $\stirling{l}{r}$ denotes the Stirling number of the second kind, counting the number of partitions of a set of size $l$ into $r$ disjoint nonempty subsets. Following a procedure similar to the one used for the bias, one can be shown that the term corresponding to the variance equals
\begin{align}\label{PRC:eq3}
    &\mathbb{E}\bigg(g(N_i')-1\bigg)^2 = e^{-\lambda_i(1-e^{-\eta})}\sum_{l=0}^{L}a_l^2 \eta^l l! M_{N_i^\ast}^{(l)}(0).
\end{align}
By plugging \eqref{PRC:eq2}, \eqref{PRC:eq3} into \eqref{PRC:eq1} and taking the supremum over $D_k$, we have
% \small
\begin{align}
    &\sup_{P\in D_k}\mathbb{E}\bigg(\frac{\hat{S}(N')-S}{k}\bigg)^2 \leq \sup_{\lambda_{\ell}\in [\frac{n}{k}, n],\;\ell \in \mathcal{L}}\mathbb{E}\bigg(\frac{\hat{S}(N')-S}{k}\bigg)^2\label{PRC:eq4}\\
    & \leq \sup_{\lambda_{\ell}\in [\frac{n}{k}, n],\;\ell \in \mathcal{L}} \frac{1}{k^2}\bigg\{\sum_{i\in \mathcal{L} }\mathbb{E}\bigg(g_L(N_i')-1\bigg)^2+(S-1)\sum_{i\in \mathcal{L}}\bigg(\mathbb{E}\bigg(g(N_i')-1\bigg)\bigg)^2\bigg\} \label{PRC:eq5}\\
    & \leq \sup_{\lambda\in [\frac{n}{k},n]} \bigg\{\frac{1}{k}e^{-\lambda(1-e^{-\eta})}\sum_{l=0}^{L}a_l^2 \eta^l l! M_{N^\ast}^{(l)}(0)+\bigg(e^{-\lambda(1-e^{-\eta})}\sum_{l=0}^{L}a_l \eta^l M_{N^\ast}^{(l)}(0)\bigg)^2\bigg\},\label{PRC:eq6}
\end{align}
\normalsize
where $N^\ast \sim Poisson(\lambda e^{-\eta})$. The inequality~\eqref{PRC:eq4} is due to the increase in the domain over which we take the supremum. The inequality~\eqref{PRC:eq5} follows from~\eqref{PRC:eq1}. The last inequality is a consequence of $|\mathcal{L}| = S\leq k$ and the fact that all terms in the summation are nonnegative. Hence we have to minimize an objective with respect to the coefficients $a_1,...,a_L$ according to
\begin{align}\label{PRC:obj1}
    \inf_{\mathbf{a}\in Poly(L)}\sup_{\lambda\in [\frac{n}{k},n]} \bigg\{\frac{1}{k}e^{-\lambda(1-e^{-\eta})}\sum_{l=0}^{L}a_l^2 \eta^l l! M_{N^\ast}^{(l)}(0)+\bigg(e^{-\lambda(1-e^{-\eta})}\sum_{l=0}^{L}a_l \eta^l M_{N^\ast}^{(l)}(0)\bigg)^2\bigg\}.
\end{align}
For the case that the Poisson repeat channel is not present, one only needs to adjust the exponential weights and substitute $\eta^lM_{N^\ast}^{(l)}(0)$ with $\lambda^l$. The resulting optimization problem reads as
\begin{align}\label{PRC:obj2}
    \inf_{\mathbf{a}\in Poly(L)}\sup_{\lambda\in [\frac{n}{k}, n]}\bigg\{\frac{e^{-\lambda}}{k}\bigg(\sum_{l=0}^{L}a_l^2\lambda^ll!\bigg)+\bigg(e^{-\lambda}\sum_{l=0}^{L}a_l\lambda^l\bigg)^2\bigg\}.
\end{align}
Note that both~\eqref{PRC:obj1} and~\eqref{PRC:obj2} are of the form of a regularized weighted Chebychev approximation problem.
For simplicity, we first focus on the noiseless case~\eqref{PRC:obj2}, as similar but more tedious arguments may be used for the noisy case~\eqref{PRC:obj1}.

If we ignore the first term in~\eqref{PRC:obj2}, the optimization problem reads as
\begin{align}\label{eq:WCapprox}
    \inf_{\mathbf{a}\in Poly(L)}\sup_{\lambda\in [\frac{n}{k}, n]}\bigg(e^{-\lambda}\sum_{l=0}^{L}a_l\lambda^l\bigg)^2 \Leftrightarrow \inf_{\mathbf{a}\in Poly(L)}\sup_{\lambda\in [\frac{n}{k}, n]}\bigg |e^{-\lambda}\sum_{l=0}^{L}a_l\lambda^l\bigg|.
\end{align}
The term $e^{-\lambda}\sum_{l=0}^{L}a_l\lambda^l$ corresponds to the bias of the estimator. It is straightforward to see that the optimal choice of $\mathbf{a}$ for the two problems are the same.
% \begin{align}\label{eq:WCapprox}
%     \inf_{\mathbf{a}\in Poly(L)}\sup_{\lambda\in [\frac{n}{k}, n]}\bigg |e^{-\lambda}\sum_{l=0}^{L}a_l\lambda^l\bigg|.
% \end{align}
Problem~\eqref{eq:WCapprox} is an exponentially weighted Chebyshev approximation problem~\cite{mason2002chebyshev}. Note that one can further upper bound~\eqref{eq:WCapprox} as follows
\begin{align}\label{eq:WCapprox2}
    \inf_{\mathbf{a}\in Poly(L)}\sup_{\lambda\in [\frac{n}{k}, n]}\bigg |e^{-\lambda}\sum_{l=0}^{L}a_l\lambda^l\bigg| \leq e^{-\frac{n}{k}}\inf_{\mathbf{a}\in Poly(L)}\sup_{\lambda\in [\frac{n}{k}, n]}\bigg |\sum_{l=0}^{L}a_l\lambda^l\bigg|,
\end{align}
resulting in a standard Chebyshev approximation problem with a solution of the form of scaled and shifted Chebyshev polynomials. Despite the fact that the authors of~\cite{wu2019chebyshev} obtained the coefficients of the Chebyshev estimator using a different interval in the supremum that account for the variance, \eqref{eq:WCapprox2} along with our extensive simulation results show that ignoring the exponential weights results in a worse bound on the risk and practical performance.

The first term $\frac{1}{k}\left(\sum_{l=0}^{L}e^{-\lambda}a_l^2\lambda^ll!\right)$, which corresponds to the variance, may be rewritten as
\begin{align*}
    & \frac{1}{k}\bigg(\sum_{l=0}^{L}e^{-\lambda}a_l^2\lambda^ll!\bigg) = \mathbf{a}^T\mathbf{M}(\lambda)\mathbf{a} \triangleq ||\mathbf{a}||_{\mathbf{M}(\lambda)}^2,\;\mathbf{M}(\lambda) \triangleq \frac{e^{-\lambda}}{k} \,Diag(\lambda^{0}0!,\lambda^{1}1!,...,\lambda^{L}L!).
\end{align*}
Clearly, $||.||_{\mathbf{M}(\lambda)}$ is a valid norm, and consequently, the first term in~\eqref{PRC:obj2} may be viewed as a regularizer. For the case including the Poisson repeat channel, since the sum of Touchard polynomials is still a polynomial, the bias term in~\eqref{PRC:obj1} also represents an exponentially weighted Chebyshev approximation problem. The variance term may be written as the weighted norm of $\mathbf{a}$ with a weight matrix $\frac{1}{k}e^{-\lambda(1-e^{-\eta})}\,Diag(0!\eta^0M_{N^\ast}^{(0)}(0),1!\eta^1M_{N^\ast}^{(1)}(0),...,L!\eta^LM_{N^\ast}^{(L)}(0))$. The resulting problem~\eqref{PRC:obj1} is once again an regularized weighted Chebyshev approximation problem.

\paragraph{Solving problems~\eqref{PRC:obj1} and \eqref{PRC:obj2}. }
%To directly solve the optimization problem of interest, we use semi-infinite programming (SIP).
Solving problem~\eqref{PRC:obj1} and \eqref{PRC:obj2} directly appears to be difficult, so we instead resort to numerically solving the epigraph formulation of the semi-infinite programs~\eqref{PRC:obj1} and \eqref{PRC:obj2} and proving that the numerical solution is asymptotically consistent.

Once again, we start with the case that excludes the Poisson repeat channel~\eqref{PRC:obj2}. The epigraph formulation of~\eqref{PRC:obj2} is a semi-infinite program of the form (\cite{boyd2004convex}, Chapter 6.1)
% \small
\begin{equation}\label{sqlosseq2}
    \begin{split}
        &\min_{t,\mathbf{a}\in Poly(L)} t \;\;\text{ s.t.}\;\; \bigg\{\frac{1}{k}\bigg(\sum_{l=0}^{L}e^{-\lambda}a_l^2\lambda^ll!\bigg)+\bigg(e^{-\lambda}\sum_{l=0}^{L}a_l\lambda^l\bigg)^2\bigg\}\leq t,\;\forall \lambda\in [\frac{n}{k}, n].
    \end{split}
\end{equation}
\normalsize
There are many algorithms that can be used to numerically solve~\eqref{sqlosseq2}: the discretization and central cutting plane method, the KKT and SQP reductions ~\cite{lopez2007semi,reemtsen1998numerical}. For simplicity, we focus on the discretization method. For this purpose, we first form a grid of the interval $[\frac{n}{k}, n]$ involving $s$ points, denoted by $\text{Grid}([\frac{n}{k}, n],s)$. Problem~\eqref{sqlosseq2} represents an LP with infinitely many quadratic constraints, which is not solvable. Hence, instead of solving~\eqref{sqlosseq2}, we focus on the relaxed problem
% \small
\begin{equation}\label{sqlosseq3}
    \begin{split}
        &\min_{t,\mathbf{a}\in Poly(L)} t \;\;\text{ s.t.}\;\; \bigg\{\frac{1}{k}\bigg(\sum_{l=0}^{L}e^{-\lambda}a_l^2\lambda^ll!\bigg)+\bigg(e^{-\lambda}\sum_{l=0}^{L}a_l\lambda^l\bigg)^2\bigg\}\leq t,\;\forall \lambda\in \text{Grid}([\frac{n}{k}, n],s).
    \end{split}
\end{equation}
\normalsize
As will be discussed in greater detail later, the solution of the relaxed problem is asymptotically consistent with the solution of the original problem (i.e., as $s \to \infty$, the optimal values of the objectives of the original and relaxed problem are the same). Problem~\eqref{sqlosseq3} is an LP with a finite number of quadratic constraints that may be solved using standard optimization tools. Unfortunately, the number of constraints scales with the length of the grid interval, which in the case of interest is linear in $n$. This appears as an undesirable feature of the approach, but can be easily mitigated through the following theorem which demonstrates that an optimal solution of the problem may be found over an interval of length proportional to the significantly smaller value of $\log\,k$ ($k/\log\,k \lesssim n$ is needed for accurate estimation~\cite{wu2019chebyshev}). We relegate the proof of this result to the Supplement.
%In order to solve this problem, we are going to show that for all $\mathbf{a}\in Poly(L)$, the optimal $\lambda$ which attain the supremum in \eqref{eq:WCapprox} will fall into a much smaller interval.
\begin{theorem}\label{lma:interval_sq}
    For any $\mathbf{a}\in Poly(L),$ $L=\lfloor c_0 \, \log\,k \rfloor$ and $c_0 = 0.558$, let
    \begin{equation*}
        g(\mathbf{a},\lambda) = \frac{1}{k}\bigg(\sum_{l=0}^{L}e^{-\lambda}a_l^2\lambda^ll!\bigg)+\bigg(e^{-\lambda}\sum_{l=0}^{L}a_l\lambda^l\bigg)^2.
    \end{equation*}
    Then, we have
    % \small
    \begin{align*}
        \sup_{\lambda\in [\frac{n}{k}, n]}g(\mathbf{a},\lambda)=
        \begin{cases}
            \sup_{\lambda\in [\frac{n}{k}, 6.5L]}g(\mathbf{a},\lambda) & \text{if }\frac{n}{k}\leq 6.5L\\
            g(\mathbf{a},\frac{n}{k}) & \text{if }\frac{n}{k}> 6.5L.
        \end{cases}
    \end{align*}
    \normalsize
%with $L=\lfloor c_0 \, \log\,k \rfloor$. \textcolor{red}{(Maybe we should remove this since we have an explanation right below \eqref{sqlosseq5}.)}
\end{theorem}
\begin{remark}\label{remark:pi_interval}
In weighted approximation theory~\cite{lubinsky2007survey}, the problem of bounding the interval over which the supremum is achieved is a topic of significant interest, with many important results readily available. For example, if we ignore the regularization term, we can directly use the Mhaskar-Saff theorem~\cite{mhaskar1984extremal,mhaskar1985does} (Theorem~\ref{thm:MRS} in the Supplement) to reduce the length of the interval in the supremum to $\frac{\pi}{2}L$. Our Theorem~\ref{lma:interval_sq} shows that even when a regularization term is present, we can still restrict the length of the interval to $6.5L$. Our proof differs from that of the more general Mhaskar-Saff theorem, since we exploit the specific structure of the problem. It remains an open problem to extend the approach of~\cite{lubinsky2007survey} used in our proof to account for more general weights.
\end{remark}

Using the previous derivations, we arrive at the following optimization problem
% \scriptsize
% \small
\begin{equation}\label{sqlosseq5}
    \begin{split}
        &\min_{t,\mathbf{a}\in Poly(L)} t \;\;\text{ s.t.}\;\; \bigg\{\frac{1}{k}\bigg(\sum_{l=0}^{L}e^{-\lambda}a_l^2\lambda^ll!\bigg)+\bigg(e^{-\lambda}\sum_{l=0}^{L}a_l\lambda^l\bigg)^2\bigg\}\leq t,\;\forall \lambda\in \text{Grid}([\frac{n}{k}, 6.5L],s).
    \end{split}
\end{equation}
\normalsize
Since $L=\lfloor c_0\, \log\, k \rfloor$ for the case excluding the Poisson repeat channel, the length of the optimization interval in~\eqref{sqlosseq5} is proportional to $\log\,k$ and thus the~\eqref{sqlosseq5} can be solved efficiently.

For the case including the Poisson repeat channel, using the same arguments as above, we have%obtain the following optimization problem
% \small
\begin{align}\label{PRC_finalobj}
    &\min_{t,\mathbf{a}\in Poly(L)}t \quad \text{s.t. } \bigg\{\frac{1}{k}e^{-\lambda(1-e^{-\eta})}\sum_{l=0}^{L}a_l^2 \eta^l l! M_{N^\ast}^{(l)}(0)+\bigg(e^{-\lambda(1-e^{-\eta})}\sum_{l=0}^{L}a_l \eta^l M_{N^\ast}^{(l)}(0)\bigg)^2\bigg\} \leq t \notag\\
    & \forall \lambda \in \text{Grid}([\frac{n}{k},C L],s),
\end{align}
\normalsize
where $C>0$ is a constant. Unfortunately, it is very hard to precisely characterize $C$. Nevertheless, we find that in practice, the choice $C = 2$ works well with $L = \lfloor \eta c_0\log k \rfloor$. Furthermore, since the Poisson repeat channel introduces an average of $\eta$ replicas of each samples, the ``cut-off'' value $L$ for $g_L$ in~\eqref{eq:gL} is set to be $\eta$ times larger than the corresponding value without Poisson repeats.
\paragraph{Convergence of the discretized method}%\label{subsec:discmethod}
For the case of objective functions and constraints that are ``well-behaved'' (see~\cite{reemtsen1991discretization} and~\cite{still2001discretization}), as $s$ grows, the solution of the relaxed semi-infinite program approaches the optimal solution of the original problem. We use the above results in conjunction with a number of properties of our objective SIP to establish the claim in the following theorem whose proof is delegated to the Supplement.
\begin{theorem}\label{thm:discretization}
    Let $s$ be the number of uniformly placed grid points on the interval~\eqref{sqlosseq5} or~\eqref{PRC_finalobj}, and let $d\triangleq \frac{6.5L-\frac{n}{k}}{s-1}$ be the length of the discretization interval. As $d \rightarrow 0$, the optimal objective value $t_d$ of the discretized SIP~\eqref{sqlosseq5} or~\eqref{PRC_finalobj} (with $\eta>0$) converges to the optimal objective value of the original SIP $t^\star$. Moreover, the optimal solution is unique. The convergence rate of $t_d$ to $t^\star$ equals $O(d^2)$. If the optimal solution of the SIP is a strict minimum of order one (i.e., if $t-t^\star \geq C||\mathbf{a}-\mathbf{a}^\star||$ for some constant $C>0$ and for all feasible neighborhoods of $\mathbf{a}^\star$), then the solution of the discretized SIP also converges to an optimal solution with rate $O(d^2)$.
\end{theorem}
\section{Experiments}\label{sec:Simulations}
Next, we compare our estimator, referred to as the Regularized Weighted Chebyshev (RWC) method, with the Good-Turing (GT) estimator, the WY estimator of~\cite{wu2019chebyshev}, the PJW estimator described in~\cite{pavlichin2017approximate} and the HOSW estimator of~\cite{yi2018data}. We do not compare our method with the estimators introduced in~\cite{valiant2013estimating,han2018local} due to their high computational complexity~\cite{wu2019chebyshev,yi2018data}.

\paragraph{Synthetic data experiments.}
We first evaluate the maximum risk under normalized squared loss of all listed estimators over six different distributions without Poisson repeats: the uniform distribution with $p_i = \frac{1}{k}$, the Zipf distributions with $p_i\propto i^{-\alpha}$, and $\alpha$ equal to $1.5$, $1$, $0.5$ or $0.25$, and the Benford distribution with $p_i \propto \log(i+1)-\log(i)$. We choose the support sizes for the Zipf and Benford distribution so that the minimum nonzero probability mass is roughly $10^{-6}$. We run the estimator $100$ times to calculate the risk. For solving~\eqref{sqlosseq5}, we use a grid with $s=1000$ points in the interval $[\frac{n}{k},6.5L],$ and $L = \lfloor 0.558 \log k \rfloor$. The GT method used for comparison first estimates the total probability of seen symbols (e.g., sample coverage) according to $\hat{C} = 1-h_1/n,$ and then estimates the support size according to $\hat{S}_{GT} = \hat{S}_{\text{c}}/\hat{C}$; here, $\hat{S}_{\text{c}}$ stands for the simple counting estimator. Note that $h_1$ equals the number of different alphabet symbols observed only once in the $n$ samples. Detailed findings are presented in the Supplement.

Figures~\ref{fig:maxresult} and~\ref{fig:maxresult_gain} indicate that the RWC estimator has a significantly better worst case performance compared to all other methods when tested on the above distributions, provided that $n\geq 0.2k$. Also, both the RWC and WY estimators have significantly better error exponents compared to GT, PJW and HOSW. Interestingly, we find the the worst case risk with normalization $(1/k)^2$ tends to severely bias the results towards a near-uniform distribution. We mitigate this issue by changing the normalization from $(1/k)^2$ to $(1/S)^2,$ which was also done in~\cite{yi2018data}. We repeat the experiment using the normalization $(1/S)^2$, corresponding to what we refer to as the RWC-S estimator. A detailed description of this algorithm and its analysis is available in the Supplement. Figures~\ref{fig:maxresult_S} and~\ref{fig:maxresult_S_gain} illustrate that the RWC-S estimator significantly outperforms all other estimators.
%Moreover, the RWC-S estimator also outperforms all known estimators on almost all tested distributions.

Next, we turn our attention to the case of the Poisson repeat channel (PRC). Our RWC-S-prc estimator requires solving~\eqref{PRC_finalobj} with $1/k$ replaced by $1/\hat{S}_{\text{c}}$, and setting $C = 2,$ $L = \lfloor 0.558\eta \log k \rfloor$ and $s = 1000$. Since the noisy support estimation problem is new there is no standard benchmark to compare it with. This is why we consider the performance of RWC-S-prc, the naive counting estimator and two simple modifications of the WY estimator, since those offer the best performance in the noiseless setting. The WY-naive method first divides the empirical counts $N'$ by $\eta$ and then applies the WY estimator. This is intuitive since each symbol is repeated $\eta$ times in expectation. The WY-prc method involves modifying the coefficients in the WY estimator with Touchard polynomial multipliers. Note that this modification does not take into account the exponential weighting and regularization term that we introduced for both the noiseless and noisy setting. In the experiments, we choose $\eta$ from $\{{0.5,1,1.5\}}$ since the replication rate is small in practice. As we can see from Figures~\ref{fig:prc_eta1},~\ref{fig:prc_eta2} and~\ref{fig:prc_eta3}, our method significantly outperforms all other methods. Notably, WY-prc performs poorly for $\eta>1$ while WY-Naive performs poorly for $\eta<1$. %In contrast, our RWC-S-prc estimator consistently outperform these baselines by a significant margin.
% \textcolor{blue}{
% Figure~\ref{fig:maxresult_S} shows that if our risk is $\E\left(\frac{\hat{S}-S}{S} \right)^2$, then our RWC-S estimator now outperform all the state-of-the-art estimators by a decent improvement. In practice, we suggest to use $\E\left(\frac{\hat{S}-S}{S} \right)^2$ as the performance measure since it more fair for the distributions in $D_k$. In contrast, $\E\left(\frac{\hat{S}-S}{k} \right)^2$ for distributions close to uniform are harder to minimize since the ground truth $S$ is large.
% }
% \textcolor{blue}{
% From the rest figure in~\ref{fig:eachdist}, we can see that our RWC-S estimator has the best performance except for Uniform distribution. Note that we also repeat the experiment with the same setting in~\cite{yi2018data} which is in appendix. We find that HOSW estimator will only work well when $n$ is sufficiently small. For example, in~\cite{yi2018data} they test their estimator with at most $2\times 10^5$ samples. Nevertheless, even in their setting our RWC-S estimator still outperform HOSW estimator.
% }
% \begin{figure}[!htb]
%   \centering
%   \subfigure[MSE normalized by $k^2$\label{fig:maxresult}]{\includegraphics[width=0.49\linewidth]{Worstcase_k.eps}}
%   \subfigure[MSE normalized by $S^2$\label{fig:maxresult_S}]{\includegraphics[width=0.49\linewidth]{Worstcase_S.eps}}
%   \vspace{-0.2cm}
%   \caption{The worst case risk for six selected probability distributions.}
% \end{figure}

\begin{figure}[!htb]
  \centering
  \subfigure[Worst case MSE$/k^2$. \label{fig:maxresult}]{\includegraphics[width=0.26\linewidth]{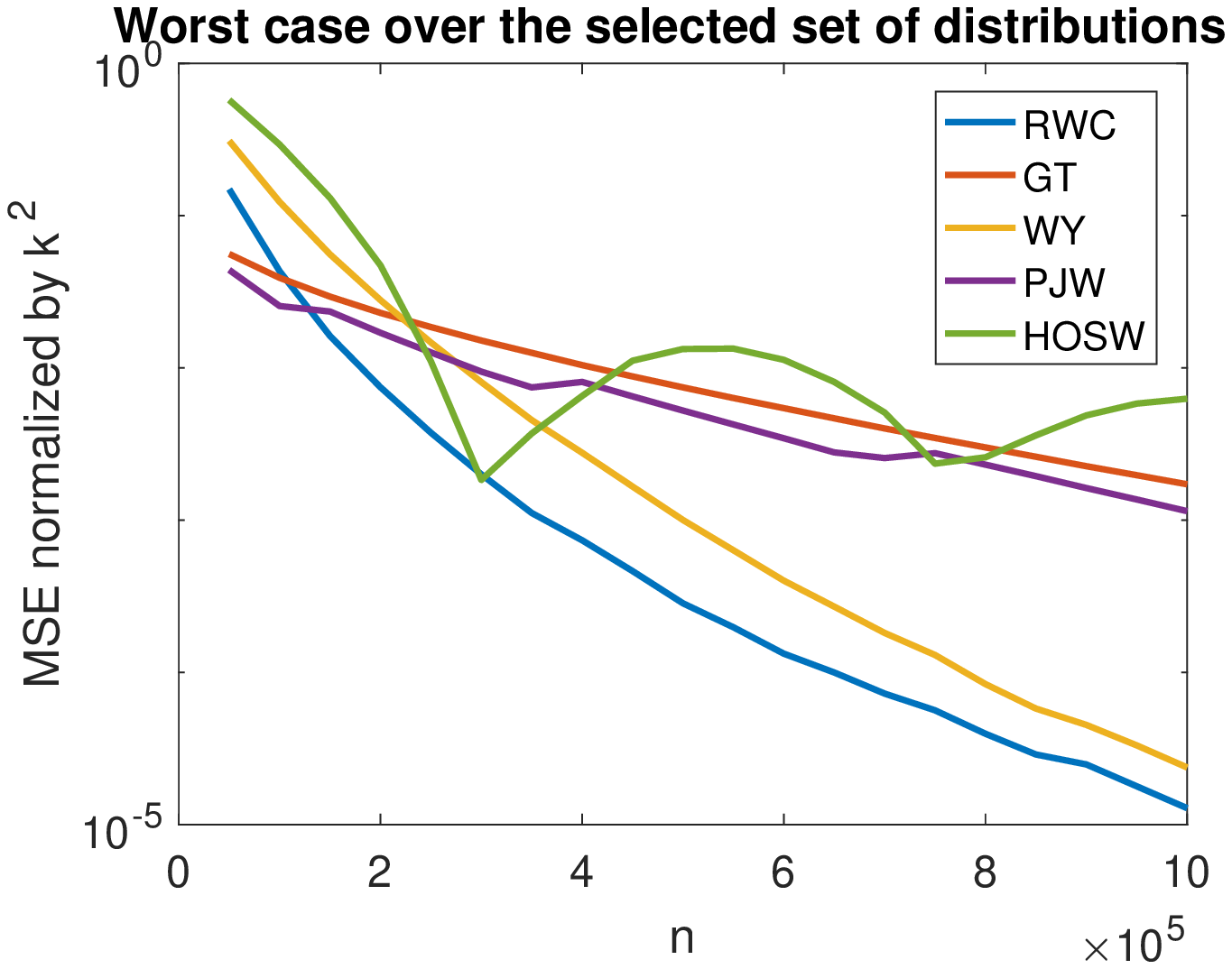}}\hspace{-1em}
  \subfigure[Gain of RWC. \label{fig:maxresult_gain}]{\includegraphics[width=0.26\linewidth]{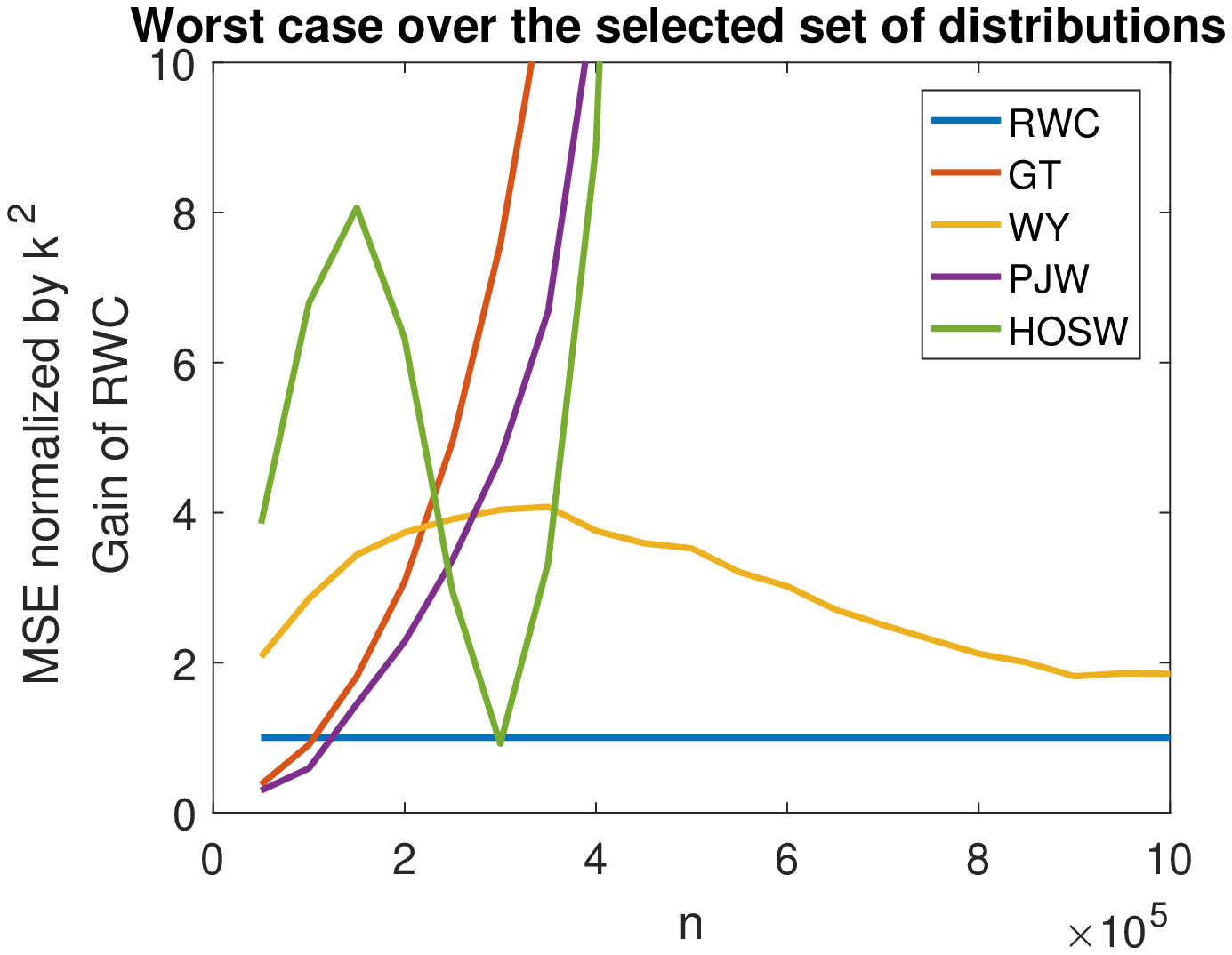}}\hspace{-1em}
  \subfigure[Worst case MSE$/S^2$. \label{fig:maxresult_S}]{\includegraphics[width=0.26\linewidth]{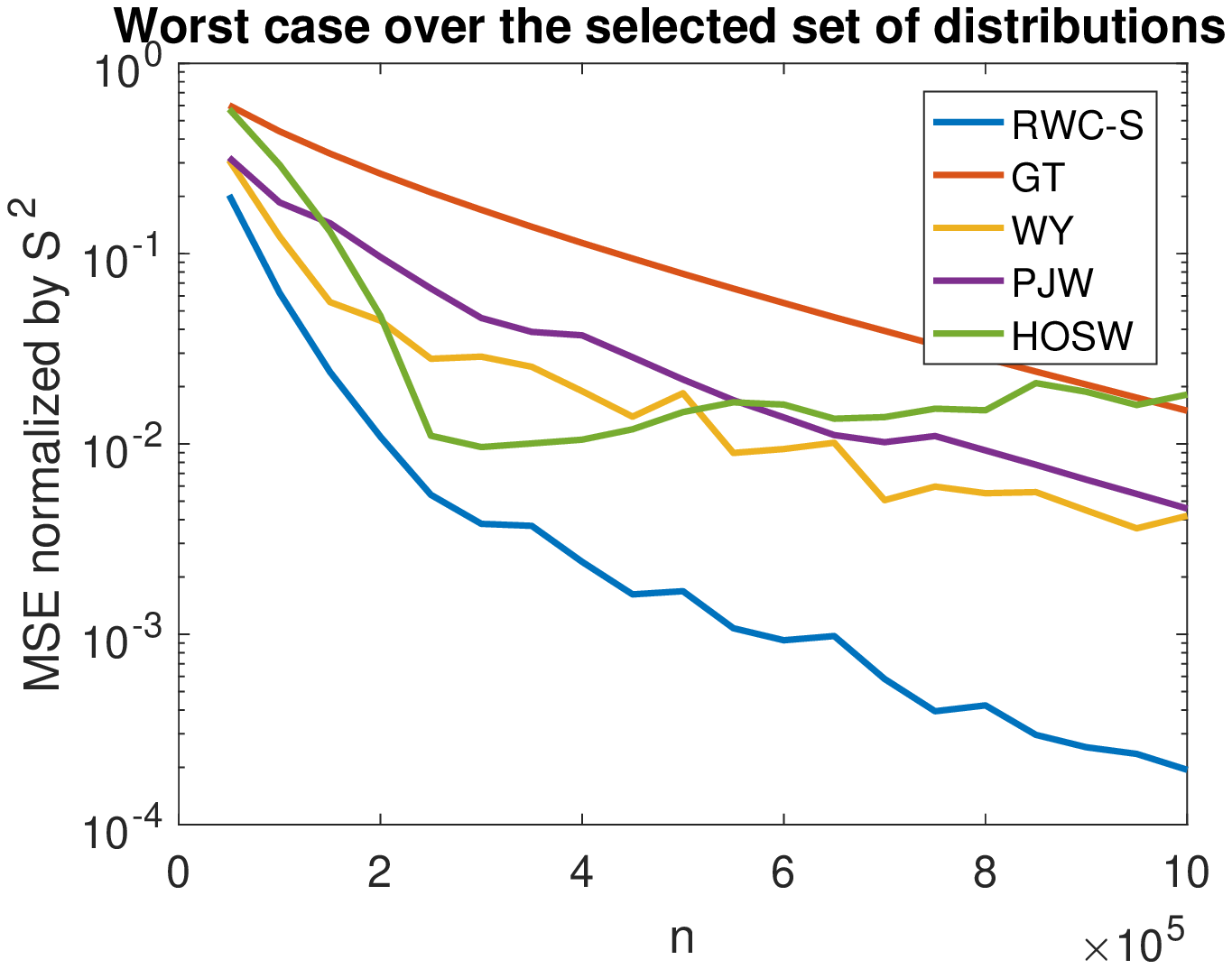}}\hspace{-1em}
  \subfigure[Gain of RWC-S. \label{fig:maxresult_S_gain}]{\includegraphics[width=0.26\linewidth]{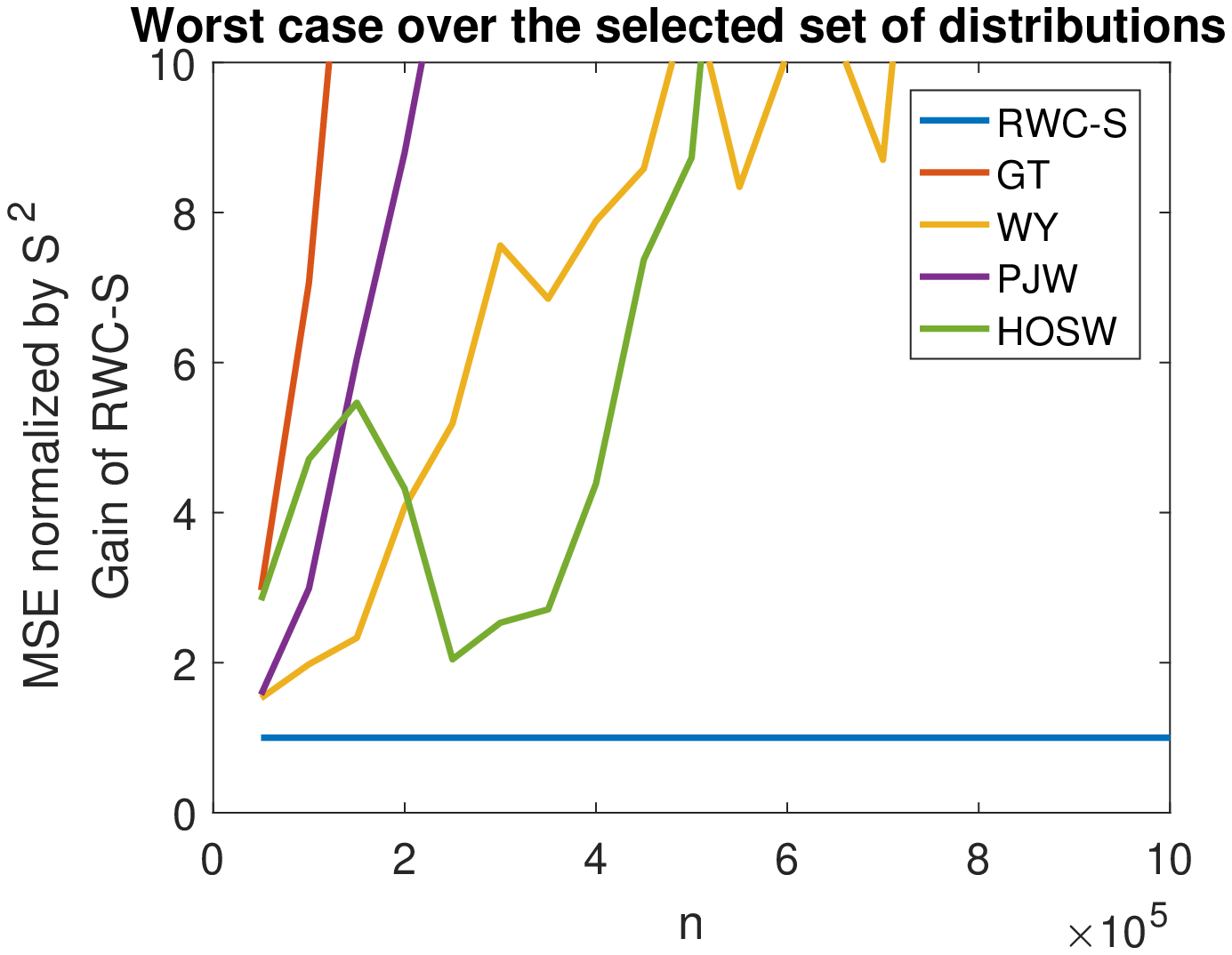}}\\
  \vspace{-0.2cm}
  \subfigure[PRC $\eta = 0.5$ \label{fig:prc_eta1}]{\includegraphics[width=0.26\linewidth]{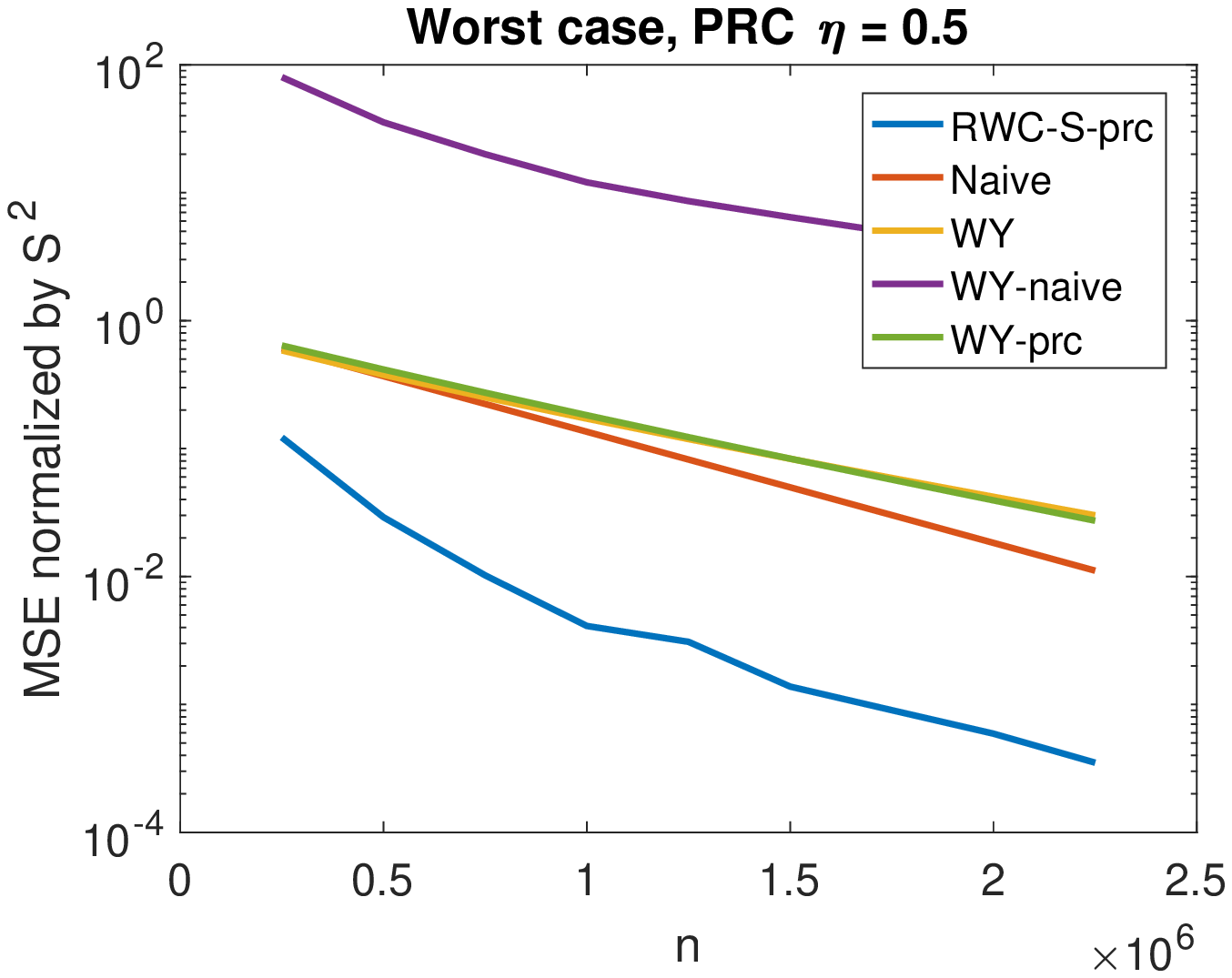}}\hspace{-1em}
  \subfigure[PRC $\eta = 1$ \label{fig:prc_eta2}]{\includegraphics[width=0.26\linewidth]{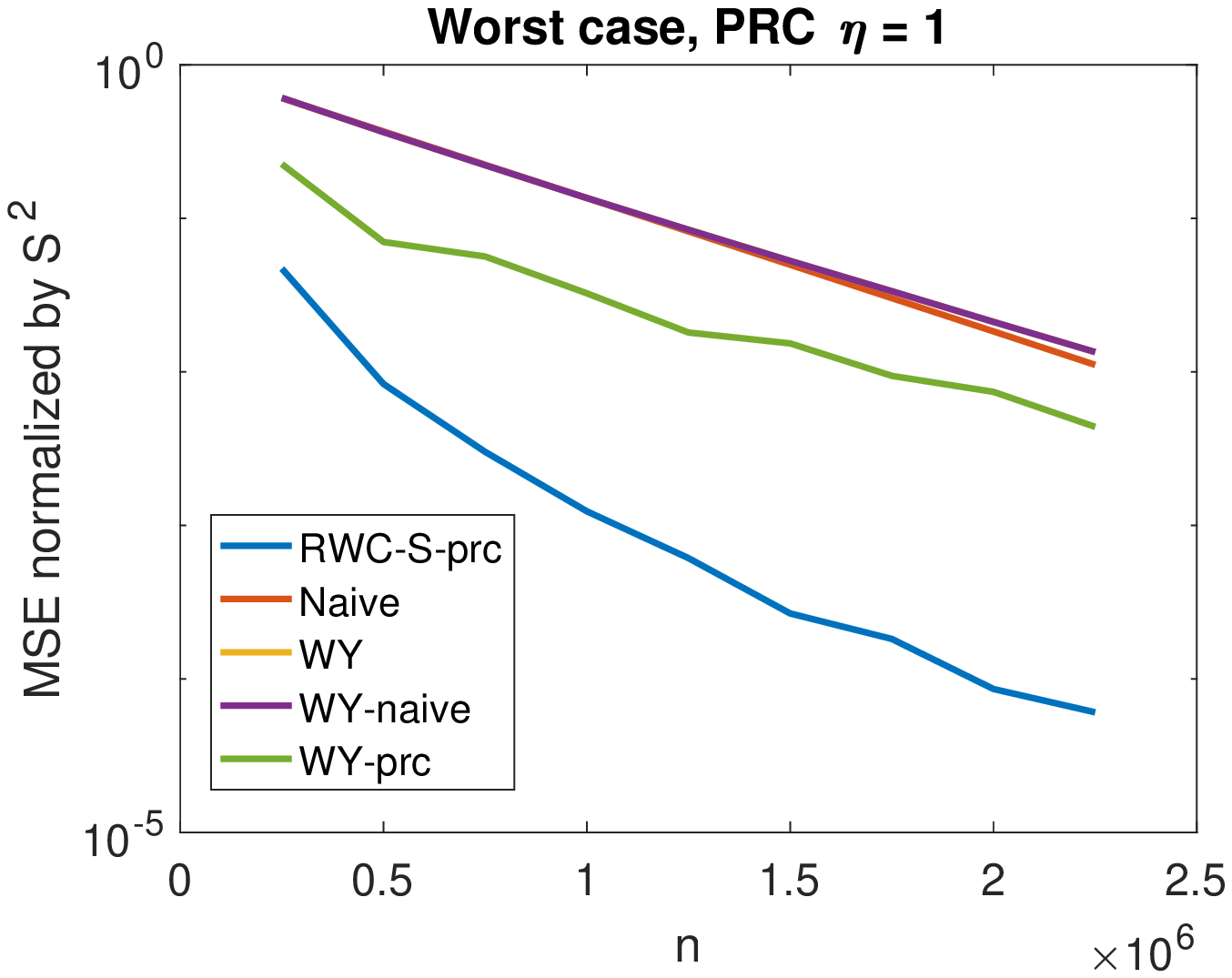}}\hspace{-1em}
  \subfigure[PRC $\eta = 1.5$ \label{fig:prc_eta3}]{\includegraphics[width=0.26\linewidth]{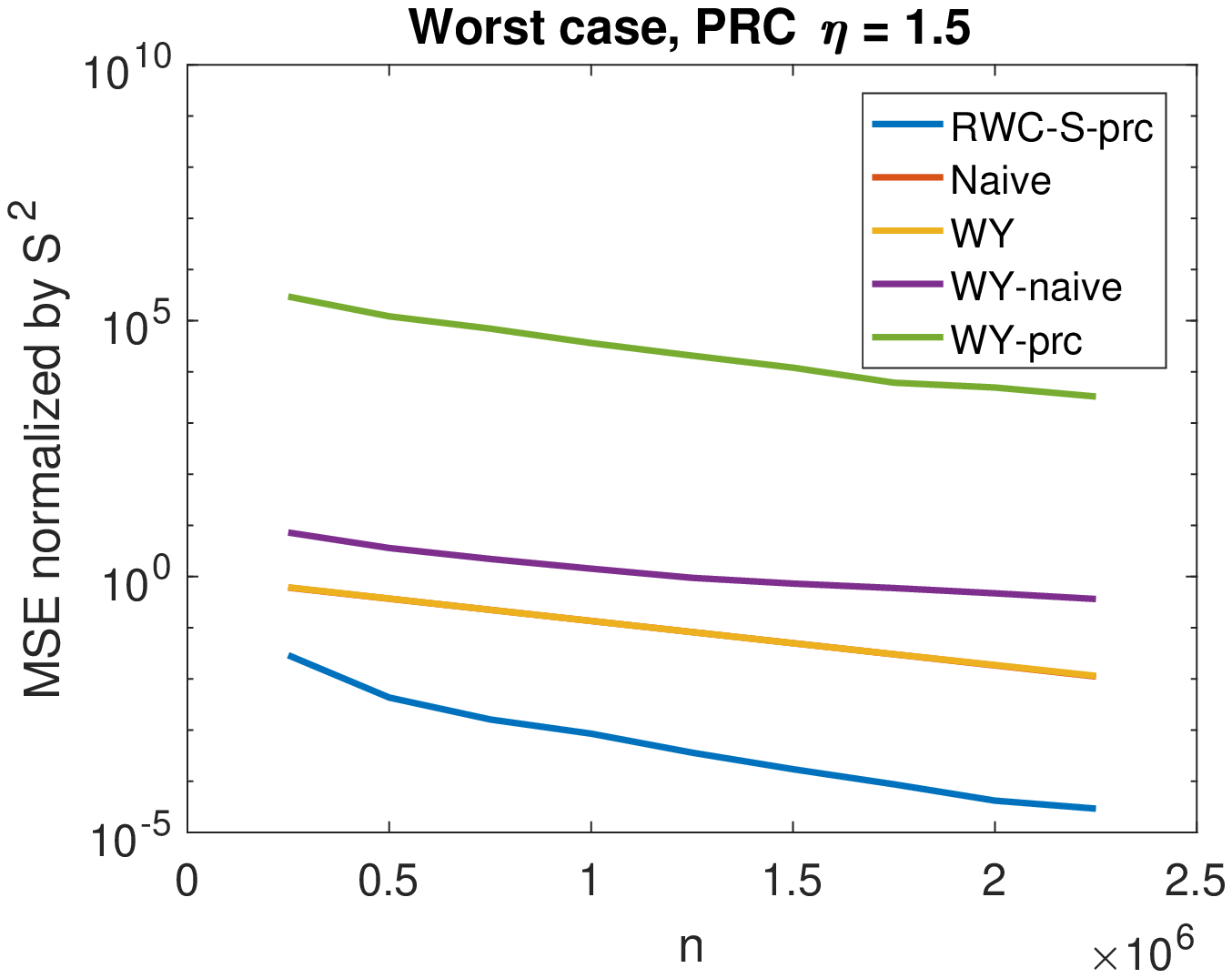}}\hspace{-1em}
  \subfigure[Hamlet MSE$/S^2$ \label{fig:Hamlet}]{\includegraphics[width=0.26\linewidth]{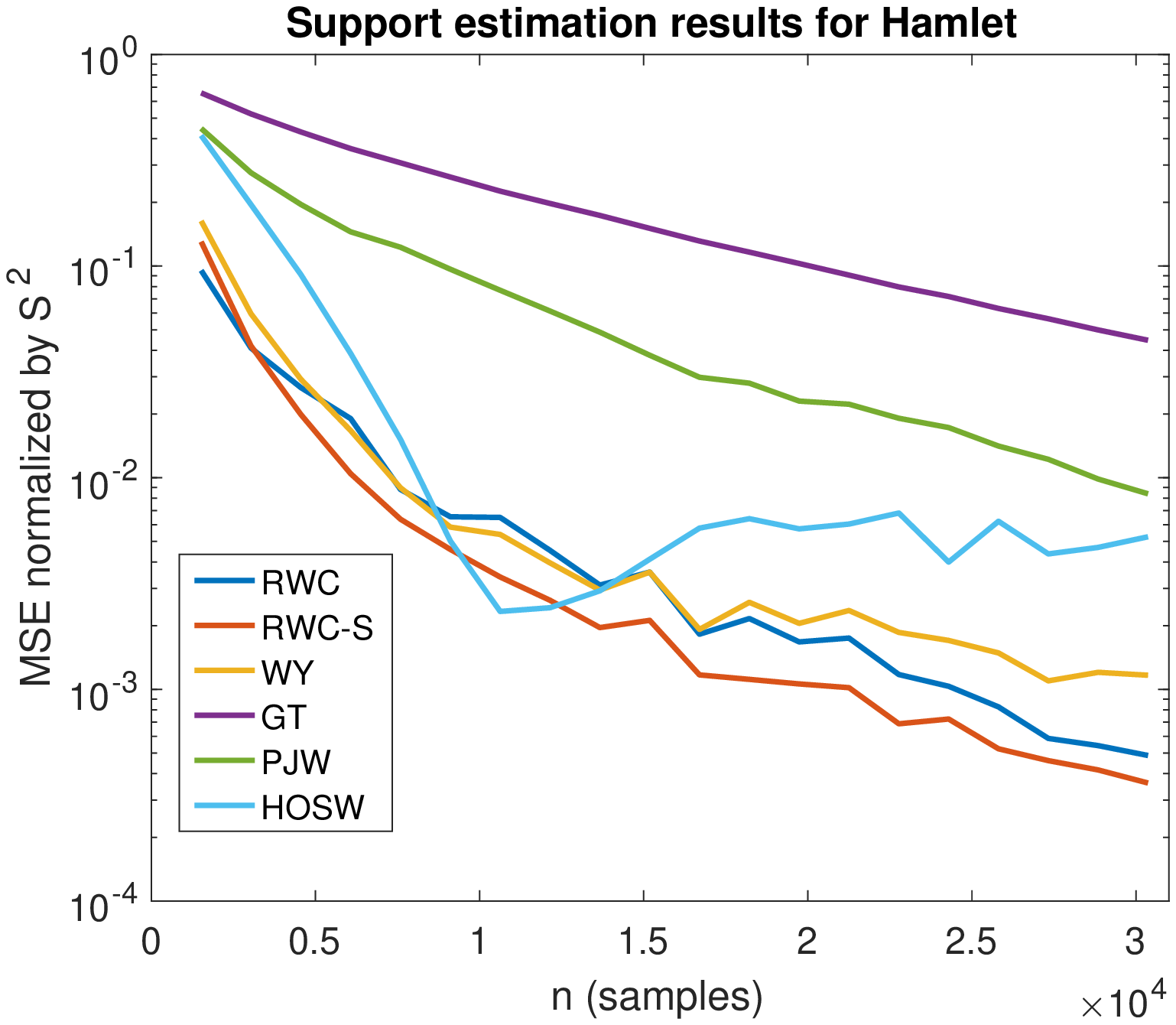}}
  \vspace{-0.2cm}
  \caption{ The $y$-axes of (a), (c), (e), (f), and (g) are in log scale. Figures (a)-(d) show demonstrate that our methods outperform state-of-the-art techniques in the noiseless setting. Figures (e)-(f) demonstrate that our method outperforms all other methods in the presence of Poisson repeats. Figure (h) illustrates the superiority of our methods on noiseless real-world data.
  }\label{fig:eachdist}
  \vspace{-0.4cm}
\end{figure}
\paragraph{Real-world data experiments.}
We start by estimating the number of distinct words in selected books, as suggested in~\cite{wu2019chebyshev,valiant2013estimating}. We use Hamlet, Othello, Macbeth and King Lear for our comparative study, with the results presented in Figure~\ref{fig:Hamlet} (for Hamlet) and the Supplement (all other plays). In the experiments, we randomly sampled words from the text with replacement and used the obtained counts to estimate the number of distinct words. For simplicity, we set $k$ to be equal the total number of words. For example, as the number of words in Hamlet equals $30,364,$ we set $k = 30,364$. As may be clearly seen, our methods significantly outperform all other competitive techniques both in terms of convergence rate and the accuracy of the estimated support.

To estimate the mutational support of the SARS-Cov-2 virus in the presence of sampling artifacts, we first create the histogram of mutations in sequenced genomes, using a reference corresponding to Patient 1 (the first infected individual that was sequenced). The mutational support of a population of individuals equals the size of the union of the individual supports. The datasets used in the study were retrieved from the GISAID repository~\cite{elbe2017data,shu2017gisaid} on 04-14-2020, and they pertain to European patients only. The analysis of datasets acquired from Asia and North America is relegated to the Supplement. We conduct three experiments: first, we examine the results of the noiseless support estimation methods (Table~\ref{table:noiseless}); next, we manually corrupt the samples by Poisson repeats with $\eta = 0.5,1,1.5$ (Table~\ref{table:Covid_artiPRC} and the Supplement for $\eta=1,1.5$). A good noisy support estimation method should produce a support close to that of its noiseless counterpart, and in this setting our method shows superior performance compared to other techniques. Finally, we also report the results of noisy support estimation on the unperturbed data (Table~\ref{table:noisy}). Note that the naive estimator gives a lower bound for the true support size, and the results of WY-Naive for $\eta = 0.5$ are erroneous. On the other hand, WY-prc produces estimates that violate the known maximum support size or negative entries for $\eta = 1.5$, as reported in the Supplement. A more detailed discussion of the relevant biological findings is also available in the Supplement.

\begin{table}[th!]
\begin{minipage}[t]{.48\linewidth}
\caption{Support estimation with synthetic Poisson repeats. We report the relative difference (in $\%$) of the results for the noisy and noiseless counterparts (the closer the value to $0,$ the better the performance).}\label{table:Covid_artiPRC}
\setlength{\tabcolsep}{2pt}
\centering
\tiny
\begin{tabular}{@{}ccccc@{}}
\toprule
Genomic region      & \multicolumn{4}{c}{$\eta$ = 0.5}                                          \\ \midrule
      & RWC-S-prc      & Naïve        & WY-Naïve & WY-prc         \\
ORF1a & \textbf{-30.5} & -50.6        & -116.5   & -64.7           \\
ORF1b & \textbf{-28.4} & -50.9        & -120     & -63.6            \\
S     & \textbf{-30.8} & -51.6        & -122.9   & -64.7           \\
ORF3a & \textbf{-32.5} & -49.5        & -96.1    & -70.6           \\
ORF6  & \textbf{-31.1} & -55.8        & -93.3    & -68.6           \\
ORF7a & \textbf{-33.6} & -54.5        & -90.1    & -68.7          \\
ORF8  & \textbf{-22.4} & -46.9        & -83      & -58.5   \\
N     & \textbf{-24}   & -45.3        & -121.9   & -56.6         \\
ORF10 & \textbf{7.7}   & -40          & -61.9    & 14.3             \\
All   & \textbf{-29.6} & -50.5        & -114.7   & -64             \\ \bottomrule
\end{tabular}
% \vspace{0.1in}

\end{minipage}\hfill
\begin{minipage}[t]{.48\linewidth}
\caption{Results of the noisy support estimation method directly applied to the real-world data. We underline the results that are obviously false (i.e., those violating the maximum support constraint, taking negative values or values smaller than the results of the naive estimator).  }\label{table:noisy}
\setlength{\tabcolsep}{2pt}
\centering
\tiny
\begin{tabular}{@{}cccccc@{}}
\toprule
Genomic region      & \multicolumn{4}{c}{$\eta$ = 0.5}                         & Maximum support \\ \midrule
      & RWC-S-prc & Naïve & WY-Naïve   & WY-prc &                 \\
ORF1a & 2346      & 911   & {\ul 465}  & 1262   & 13203           \\
ORF1b & 1106      & 477   & {\ul 230}  & 665    & 8087            \\
S     & 553       & 246   & {\ul 104}  & 353    & 3822            \\
ORF3a & 210       & 99    & {\ul 70}   & 112    & 828             \\
ORF6  & 106       & 52    & {\ul 42}   & 64     & 186             \\
ORF7a & 139       & 66    & {\ul 55}   & 80     & 366             \\
ORF8  & 56        & 32    & {\ul 28}   & 37     & 366             \\
N     & 267       & 139   & {\ul 74}   & 188    & 1260            \\
ORF10 & 46        & 30    & {\ul 28}   & 32     & 117             \\
All   & 4955      & 2118  & {\ul 1156} & 2867   & 29132           \\ \bottomrule
\end{tabular}
% \vspace{0.1in}

\end{minipage}\hfill
\end{table}

\bibliography{example_paper}
\bibliographystyle{IEEEtran}

\clearpage
\section*{Supplement}

\section{Proof of Theorem~\ref{lma:interval_sq}}

     To prove the result, we need to show that $\forall \lambda\geq 6.5L$, $\frac{\partial }{\partial \lambda}g(\mathbf{a},\lambda)<0$. The derivative of the first term in $g$ equals
    \begin{align*}
        & \frac{\partial}{\partial \lambda}\frac{1}{k}\bigg(\sum_{l=0}^{L}e^{-\lambda}a_l^2\lambda^ll!\bigg) = \frac{1}{k}\bigg(\sum_{l=0}^{L}(\frac{l}{\lambda}-1)e^{-\lambda}a_l^2\lambda^ll!\bigg).
    \end{align*}
    Clearly, the right hand side in the above expression is negative for all $\lambda>L$. The second term of the derivative equals
    \begin{align*}
        & \frac{\partial}{\partial \lambda}\bigg(e^{-\lambda}\sum_{l=0}^{L}a_l\lambda^l\bigg)^2 \\
        & = 2\bigg(e^{-\lambda}\sum_{l=0}^{L}a_l\lambda^l\bigg)\bigg(-e^{-\lambda}\sum_{l=0}^{L}a_l\lambda^l + e^{-\lambda}\sum_{l=0}^{L}
        \frac{l}{\lambda} a_l\lambda^l\bigg)\\
        & = 2e^{-2\lambda}\bigg(\sum_{l=0}^{L}a_l\lambda^l\bigg)\bigg(\sum_{l=0}^{L}
        (\frac{l}{\lambda}-1) a_l\lambda^l\bigg).
    \end{align*}
    To analyze the two terms of the derivative, we introduce the vectors $\mathbf{y},\mathbf{z},\mathbf{1}$ and the diagonal matrix $\mathbf{D}$ according to
    \begin{align*}
        &\mathbf{y} = (a_0\lambda^0,a_1\lambda^1,...,a_L\lambda^L)^T,\\
        &\mathbf{z} = ((\frac{0}{\lambda}-1),(\frac{1}{\lambda}-1),...,(\frac{L}{\lambda}-1))^T,\\
        &\mathbf{1} = (1,1,...,1)^T,\\
        & D_{ii} = (-1+\frac{i-1}{\lambda})\frac{(i-1)!}{\lambda^{(i-1)}}.
    \end{align*}
    Consequently, we have
    \begin{align*}
        & \frac{\partial}{\partial \lambda}\frac{1}{k}\bigg(\sum_{l=0}^{L}e^{-\lambda}a_l^2\lambda^ll!\bigg) = \frac{e^{-\lambda}}{k}\mathbf{y}^T\mathbf{D}\mathbf{y},\\
        & \frac{\partial}{\partial \lambda}\bigg(e^{-\lambda}\sum_{l=0}^{L}a_l\lambda^l\bigg)^2 = 2e^{-2\lambda}\mathbf{y}^T\mathbf{1}\mathbf{z}^T\mathbf{y} = e^{-2\lambda}\mathbf{y}^T(\mathbf{1}\mathbf{z}^T+\mathbf{z}\mathbf{1}^T)\mathbf{y}.
    \end{align*}
    Therefore,
    \begin{align*}
        \frac{\partial}{\partial \lambda}g(\mathbf{a},\lambda) = e^{-2\lambda}\mathbf{y}^T\left ( \frac{e^{\lambda}}{k}\mathbf{D} +(\mathbf{1}\mathbf{z}^T+\mathbf{z}\mathbf{1}^T)  \right ).\mathbf{y}
    \end{align*}
    To show that $\frac{\partial}{\partial \lambda}g(\mathbf{a},\lambda)<0$ for all polynomials of degree $L$ whenever $\lambda > CL$, we show that the matrix $\left ( \frac{e^{\lambda}}{k}\mathbf{D} +(\mathbf{1}\mathbf{z}^T+\mathbf{z}\mathbf{1}^T)  \right )$ is negative-definite whenever $\lambda > CL$, for some constant $C>0$. It suffices to show that the sum of the maximum eigenvalues of $\frac{e^{\lambda}}{k}\mathbf{D}$ and $(\mathbf{1}\mathbf{z}^T+\mathbf{z}\mathbf{1}^T)$ is negative, since $\frac{e^{\lambda}}{k}\mathbf{D}$ is a diagonal matrix. Thus, we turn our attention to determining the maximum eigenvalues of these two matrices. For $\frac{e^{\lambda}}{k}\mathbf{D}$, the maximum eigenvalue satisfies
    \begin{align*}
        \frac{e^{\lambda}}{k}\max_{i\in\{0,1,...,L\}} \left(-1+\frac{i}{\lambda}\right) \frac{i!}{\lambda^{i}} \leq -\frac{e^{\lambda}}{2k}\min_{i\in\{0,1,...,L\}}\frac{i!}{\lambda^{i}},
    \end{align*}
    since for $\lambda>2L$, one has $(-1+\frac{i}{\lambda})\leq -\frac{1}{2}$. When $\lambda>L$, it is clear that $\frac{i!}{\lambda^{i}}$ is decreasing in $i$, for $i\in\{0,1,...,L\}$, so that
    \begin{align*}
        \min_{i\in\{0,1,...,L\}}\frac{i!}{\lambda^{i}} = \frac{L!}{\lambda^L} \geq \left(\frac{L}{e\lambda}\right)^L.
    \end{align*}
    The last inequality is a consequence of Stirling's formula, which asserts that $n! \geq (\frac{n}{e})^n$. Combining the above expressions, we obtain
    \begin{align*}
        \frac{e^{\lambda}}{k}\max_{i\in\{0,1,...,L\}}\left(-1+\frac{i}{\lambda}\right)\frac{i!}{\lambda^{i}} \leq -\frac{e^{\lambda}}{2k}\left(\frac{L}{e\lambda}\right)^L.
    \end{align*}
    Next, we derive an upper bound on maximum eigenvalue of the second matrix. The $i,j$ entry of the matrix $(\mathbf{1}\mathbf{z}^T+\mathbf{z}\mathbf{1}^T)$ equals $\frac{i+j-2}{\lambda} - 2,$ and all these values are negative when $\lambda>L$. Moreover, it is clear that the matrix of interest has rank equal to $2$.
%    \begin{align*}
%        \mathbf{1}\mathbf{z}^T+\mathbf{z}\mathbf{1}^T =
%        \begin{bmatrix}
%            \frac{0}{\lambda}-2 & \frac{1}{\lambda}-2 & \cdots & \frac{L}{\lambda}-2\\
%            \frac{1}{\lambda}-2 & \frac{2}{\lambda}-2 & \cdots & \frac{L+1}{\lambda}-2\\
%            \vdots               & \ddots              & \cdots & \vdots\\
%            \frac{L}{\lambda}-2 & \cdots               & \cdots & \frac{2L}{\lambda}-2
%        \end{bmatrix}
%    \end{align*}
Therefore, the matrix has exactly two nonzero eigenvalues.

Let $\mathbf{A} = -(\mathbf{1}\mathbf{z}^T+\mathbf{z}\mathbf{1}^T)$. All entries of $\mathbf{A}$ are positive whenever $\lambda>L$. By Gershgorin's theorem, we can upper bound the maximum eigenvalues of the matrix $\mathbf{A}$ by its maximum row sum. It is obvious that the maximum row sum equals
    \begin{align*}
        2(L+1) - \frac{L(L+1)}{2\lambda}.
    \end{align*}
    Moreover, the trace of $\mathbf{A}$ equals
    \begin{align*}
        2(L+1) - \frac{L(L+1)}{\lambda}.
    \end{align*}
    This implies that the minimum eigenvalue of $\mathbf{A}$ is lower bounded by $- \frac{L(L+1)}{2\lambda},$ which directly implies that the maximum eigenvalue of $(\mathbf{1}\mathbf{z}^T+\mathbf{z}\mathbf{1}^T)$ is upper bounded by $\frac{L(L+1)}{2\lambda}$.

Summing up the two previously derived upper bounds gives
    \begin{align*}
        h(\lambda)\triangleq -\frac{e^{\lambda}}{2k}\left(\frac{L}{e\lambda}\right)^L+\frac{L(L+1)}{2\lambda},
    \end{align*}
    whenever $\lambda>2L$.
    Note that $h(\lambda)<0$ is equivalent to
    \begin{align}
        &\frac{L(L+1)}{2\lambda} < \frac{e^{\lambda}}{2k}\left(\frac{L}{e\lambda}\right)^L\nonumber\\
        & \Leftrightarrow \log(L) + \log(L+1) + \log(k) - L\log(L) + L < \lambda + \log(\lambda) - L\log(\lambda).\label{app:eq1}
    \end{align}
    The function $\lambda + \log(\lambda) - L\log(\lambda)$ is nondecreasing in $\lambda$ whenever $\lambda> L$ since
    \begin{align*}
        \frac{d}{d\lambda}(\lambda + \log(\lambda) - L\log(\lambda)) = 1-\frac{L-1}{\lambda}.
    \end{align*}
    By the definition of $L = \lfloor c_0\log(k) \rfloor$, we also have $\log(k)\leq \frac{L+1}{c_0}$. Using $\log(x+1)\leq x,$ which holds $\forall x\geq 1$. Hence $\forall \lambda > CL$ where $C>2$, the sufficient condition for \eqref{app:eq1} to hold is
    \begin{align*}
        &\log(L) + L + \frac{L+1}{c_0} - L\log(L) + L < CL + \log(CL) - L\log(CL).
        % &\Rightarrow \log(L) + \log(L+1) + \log(k) - L\log(L) + L\\
        % & < \lambda + \log(\lambda) - L\log(\lambda).
    \end{align*}
    Rearranging terms leads to
        \begin{align*}
        \left(C-\log(C) -2 -\frac{1}{c_0} \right)L + \log(C) > \frac{1}{c_0}.
    \end{align*}
    Sufficient conditions that ensure that the above inequality holds are $\log(C)\geq \frac{1}{c_0}$ and $(C-\log(C) -2 -\frac{1}{c_0})>0$. The first condition implies $C\geq e^{\frac{1}{c_0}} = 6.0021,$ while the second condition holds with $C = 6.5$, for which the first condition is also satisfied. This completes the proof.

\section{Proof of Theorem~\ref{thm:discretization}}

The proof consists of two parts. In the first part, we establish the conditions for convergence, while in the second part, we determine the convergence rate. For simplicity, we present the proofs for the case without Poisson repeats. We then outline how the analysis can be modified to account for the repeats.
\subsection{Proof of convergence}
We start by introducing the relevant terminology. Let $\Pi \subset \mathbb{R}^{L+1}$ be a closed set of parameters, and let $f$ be a continuous functional on $\Pi$. Assume that $B \subset \mathbb{R}$ is compact and that $g:\, \Pi\mapsto \mathcal{C}(B)$ is a continuous mapping from $\Pi$ into $\mathcal{C}(B),$ where $\mathcal{C}(B)$ is the space of continuous functions over $B$ equipped with the supremum norm $||\cdot||_{\infty}$. For each $D\subset B$ let
    \begin{align*}
        M(D) = \{\mathbf{c}\in \Pi|\,g(\mathbf{c},x)\leq 0, x\in D\}
    \end{align*}
    denote the set of feasible points of the optimization problem
    \begin{align*}
        \text{min }f(\mathbf{c})\text{ over }\mathbf{c}\in M(D).
    \end{align*}
    Assuming that $M(D)\neq \emptyset$, let
    \begin{align*}
        \mu(D) = \inf\{f(\mathbf{c})|\mathbf{c}\in M(D)\},
    \end{align*}
    and define the level set
    \begin{align*}
        \text{Level}(\mathbf{c}_0,D) = \{\mathbf{c}\in \Pi |\, f(\mathbf{c})\leq f(\mathbf{c}_0)\}\cap M(D).
    \end{align*}
We also make the following two assumptions:
\begin{assumption}[Fine grid]\label{assp:grid}
    Let $\mathbb{N}_0=\mathbb{N}\cup\{0\}$. There exists a sequence $\{B_i\}$ of compact subsets of $B$ with $B_{i}\subset B_{i+1},\, i\in\mathbb{N}_0,$ for which $\lim_{i\rightarrow \infty}h(B_{i},B) = 0,$ such that
    \begin{align*}
        h(B_i,B) = \sup_{x\in B}\inf_{y\in B_i}||x-y||.
    \end{align*}
    %and $||\cdot||_{\infty}$ is the maximum norm in $\mathbb{R}$
\end{assumption}
\begin{assumption}[Bounded level set]\label{assp:blevelset}
    $M(B)$ is nonempty, and there exists a $\mathbf{c}_0\in M(B)$ such that the level set $\text{Level}(\bold{c}_0,B_0)$ is bounded
    and hence compact in $\mathbb{R}^{L+1}$.
\end{assumption}
\begin{theorem}[Convergence of the discretized method, Theorem 2.1 in~\cite{reemtsen1991discretization}]\label{thm:convergence}
    Under assumptions~\ref{assp:grid} and~\ref{assp:blevelset}, the solution of the discretized problem converges to the optimal solution.
    More formally, we have
    \begin{align*}
        &\mu(B_i)\leq \mu(B_{i+1}) \leq \mu(B),\forall t\in\mathbb{N}_0\\
        &\lim_{i\rightarrow \infty} \mu(B_i) = \mu(B).
    \end{align*}
    If $\bold{c}^{\ast}$ is the unique optimal solution of the original problem, and $\bold{c}^{\ast}_{i}$ is the optimal solution of the discretized relaxation with grid $B_i$, then
    \begin{align*}
        \lim_{i \rightarrow \infty} ||\bold{c}^{\ast}-\bold{c}^{\ast}_{i}||_2 = 0.
    \end{align*}
\end{theorem}
%Thus if we can show that our original problem \eqref{sqlosseq2} satisfies assumptions \ref{assp:grid} and \ref{assp:blevelset}, then by theorem \ref{thm:convergence} we know that with $s$ large enough such the the grid is arbitrary fine, then our solution from discretization method will be arbitrary close to the original optimal solution.
It is straightforward to see that our chosen grid is arbitrary fine. Hence, we only need to prove that there exists a $\mathbf{c}_0$
such that the level set $\text{Level}(\mathbf{c}_0,D)$ is bounded.

Let $\mathbf{c} = (\mathbf{a};t)$ and note that in our setting, $f(\mathbf{c}) = t$. Rewrite $g(\mathbf{c},\lambda)$ in matrix form as
\begin{align*}
  %  & g(\mathbf{a},\lambda) \triangleq \bigg\{\frac{1}{k}\bigg(\sum_{l=0}^{L}e^{-\lambda}a_l^2\lambda^ll!\bigg)+\bigg(e^{-\lambda}P_L(\lambda,\mathbf{a})\bigg)^2\bigg\}\\
   % & \mathbf{M}(\lambda) \triangleq \frac{e^{-\lambda}}{k} Diag(\lambda^{0}0!,\lambda^{1}1!,...,\lambda^{L}L!)\\
   % & \Lambda \triangleq e^{-\lambda}(\lambda^{0},\lambda^{1},...,\lambda^{L})^T\\
     g(\mathbf{c},\lambda) = \mathbf{a}^T\mathbf{M}(\lambda)\mathbf{a}+\mathbf{a}^T\mathbf{\Lambda}\mathbf{\Lambda}^T\mathbf{a}-t,
\end{align*}
where
\begin{align*}
    \mathbf{\Lambda} \triangleq e^{-\lambda}(\lambda^{0},\lambda^{1},...,\lambda^{L})^T.
\end{align*}
Note that only $a_1,...a_L$ are allowed to vary since we fixed $a_0 = -1$. Obviously, $\mathbf{\Lambda}\mathbf{\Lambda}^T$ is positive semi-definite and the previously introduced $\mathbf{M}(\lambda)$ is positive definite for all
$\lambda>0$. Since the constraints on $g$ in~\eqref{sqlosseq5} are positive definite with respect to $a_1,...a_L,$ $g$ is coercive in $a_1,...a_L$. Furthermore, for any given $t$, the set of feasible coefficients $a_1,...a_L$ is bounded. Therefore, given a $t_0$, the level set $\text{Level}(\mathbf{c}_0,B_0)$ is bounded. This ensures that Assumption~\ref{assp:blevelset} holds for our optimization problem.

Next, we prove the uniqueness of the optimal solution $\mathbf{c}^\star$. Note that proving this result is
equivalent to proving the uniqueness of $\mathbf{a}^\star$. Hence, we once again refer to the original
minmax formulation of our problem,
\begin{equation}
    \inf_{\mathbf{a}:a_0=-1}\sup_{\lambda\in [\frac{n}{k},6.5L]}\mathbf{a}^T(\mathbf{M}(\lambda)+\mathbf{\Lambda}\mathbf{\Lambda}^T)\mathbf{a}\triangleq \inf_{\mathbf{a}:a_0=-1}\sup_{\lambda\in [\frac{n}{k},6.5L]} h_{\lambda}(\mathbf{a})
\end{equation}
Clearly, $\forall \lambda\in [\frac{n}{k},6.5L],$ the function $h_{\lambda}(\mathbf{a})$ is strictly convex since
$(\mathbf{M}(\lambda)+\mathbf{\Lambda}\mathbf{\Lambda}^T) \succ 0,\;\forall \lambda\in [\frac{n}{k},6.5L].$
Taking the supremum over $\lambda$ preserves strict convexity since $\forall \theta\in(0,1),$ one has
\begin{align*}
    &\sup_{\lambda\in [\frac{n}{k},6.5L]} h_{\lambda}(\theta\mathbf{x}+(1-\theta)\mathbf{y})\\
    &< \sup_{\lambda\in [\frac{n}{k},6.5L]} \theta h_{\lambda}(\mathbf{x})+(1-\theta)h_{\lambda}(\mathbf{y}) \\
    &\leq \sup_{\lambda\in [\frac{n}{k},6.5L]} \theta h_{\lambda}(\mathbf{x})+\sup_{\lambda'\in [\frac{n}{k},6.5L]}(1-\theta)h_{\lambda'}(\mathbf{y}).
\end{align*}
Hence $\sup_{\lambda\in [\frac{n}{k},6.5L]} h_{\lambda}(\mathbf{a})$ is strictly convex, which consequently implies the uniqueness of $\mathbf{a}^\star$ and hence $\mathbf{c}^\star$.

For the case of samples passed through a Poisson channel, it is not hard to see that the constraints are again strictly convex in $\mathbf{a}$, where one only need to replace $\mathbf{M(\lambda)},\mathbf{\Lambda}$ by
\begin{align*}
    & \frac{1}{k}e^{-\lambda(1-e^{-\eta})}\,Diag(0!\eta^0M_{N^\ast}^{(0)}(0),1!\eta^1M_{N^\ast}^{(1)}(0),...,L!\eta^LM_{N^\ast}^{(L)}(0))\\
    & e^{-\lambda(1-e^{-\eta})}(\eta^0M_{N^\ast}^{(0)}(0),\eta^1M_{N^\ast}^{(1)}(0),...,\eta^LM_{N^\ast}^{(L)}(0))^T
\end{align*}
respectively. 
Thus, a similar analysis is possible and the details are omitted. The proof above along with the previous observation prove the convergence result of Theorem~\ref{thm:discretization}.

\subsection{Proof for the convergence rate}
In what follows, and for reasons of simplicity, we omit the constraint $a_0 = -1$ in the SIP formulation. The described proof only requires small modifications to accommodate $a_0 = -1$.

Recall that we used $B_d$ to denote the grid with grid spacing $d$. In order to use the results in~\cite{still2001discretization}, we require the convergence assumption below.

\begin{assumption}\label{assump:rate1}
    Let $\bar{\mathbf{c}}$ be a local minimizer of an SIP. There exists a local solution $\mathbf{c}_d$ of the discretized SIP with grid $B_d$ such that
    \begin{align*}
        ||\mathbf{c}_d-\bar{\mathbf{c}}|| \rightarrow 0.
    \end{align*}
\end{assumption}
This assumption is satisfied for the SIP of interest as shown in the first part of the proof.

\begin{assumption}\label{assump:rate2}
The following hold true:
    \begin{itemize}
        \item There is a neighborhood $\bar{U}$ of $\bar{\mathbf{c}}$ such that the function $\frac{\partial^2}{\partial \lambda^2}g(\mathbf{c},\lambda)$ is continuous on $\bar{U}\times B$.
        \item The set $B$ is compact, nonempty and explicitly given as the solution set of a set of inequalities, $B = \{\lambda\in \mathbb{R}| v_i(\lambda)\leq 0,i\in I\},$ where $I$ is a finite index set and $v_i\in C^2(B)$.
        \item For any $\bar{\lambda}\in B$, the vectors $\frac{\partial}{\partial \lambda}v_i(\bar{\lambda}),i\in \{i\in I| v_i(\bar{\lambda}) = 0\}$ are linearly independent.
    \end{itemize}
\end{assumption}
Recall that our objective is of the form
\begin{align*}
  %  & g(\mathbf{a},\lambda) \triangleq \bigg\{\frac{1}{k}\bigg(\sum_{l=0}^{L}e^{-\lambda}a_l^2\lambda^ll!\bigg)+\bigg(e^{-\lambda}P_L(\lambda,\mathbf{a})\bigg)^2\bigg\}\\
   % & \mathbf{M}(\lambda) \triangleq \frac{e^{-\lambda}}{k} Diag(\lambda^{0}0!,\lambda^{1}1!,...,\lambda^{L}L!)\\
   % & \Lambda \triangleq e^{-\lambda}(\lambda^{0},\lambda^{1},...,\lambda^{L})^T\\
     g(\mathbf{c},\lambda) = \mathbf{a}^T\mathbf{M}(\lambda)\mathbf{a}+\mathbf{a}^T\mathbf{\Lambda}\mathbf{\Lambda}^T\mathbf{a}-t,
\end{align*}
where
\begin{align*}
    &\mathbf{\Lambda} \triangleq e^{-\lambda}(\lambda^{0},\lambda^{1},...,\lambda^{L})^T,\;\mathbf{c} = (\mathbf{a};t),\\
    &\mathbf{M}(\lambda) \triangleq \frac{e^{-\lambda}}{k} \,Diag(\lambda^{0}0!,\lambda^{1}1!,...,\lambda^{L}L!).
\end{align*}
It is straightforward to see that the first condition in Assumption~\ref{assump:rate2} holds. For the second condition, recall that $B = [\frac{n}{k}, 6.5L]$. Hence, the second condition can be satisfied by choosing $I = \{1\}$, $v_1(\lambda) = (\lambda-\frac{n}{k})(\lambda-6.5L)$. Since we only have one variable $v_1$, it is also easy to see that the third condition is met.
\begin{assumption}\label{assump:rate3}
    The set $B$ satisfies Assumption~\ref{assump:rate2} and all the sets $B_d$ contain the boundary points $\frac{n}{k}, 6.5L$.
\end{assumption}
This assumption also clearly holds for the grid of choice. Note that it is crucial to include the boundary points for the proof in~\cite{still2001discretization} to be applicable.
\begin{assumption}\label{assump:rate4}
    $\nabla_{\mathbf{c}} g(\mathbf{c},\lambda)$ is continuous on $\bar{U}\times B$, where $\bar{U}$ is a neighborhood of $\bar{\mathbf{c}}$. Moreover, there exists a vector $\xi$ such that
    \begin{align*}
        \nabla_{\mathbf{c}}g(\bar{\mathbf{c}},\lambda)^T\xi\leq -1,\;\forall \lambda\in B.
    \end{align*}
\end{assumption}
Note that $\nabla_{\mathbf{c}}g(\mathbf{c},\lambda) = [\nabla_{\mathbf{a}}g(\mathbf{c},\lambda);\nabla_{t}g(\mathbf{c},\lambda)]$ and
\begin{align*}
    \nabla_{\mathbf{a}}g(\mathbf{c},\lambda) = 2(\mathbf{M}(\lambda)+\mathbf{\Lambda}\mathbf{\Lambda}^T)\mathbf{a}.
\end{align*}
Also note that $\forall \lambda\in B$, $\mathbf{M}(\lambda)+\mathbf{\Lambda}\mathbf{\Lambda}^T$ is positive definite. Hence choosing $\xi$ to be colinear with and of the same direction as $[-\mathbf{a}^T \; 1]^T$, as well as of sufficiently large norm will allow us to satisfy the inequality
\begin{align*}
    \nabla_{\mathbf{c}}g(\bar{\mathbf{c}},\lambda)^T\xi\leq -1,\;\forall \lambda\in B.
\end{align*}
Hence, Assumption~\ref{assump:rate4} holds as well. The next results follow from the above assumptions and observations, and the results in~\cite{still2001discretization}.
\begin{lemma}[Corollary 1 in~\cite{still2001discretization}]\label{sup:discreterate1}
    Let $t_d$ be the optimal objective value of the discretized SIP used for support estimation with the grid $B_d,$ and let $t^\star$ be the optimal objective value for the original SIP. Since Assumptions~\ref{assump:rate1},\ref{assump:rate2},\ref{assump:rate3},\ref{assump:rate4} hold, then for some $c_3>0$ and $d$ sufficiently small, we have
    \begin{align*}
        0\leq t^\star-t_d \leq c_3d^2.
    \end{align*}
\end{lemma}
Consequently, $t_d\rightarrow t^\star$ with a convergence rate of $O(d^2)$.
\begin{lemma}[Theorem 2 in~\cite{still2001discretization}]
\label{sup:discreterate2}
    Assume that all assumptions in Lemma~\ref{sup:discreterate1} hold. If there exists a constant $c_4>0$ such that
    \begin{align*}
        t-\bar{t} \geq c_4||\mathbf{c}-\bar{\mathbf{c}}||,\;\forall \mathbf{c}\in M(B)\cap \bar{U},
    \end{align*}
    then for sufficiently small $d$ and $\sigma>0$ we have
    \begin{align*}
        ||\mathbf{c}_d-\bar{\mathbf{c}}||\leq \sigma d^2.
    \end{align*}
\end{lemma}
This result implies that if $\bar{\mathbf{c}}$ is also a strict minimum of order one, then the solution of the discretized SIP converges to that of the the original SIP with rate $O(d^2)$. For the Poisson repeat channel, the constraints are also strictly convex in $\mathbf{a}$. Therefore, a similar analysis is possible and the details are omitted once again. Combining these results completes the proof.

\section{Theoretical results supporting Remark~\ref{remark:pi_interval}}
The result described in the main text follows from Theorem 6.2 in \cite{lubinsky2007survey}, originally proved in~\cite{mhaskar1984extremal,mhaskar1985does} and~\cite{rakhmanov1984asymptotic}.
\begin{theorem}[Theorem 6.2 from~\cite{lubinsky2007survey}]\label{thm:MRS}
    Let $W(x) = \exp(-Q(x))$ be a weight function, where $Q:\mathbb{R} \mapsto [0,\infty) $ is even, convex, diverging for $x \to \infty$, and such that
    \begin{equation*}
        0 = Q(0) < Q(x), \forall x \neq 0.
    \end{equation*}
    %\textcolor{red}{(I find that we use $P$ for both polynomial and some probability distribution accidentally... Do we need to change one of them?)}
    Then, for any polynomial $P(x)$ of degree $\leq L$, not identical to zero, one has
    \begin{align*}
        &\sup_{x\in \mathbb{R}}|P(x)W(x)| = \sup_{x\in [-M_L,M_L]}|P(x)W(x)|,\\
        &\sup_{x\in \mathbb{R}\setminus [-M_L,M_L]}|P(x)W(x)| < \sup_{x\in [-M_L,M_L]}|P(x)W(x)|.
    \end{align*}
    Here, $M_L$ stands for the \textit{Mhaskar-Rakhmanov-Saff} (MSF) number, which is the smallest positive root of the integral equation
    \begin{equation}\label{app:MRSnumber}
        L = \frac{2}{\pi}\int_{0}^{1}\frac{M_LtQ'(M_Lt)}{\sqrt{1-t^2}}dt.
    \end{equation}
\end{theorem}
In our setting, the weight equals $\exp(-x)$. Solving~\eqref{app:MRSnumber} gives us an MSF number equal to $M_L = \frac{\pi}{2}L$. Thus, we can restrict our optimization interval to $[\frac{n}{k},\frac{\pi}{2}L +\frac{n}{k}]$. If there is no regularization term, the optimal interval reduces to $[\frac{n}{k},\frac{\pi}{2}L +\frac{n}{k}]$.
% Moreover, the Theorem 6.3, 6.4 in \cite{lubinsky2007survey} states that the sup-norm outside the \textit{Mhaskar-Rakhmanov-Saff} interval can be upper bound by an exponentially decaying term. In addition, from the proof of Lemma~\ref{lma:interval_sq} we know that the derivative of the regularization term is also negative when $\lambda>L$. This is how we come up with the interval $[\frac{n}{k},\frac{\pi}{2}L + \frac{n}{k}]$.

\section{Construction of the RWC-S estimator}
We introduce the optimization problem needed for minimizing the risk $\E \left( \frac{S-\hat{S}}{S}\right)^2$. Poissonization arguments once again establish that
\begin{align*}
    &\mathbb{E}\left(\frac{S-\hat{S}}{S}\right)^2 = \frac{1}{S^2}\bigg\{\sum_{i\in \mathcal{L}}\bigg(\sum_{l=0}^{L}e^{-\lambda_i}a_l^2\lambda_i^ll!\bigg)+\sum_{i\neq j\in \mathcal{L}}\bigg(e^{-\lambda_i}\sum_{l=0}^{L}a_l\lambda_i^l\bigg)\bigg(e^{-\lambda_j}\sum_{l=0}^{L}a_l\lambda_j^l\bigg)\bigg\}.
\end{align*}
Taking the supremum over $D_k$, one can further upper bound the risk as
\begin{align}
    %&\sup_{P\in D_k}\mathbb{E}(\frac{S-\hat{S}}{k})^2 \\
    &\leq \sup_{\lambda_{\ell}\in [\frac{n}{k}, n],\;\ell \in \mathcal{L}} \frac{1}{S^2}\bigg\{\sum_{i\in \mathcal{L}}\bigg(\sum_{l=0}^{L}e^{-\lambda_i}a_l^2\lambda_i^ll!\bigg)+\sum_{i\neq j\in \mathcal{L}}\bigg(e^{-\lambda_i}\sum_{l=0}^{L}a_l\lambda_i^l\bigg)\bigg(e^{-\lambda_j}\sum_{l=0}^{L}a_l\lambda_j^l\bigg)\bigg\}\nonumber\\
    & \leq \sup_{\lambda\in [\frac{n}{k}, n]}\bigg\{\frac{1}{S}\bigg(\sum_{l=0}^{L}e^{-\lambda}a_l^2\lambda^ll!\bigg)+\bigg(e^{-\lambda}\sum_{l=0}^{L}a_l\lambda^l\bigg)^2\bigg\}\nonumber\\
    &\leq \sup_{\lambda\in [\frac{n}{k}, n]}\bigg\{\frac{1}{\hat{S}_{c}}\bigg(\sum_{l=0}^{L}e^{-\lambda}a_l^2\lambda^ll!\bigg)+\bigg(e^{-\lambda}\sum_{l=0}^{L}a_l\lambda^l\bigg)^2\bigg\} \label{ratiolosseq3},
\end{align}
where the last inequality is due to the fact that $\hat{S}_{c}\leq S$. Note that the only difference between~\eqref{ratiolosseq3} and~\eqref{PRC:obj2}
is in terms of changing $1/k$ to $1/\hat{S}_c$ in the first term. In view of Theorem~\ref{lma:interval_sq}, \eqref{ratiolosseq3} is optimized by the solution of the following problem:
%After the similar analysis we can show that the following SIP will optimize the upper bound of squared ratio loss
%\begin{equation}\label{ratiolosseq1}
%    \begin{split}
 %       &\min_{t,a_1,...,a_L} t \;\;\;subject\;to\\
%        & \bigg\{\frac{1}{S}\bigg(\sum_{l=0}^{L}e^{-\lambda}a_l^2\lambda^ll!\bigg)+\bigg(e^{-\lambda}P_L(\lambda,\mathbf{a})\bigg)^2\bigg\}\leq t\\
%        & \forall \lambda\in Grid([\frac{n}{k}, \frac{\pi}{2}L+\frac{n}{k}],s), \text{with }a_0=-1.
%    \end{split}
%\end{equation}
%Note that since $S$ is unknown we need to replace it by some data-driven value. Interestingly, we know that $\hat{S}_{seen}\leq S$ and use $\hat{S}_{seen}$ will form an upper bound in the primal form for $\E (\frac{S-\hat{S}}{S})^2$. Thus we suggest to solve the following optimization problem for the squared ratio loss.

\begin{equation}\label{ratiolosseq2}
    \begin{split}
        &\min_{t,\mathbf{a}\in Poly(L)} t \;\;\;\text{ s.t.}\\
        & \bigg\{\frac{1}{\hat{S}_{c}}\bigg(\sum_{l=0}^{L}e^{-\lambda}a_l^2\lambda^ll!\bigg)+\bigg(e^{-\lambda}\sum_{l=0}^{L}a_l\lambda^l\bigg)^2\bigg\}\leq t,\; \forall \lambda\in \text{Grid}([\frac{n}{k}, 6.5L],s).
    \end{split}
\end{equation}

\section{Detailed description of the SARS-Cov-2 genomic data}
\begin{figure}[ht!]
		\centering
		\includegraphics[width=5.5in]{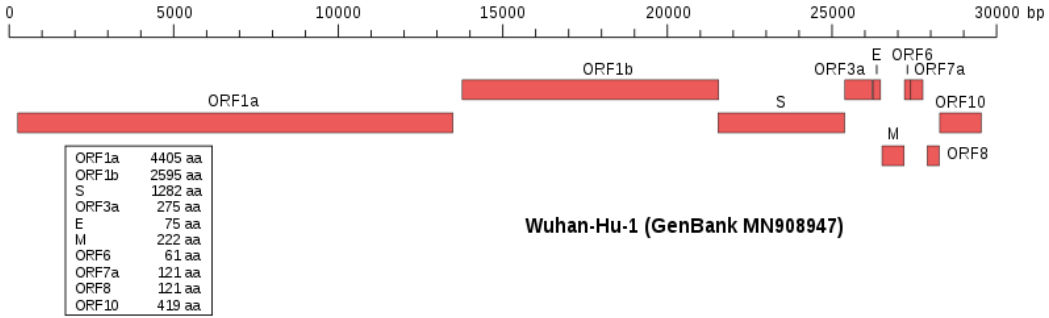}
		\caption{Organization of the SARS-Cov-2 genome (Source: Wikipedia). }
	\label{fig:genome}
	\vspace{-0.12in}
\end{figure}

\paragraph{Organization of the SARS-Cov-2 genome.} A breakdown of the genomic structure of SARS-Cov-2 is shown in Figure~\ref{fig:genome}, and described in detail in more detail in~\cite{mousavizadeh2020genotype}. SARS-Cov-2 comprises the following open reading frames: ORF1a and ORF1b, spike (S), membrane (M), envelope (E), and nucleocapsid (N), as well as ORF 10. Since all these ORFs encode proteins that have different roles in the process of evading the immune system of the host, we perform the mutational support analysis in the presence of sampling artifacts for each individual region. Note that the overall support is the sum of the mutational supports of all ORF regions.

\paragraph{Data acquisition. } We used data from the GISAID EpiCov repository~\cite{shu2017gisaid} which contains sequenced viral strains collected from patients across the world. We downloaded the data aggregated by the date of 04-14-2020. We filtered out all incomplete genomic datasets, resulting in a total of $8,893$ samples. To obtain the mutation counts, we first used the alignment software MUSCLE~\cite{edgar2004muscle} with respect to the reference of Patient 1, published under the name Wuhan-Hu-1 and collected at the Central Hospital of Wuhan, December 2019 (GenBank accession number MN909847). For each aligned pair of samples we generate a list containing the positions in the reference genome in which the patient aligned to the reference has a mutation. The mutation profile lists are aggregated producing a mutation histogram for each of the viral genome regions depicted in Figure~\ref{fig:genome}. To counter alignment artifacts caused by sequencing errors, we removed all gaps encountered in the prefixes and suffixes and sufficiently long gaps ($>$ 10 nts) within the alignments.

For the experiments, we partitioned the mutation histograms based on geographic location (Asia/ North America/ Europe). Since the number of samples across the three regions varies significantly, we subsampled the sequence sets to arrive at $636$ samples from Asia and $1774$ samples from Europe and North America.

\paragraph{Interpretation of the results. } The mutational supports without Poisson repeats may be used to estimate how quickly SARS-Cov-2 as well as any other virus mutates early on in the infection. The Poisson repeat model, as previously mentioned, accounts for resampling of patients that may have tested positive in a previous round or that may have been exposed to new sources of infection. Erroneous samples are accounted for through deletions. The mean value of the Poisson repeats is chosen to accommodate the fact that most individuals are sequenced once or not at all, but that certain high-risk groups (such as health workers) may be subject to multiple tests.

\paragraph{Remark} The estimators are based on the assumption that the symbol counts are independent and identically distributed, which may clearly not be the case when analyzing viral mutations (i.e., mutations at difference locations in the genomes may and are expected to be correlated). However, it gives good performance even if the i.i.d. assumption is violated. See experiments on Shakespeare's plays where the i.i.d. assumption does not hold for example.

\section{Additional experimental results}

\begin{figure}[!htb]
  \centering
%   \subfigure[Weighted vs classical Chebyshev approximation. \label{fig:BiasforexpW}]{\includegraphics[width=0.245\linewidth]{BiasforexpW.eps}}
%   \subfigure[The coefficients of $g_L$. \label{fig:gLcompare}]{\includegraphics[width=0.245\linewidth]{g_L.eps}}
  \subfigure[Uniform distribution.]{\includegraphics[width=0.32\linewidth]{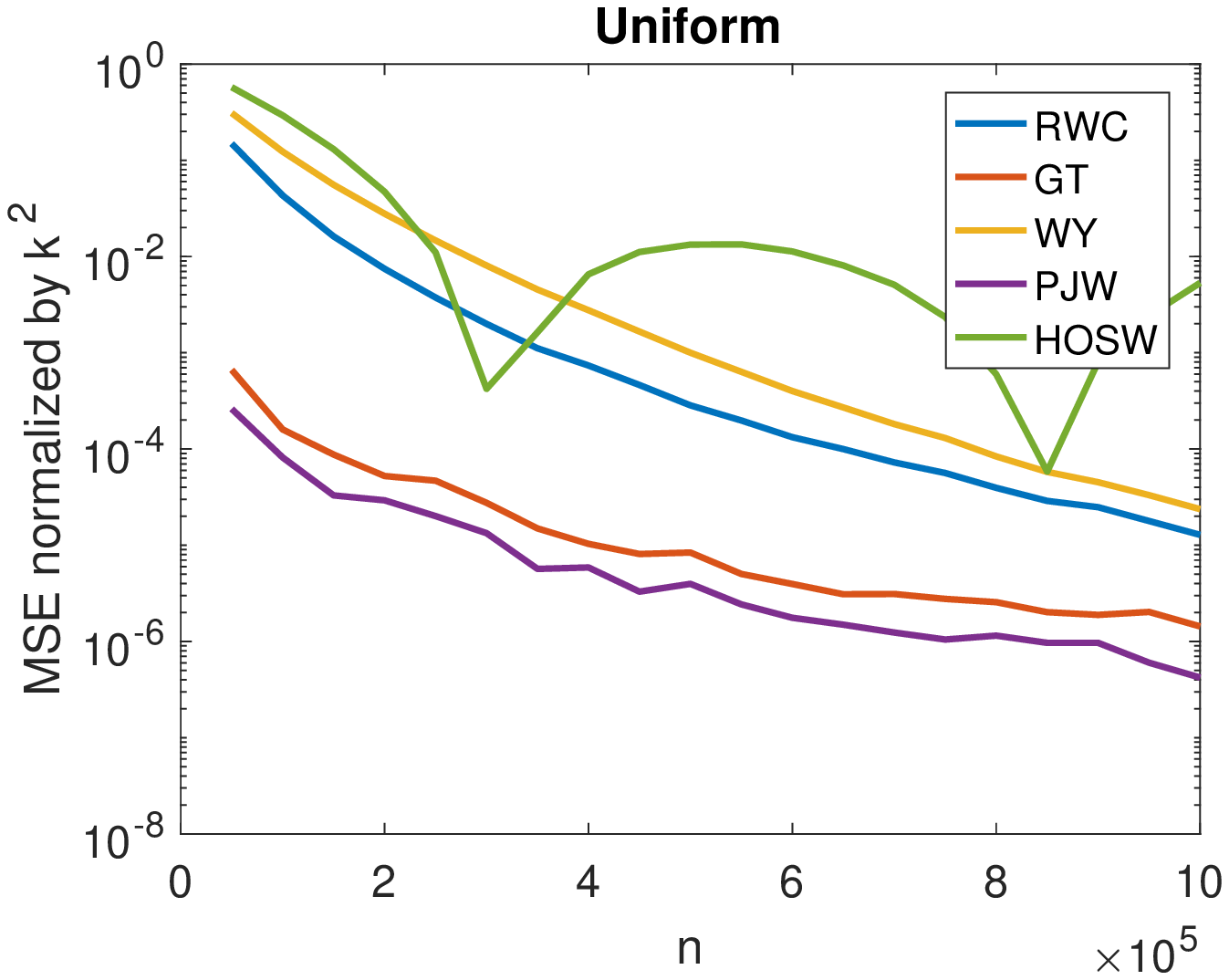}}
  \subfigure[Benford distribution.]{\includegraphics[width=0.32\linewidth]{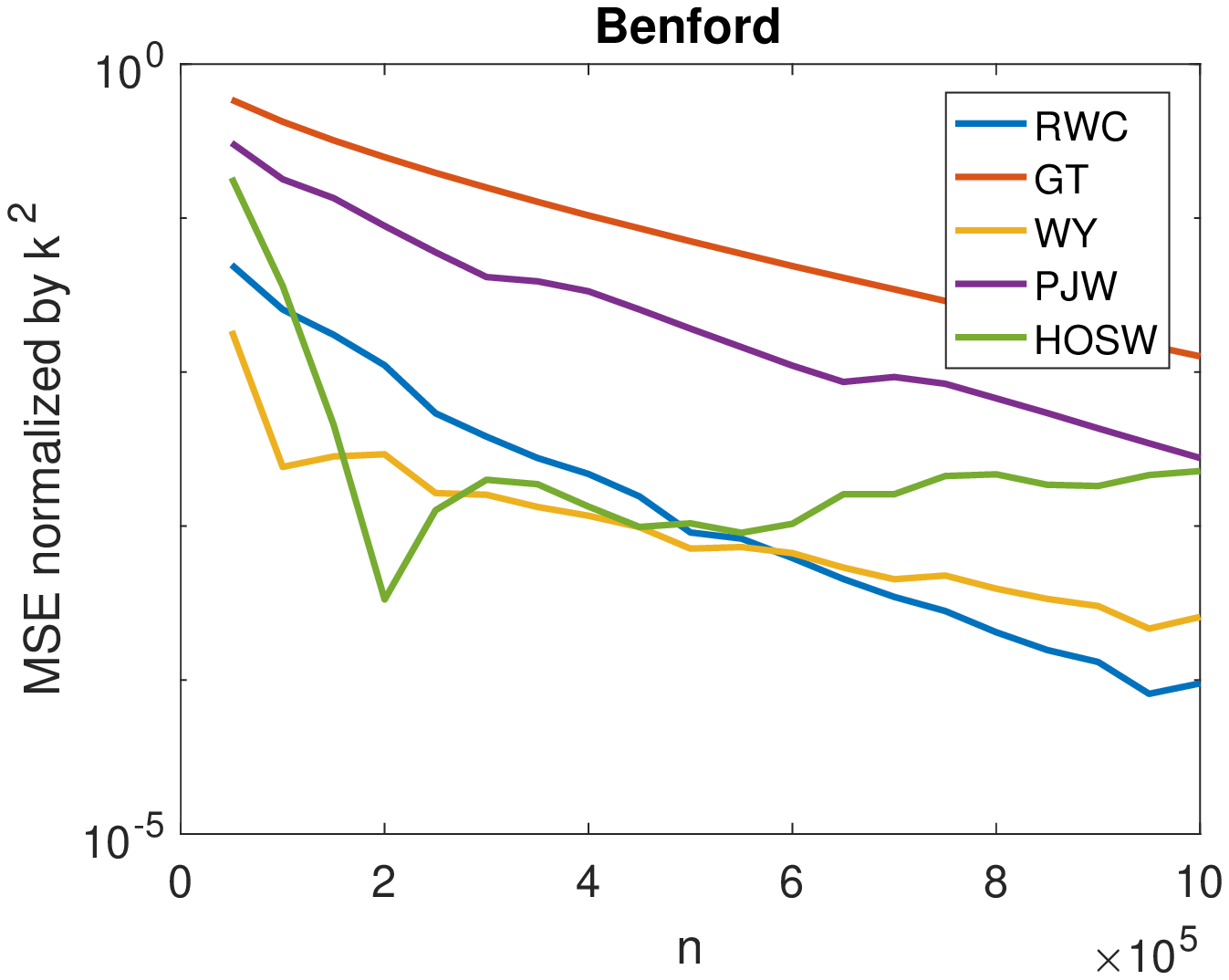}}
  \subfigure[Zipf$(1.5)$ distribution.]{\includegraphics[width=0.32\linewidth]{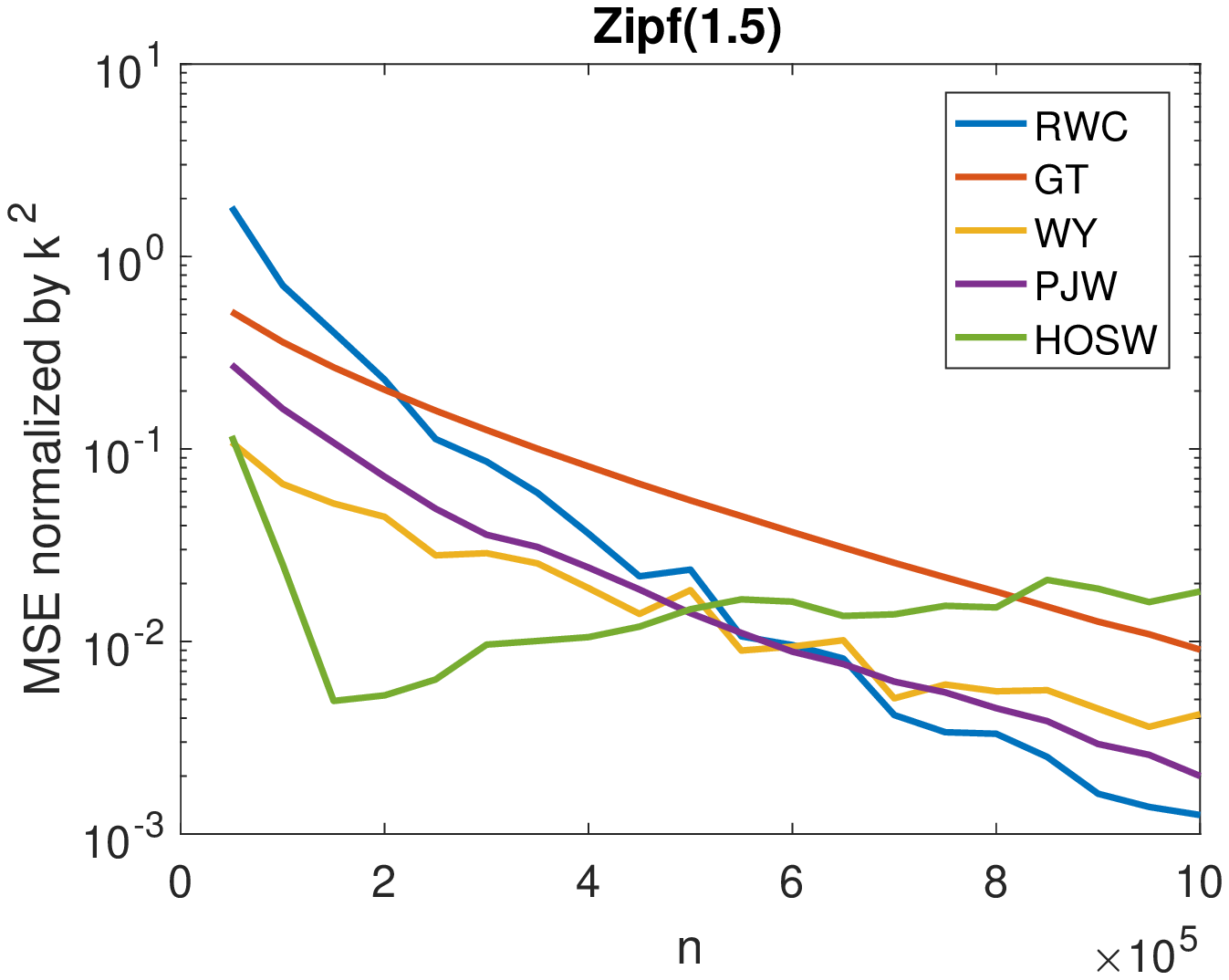}}
  \subfigure[Zipf$(1)$ distribution.]{\includegraphics[width=0.32\linewidth]{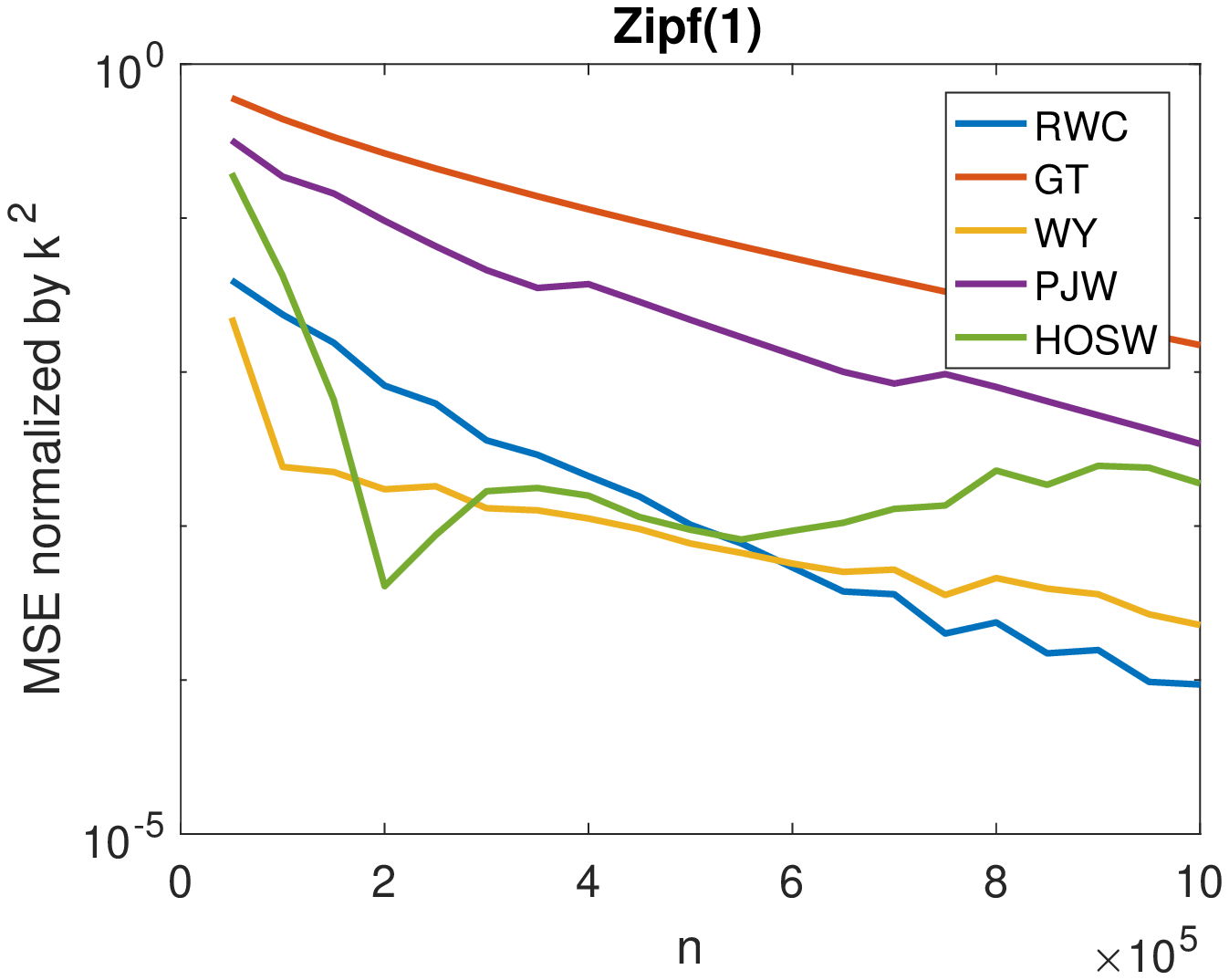}}
  \subfigure[Zipf$(0.5)$ distribution.]{\includegraphics[width=0.32\linewidth]{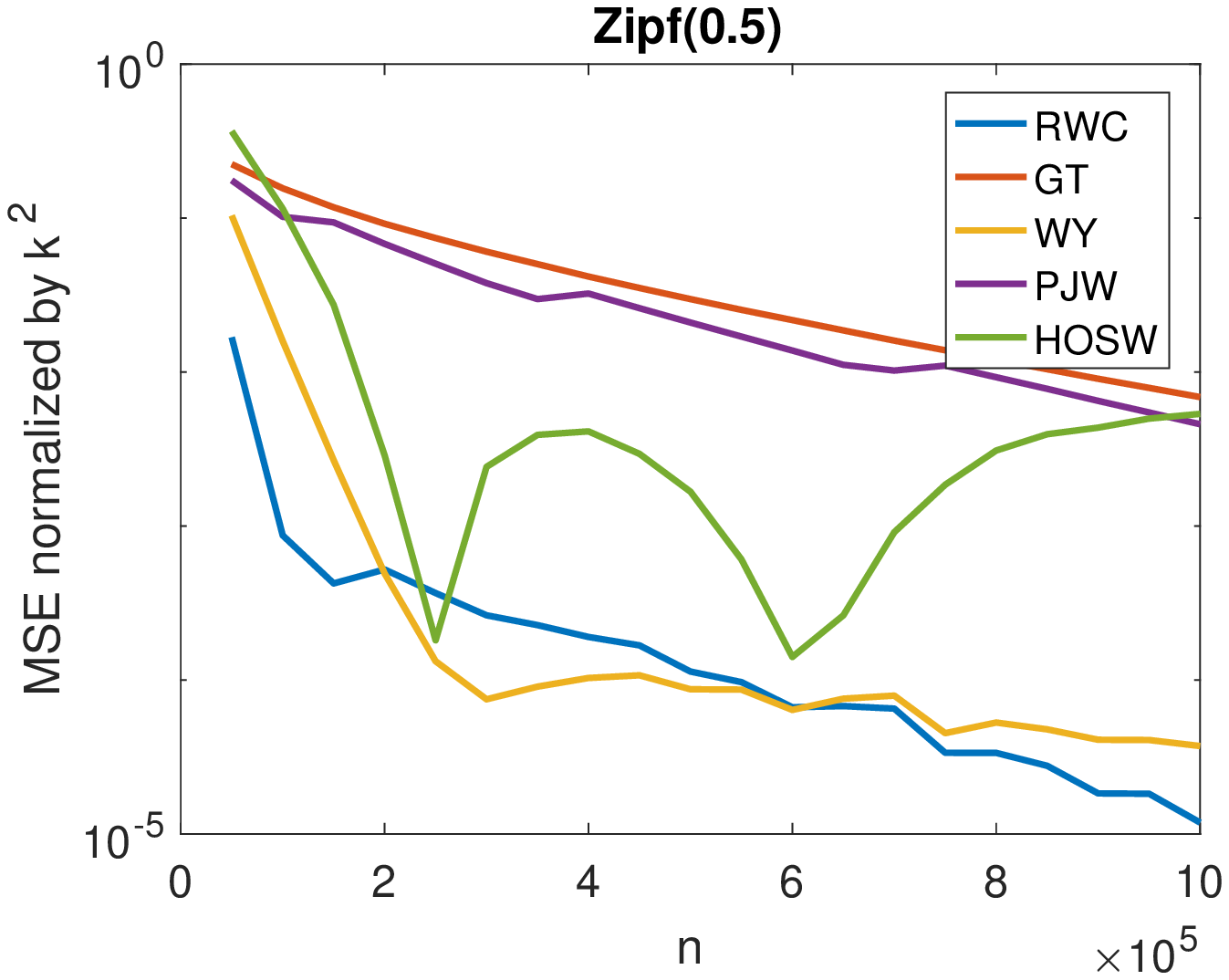}}
  \subfigure[Zipf$(0.25)$ distribution.]{\includegraphics[width=0.32\linewidth]{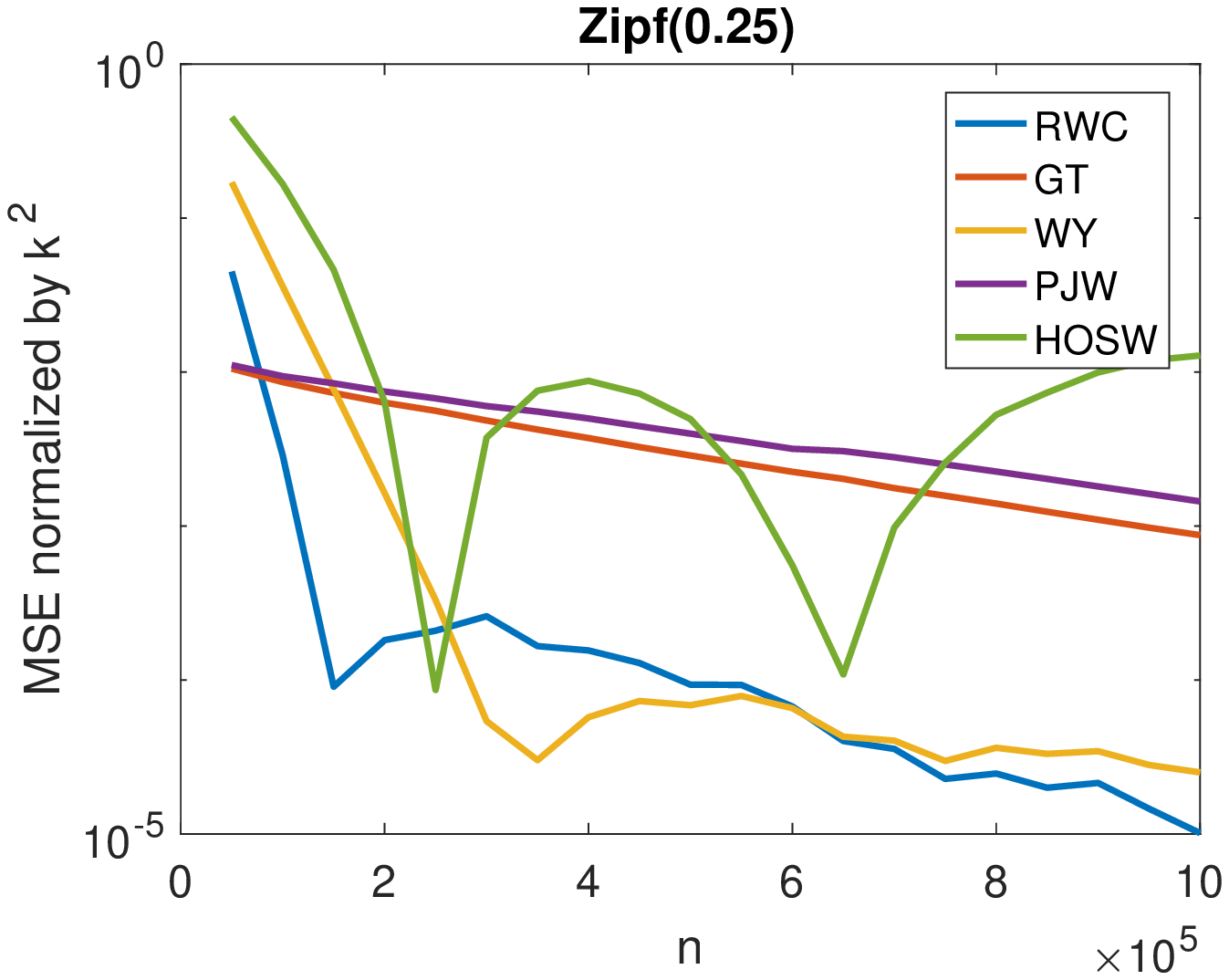}}
  \subfigure[Uniform distribution.]{\includegraphics[width=0.32\linewidth]{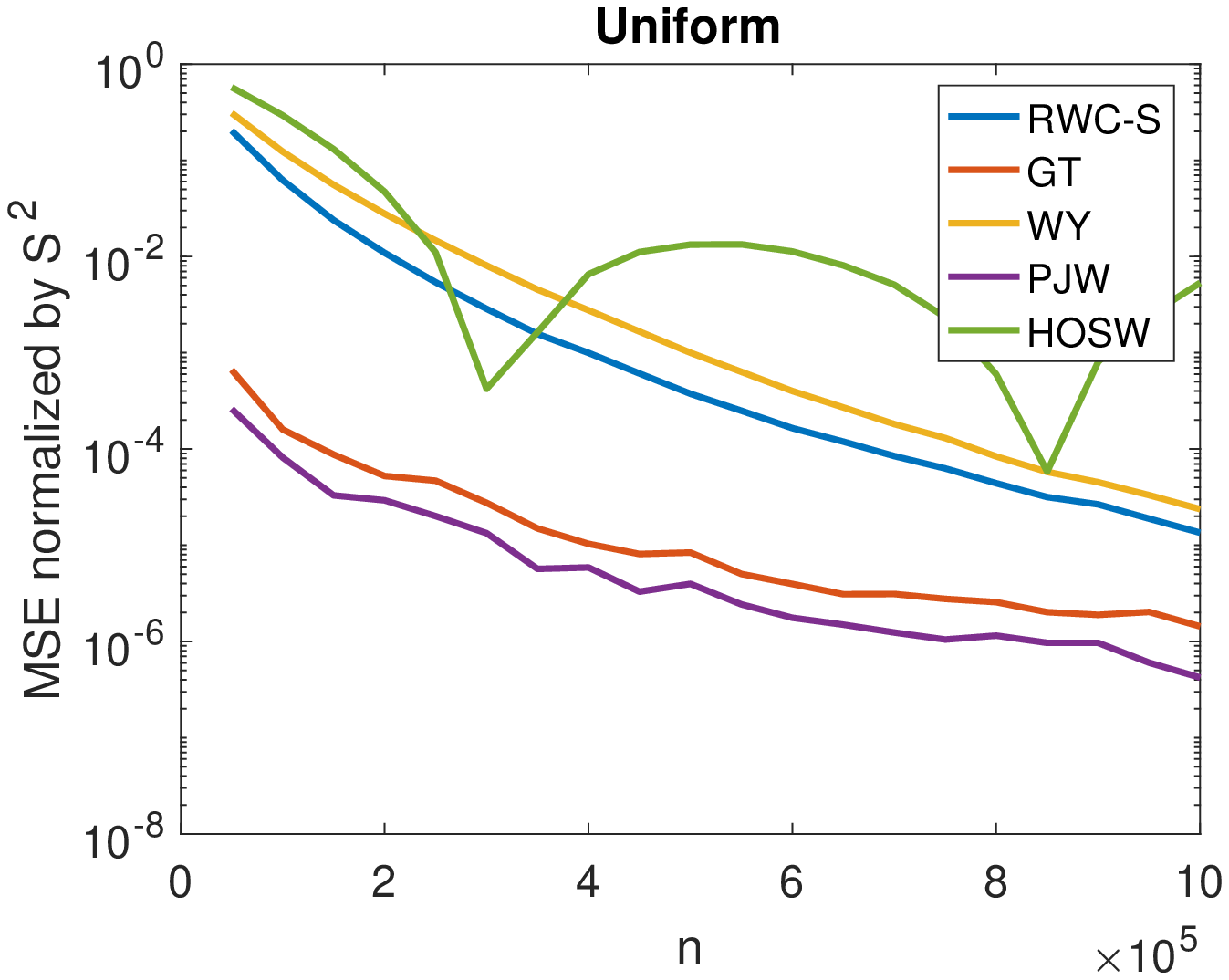}}
  \subfigure[Benford distribution.]{\includegraphics[width=0.32\linewidth]{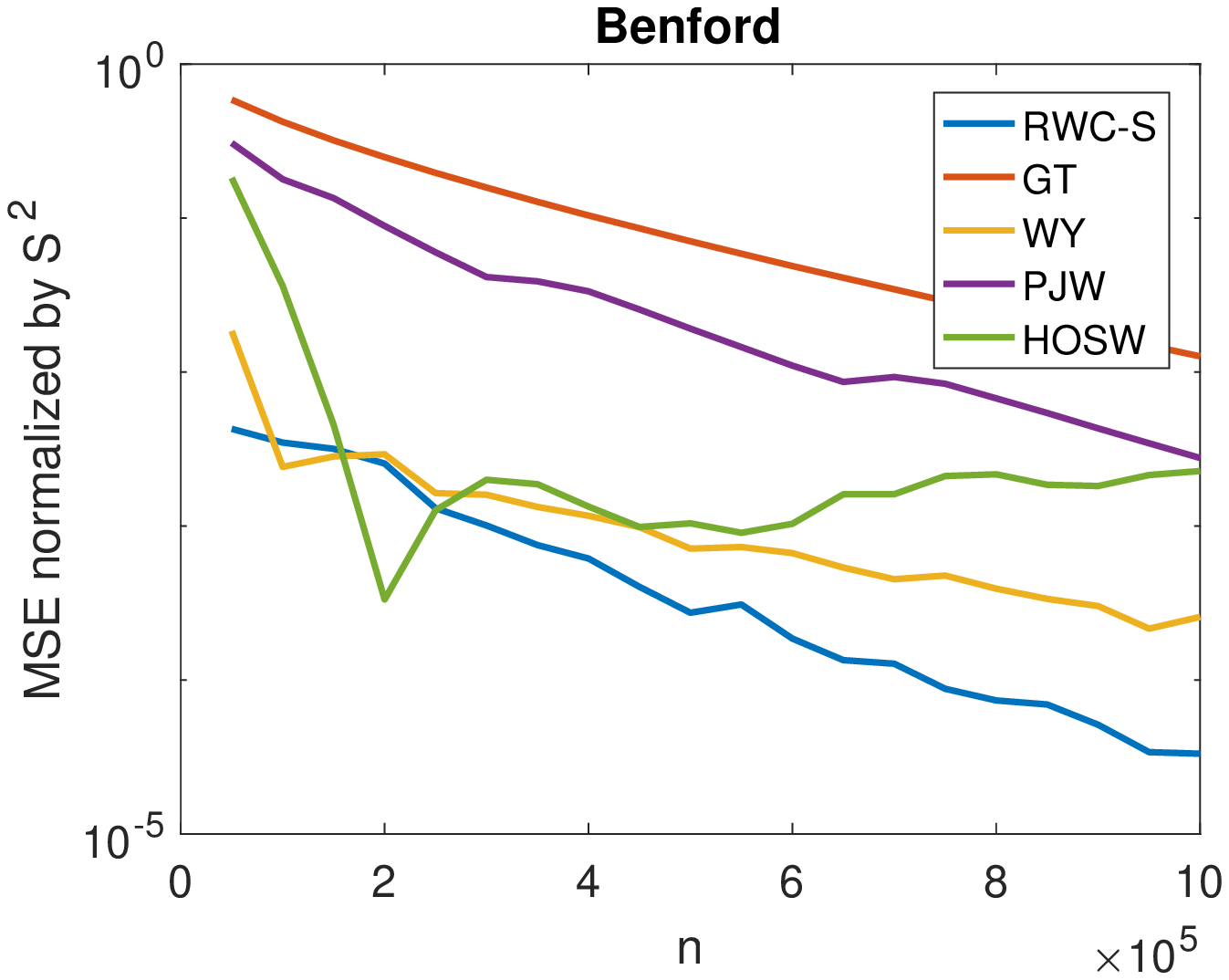}}
  \subfigure[Zipf$(1.5)$ distribution.]{\includegraphics[width=0.32\linewidth]{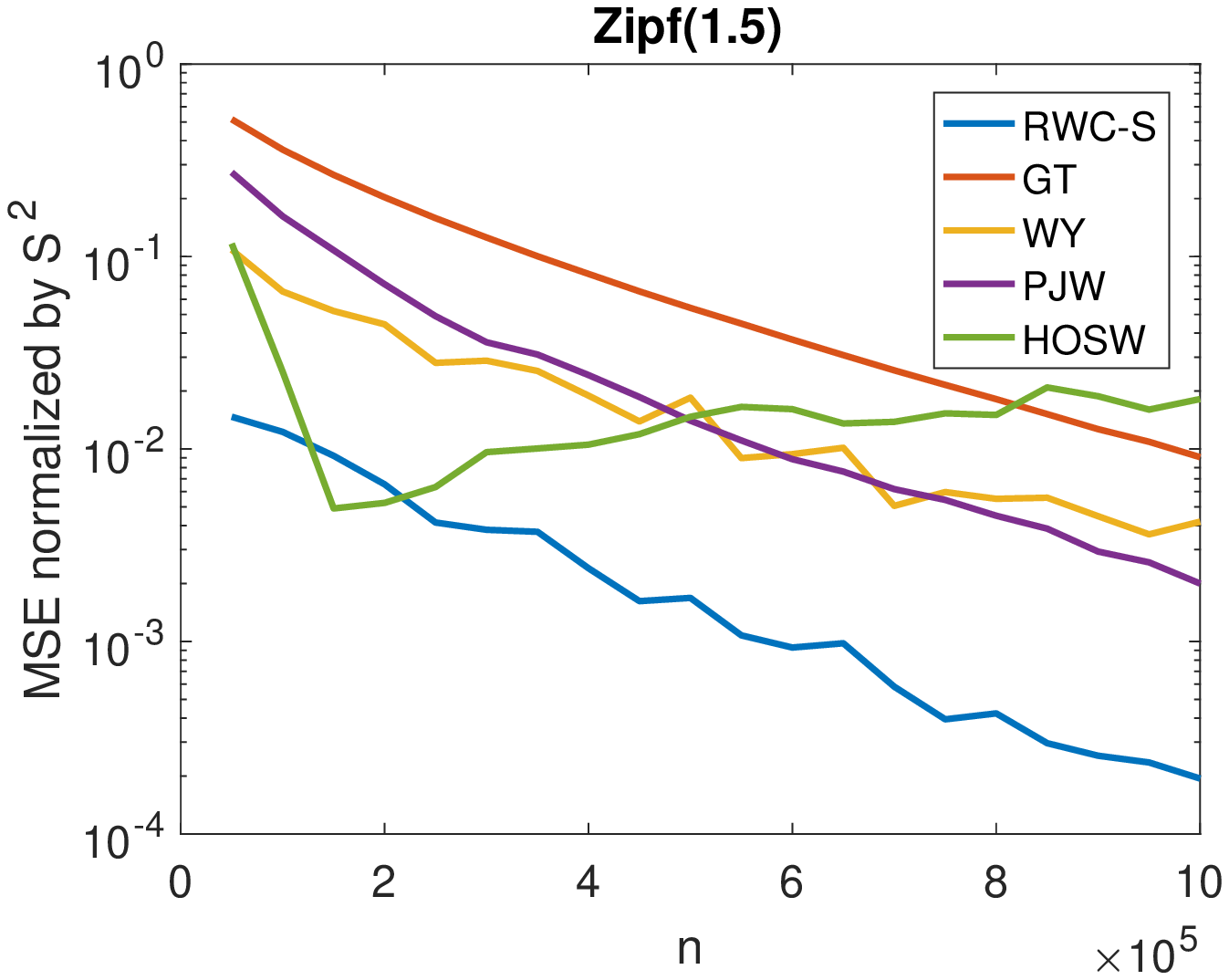}}
  \subfigure[Zipf$(1)$ distribution.]{\includegraphics[width=0.32\linewidth]{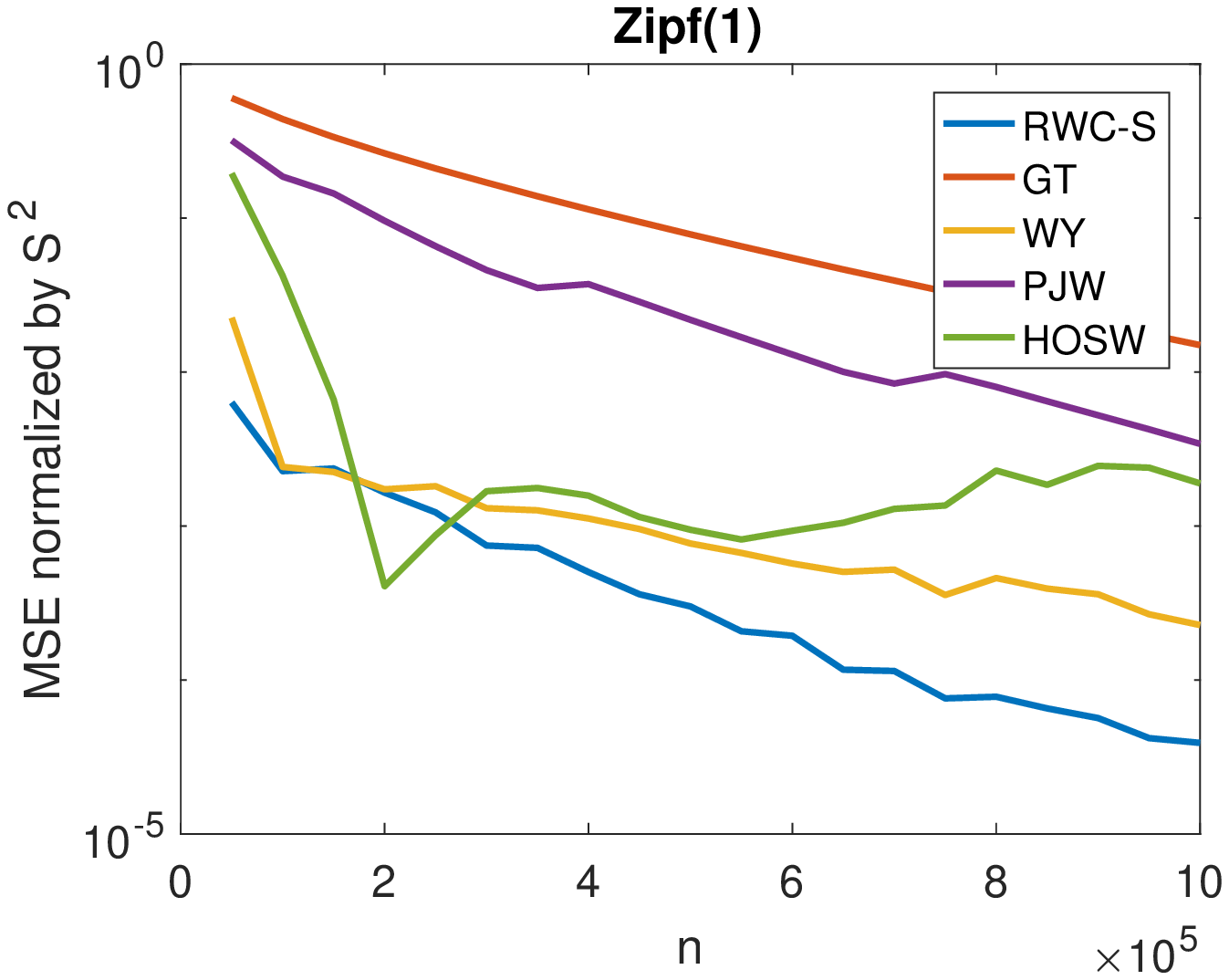}}
  \subfigure[Zipf$(0.5)$ distribution.]{\includegraphics[width=0.32\linewidth]{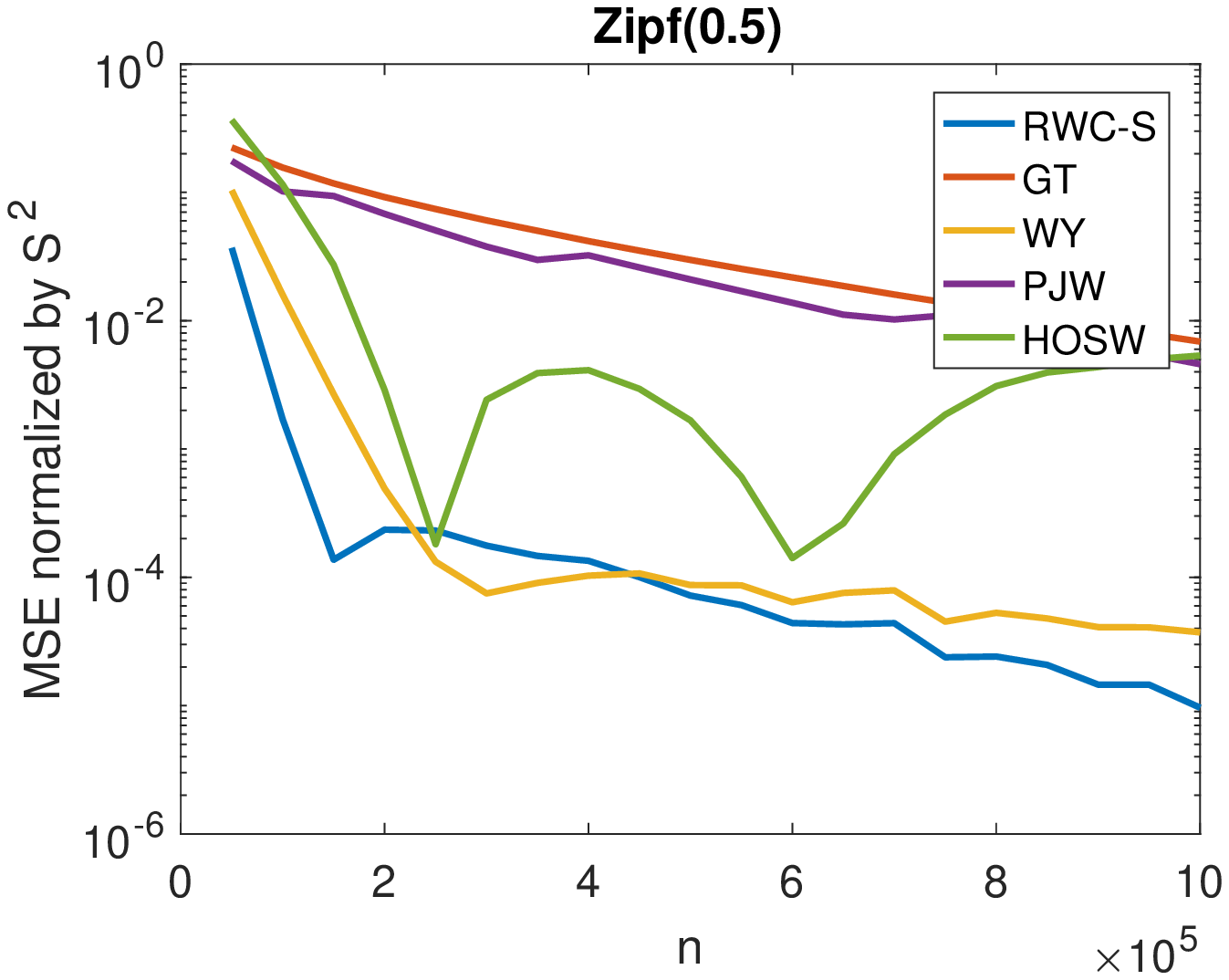}}
  \subfigure[Zipf$(0.25)$ distribution.]{\includegraphics[width=0.32\linewidth]{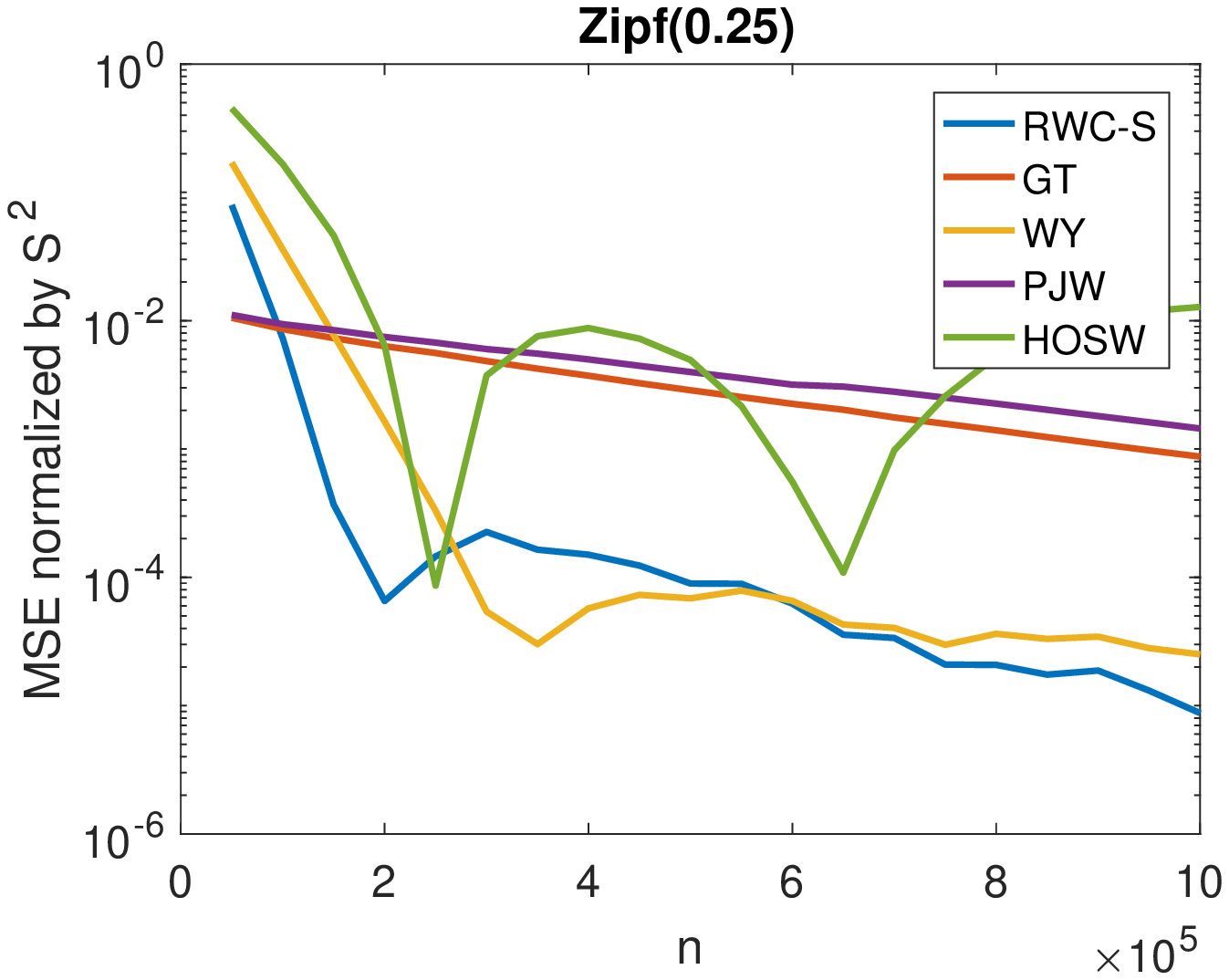}}
  \vspace{-0.2cm}
  \caption{ The figures plot the MSE of all estimators considered on each tested distributions without Poisson repeats. The $y$-axis is on the log scale.}
\end{figure}

\begin{figure}
    \centering
    \subfigure[Uniform distribution.]{\includegraphics[width=0.32\linewidth]{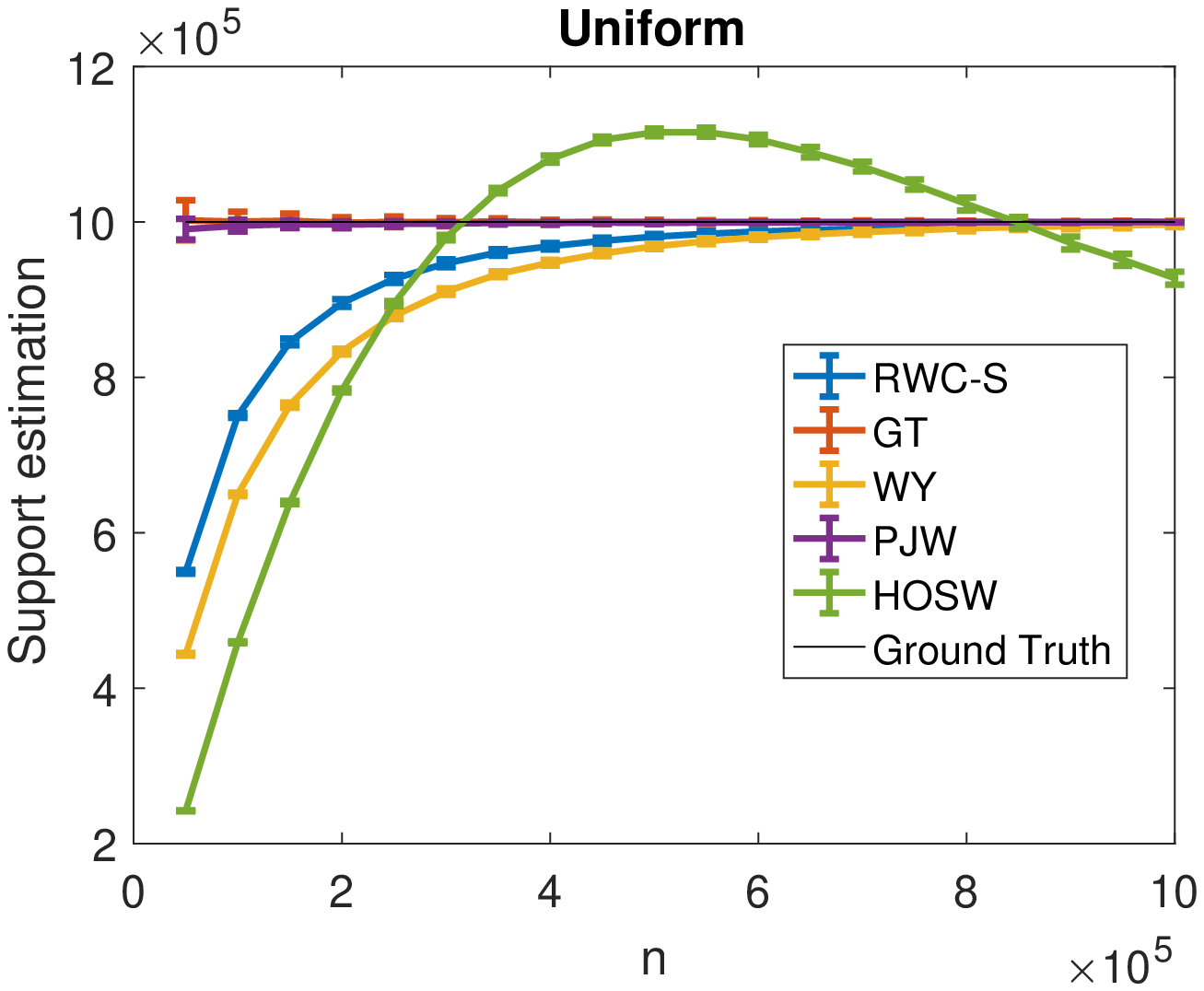}}
  \subfigure[Benford distribution.]{\includegraphics[width=0.32\linewidth]{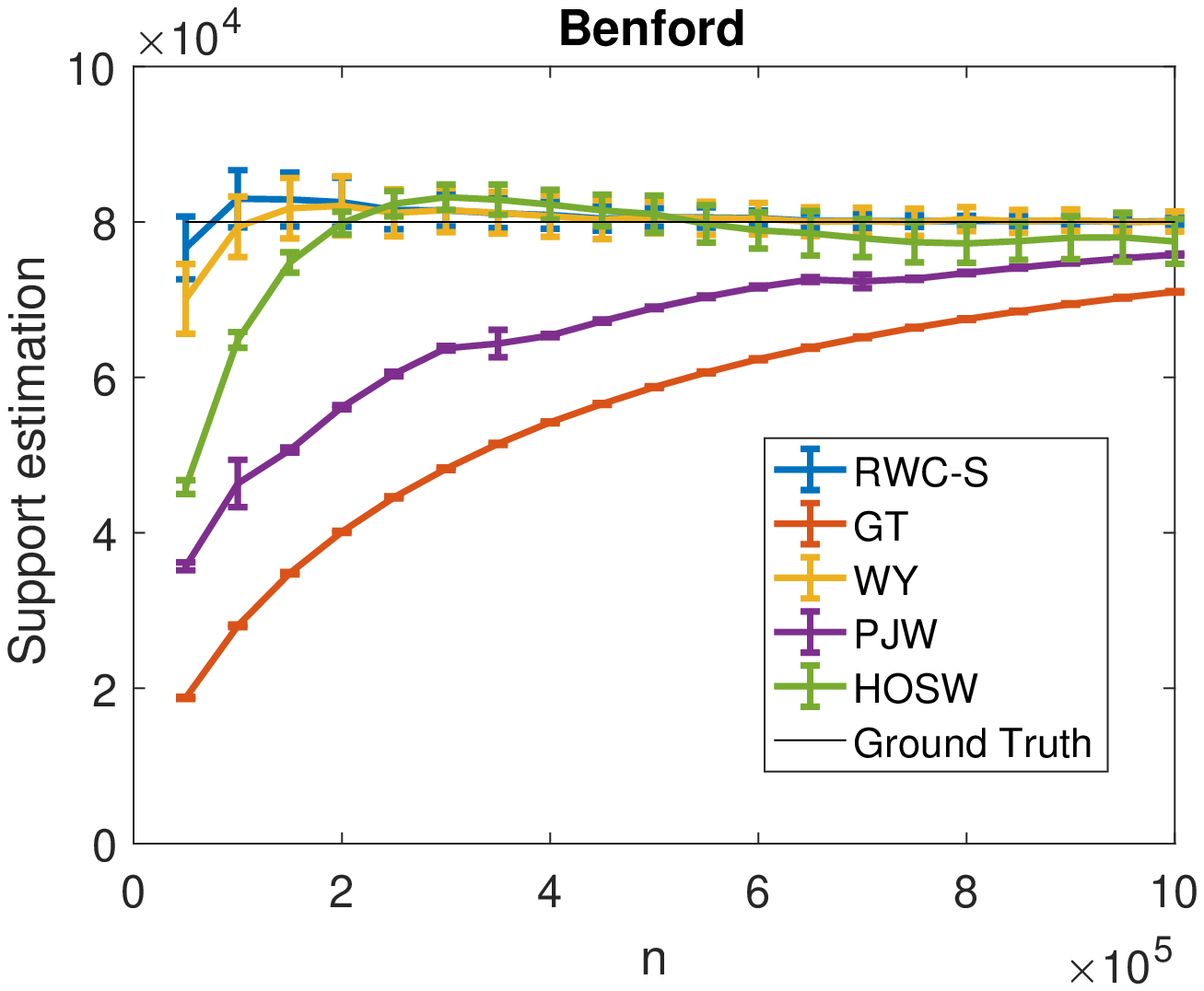}}
  \subfigure[Zipf$(1.5)$ distribution.]{\includegraphics[width=0.32\linewidth]{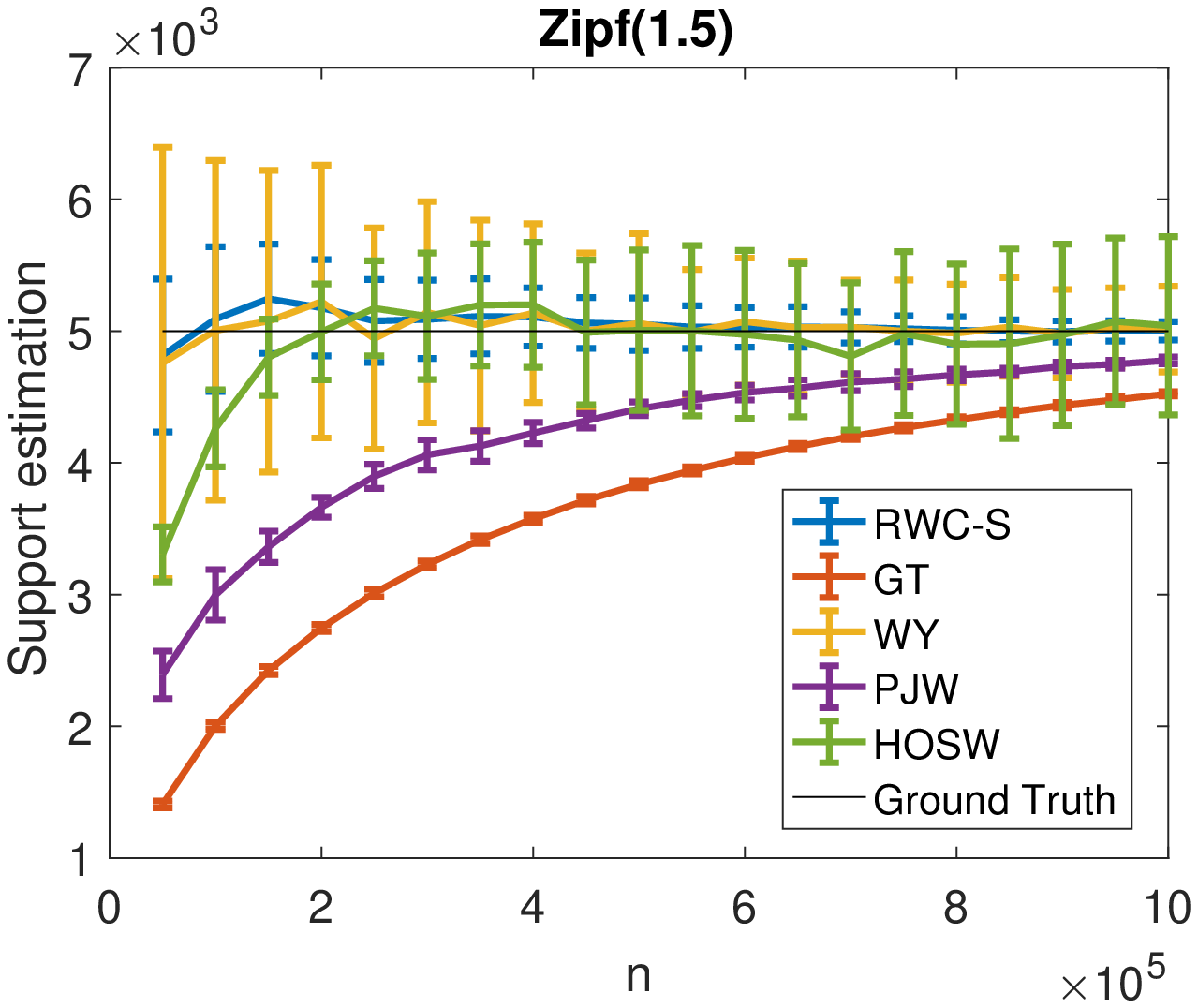}}
  \subfigure[Zipf$(1)$ distribution.]{\includegraphics[width=0.32\linewidth]{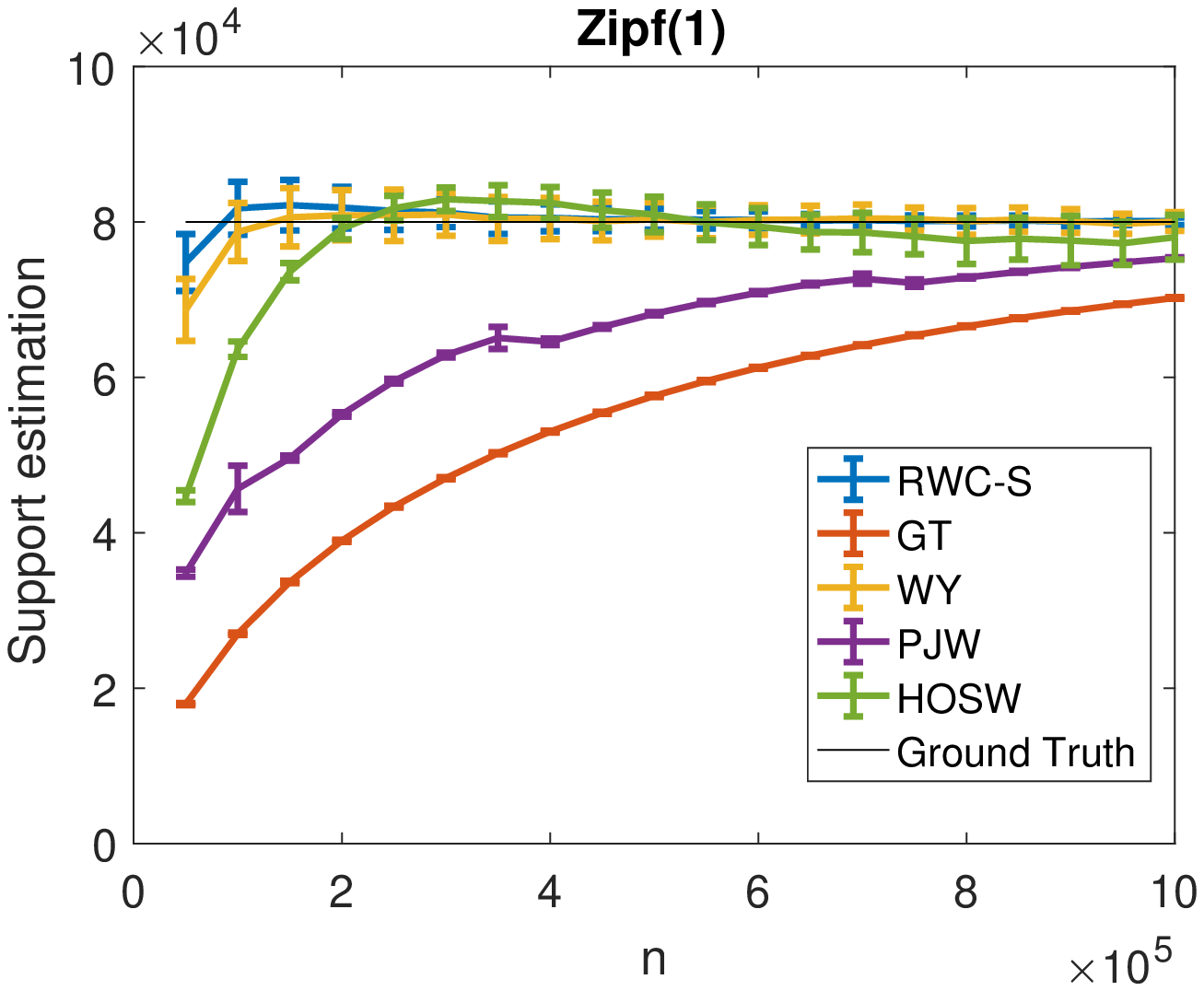}}
  \subfigure[Zipf$(0.5)$ distribution.]{\includegraphics[width=0.32\linewidth]{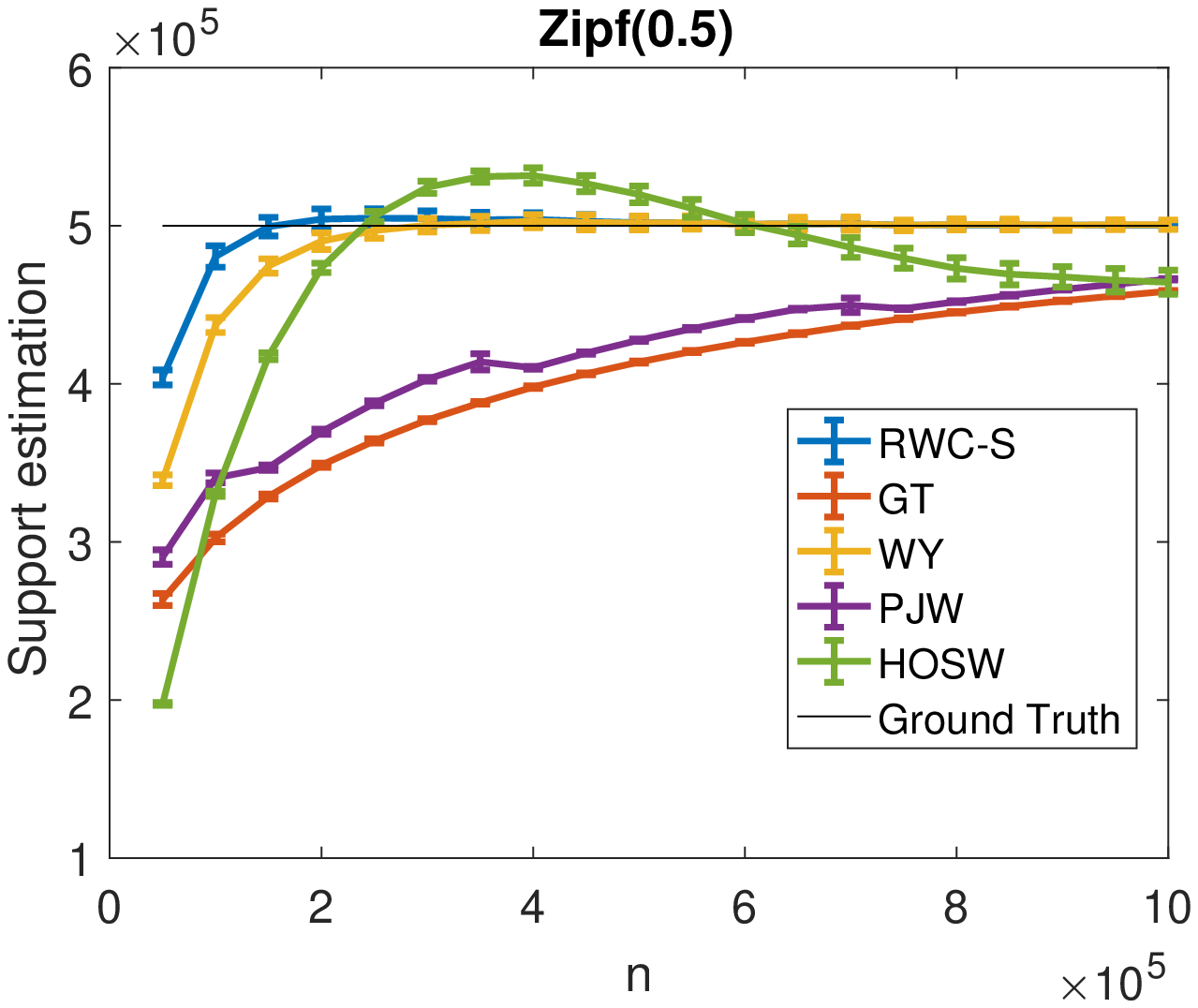}}
  \subfigure[Zipf$(0.25)$ distribution.]{\includegraphics[width=0.32\linewidth]{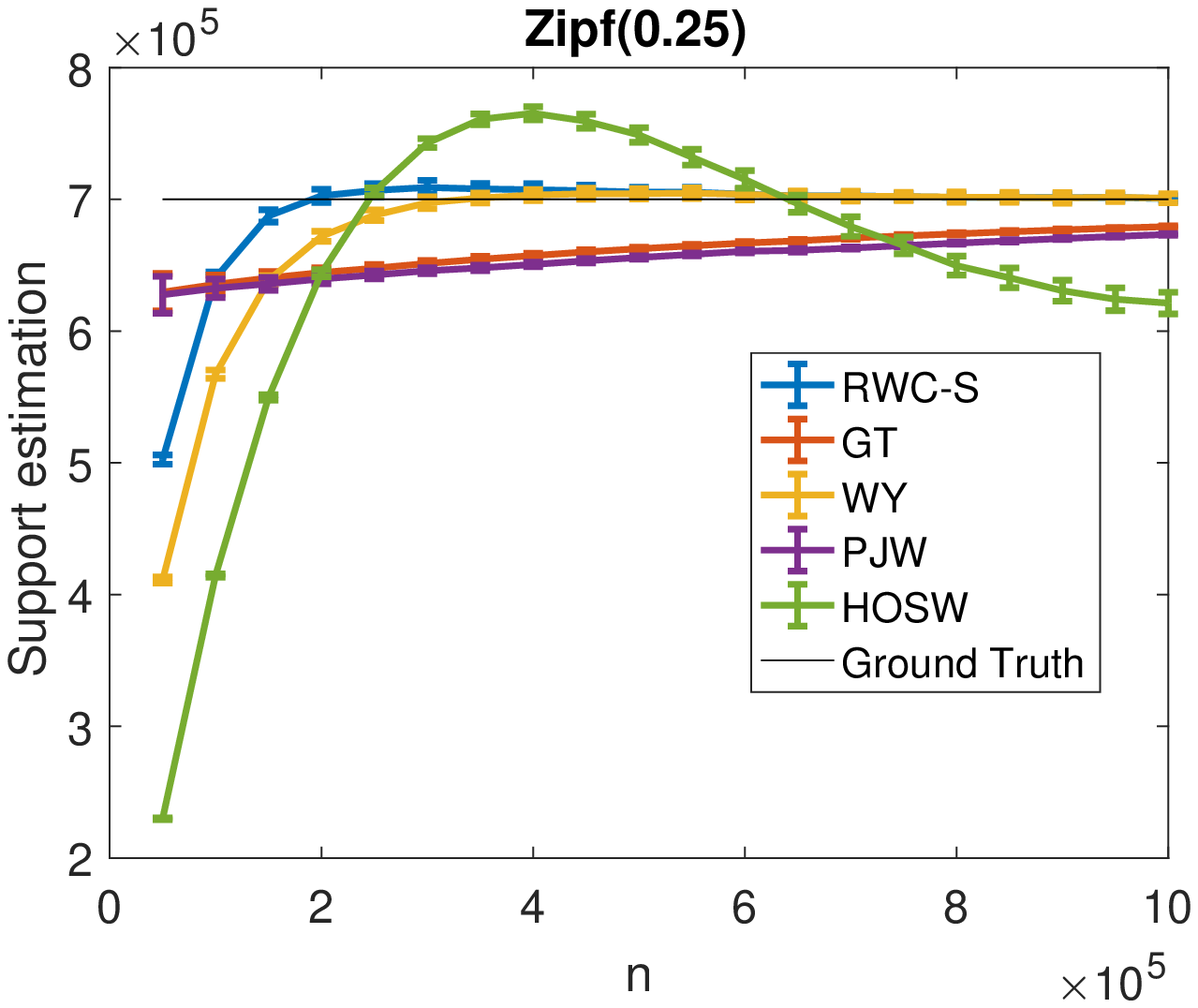}}
    \caption{The figures plot the mean and variance of the estimators for the tested distributions without Poisson repeats. The $y$-axis is on the log scale.}
\end{figure}

\begin{figure}
    \centering
    \subfigure[Uniform distribution.]{\includegraphics[width=0.32\linewidth]{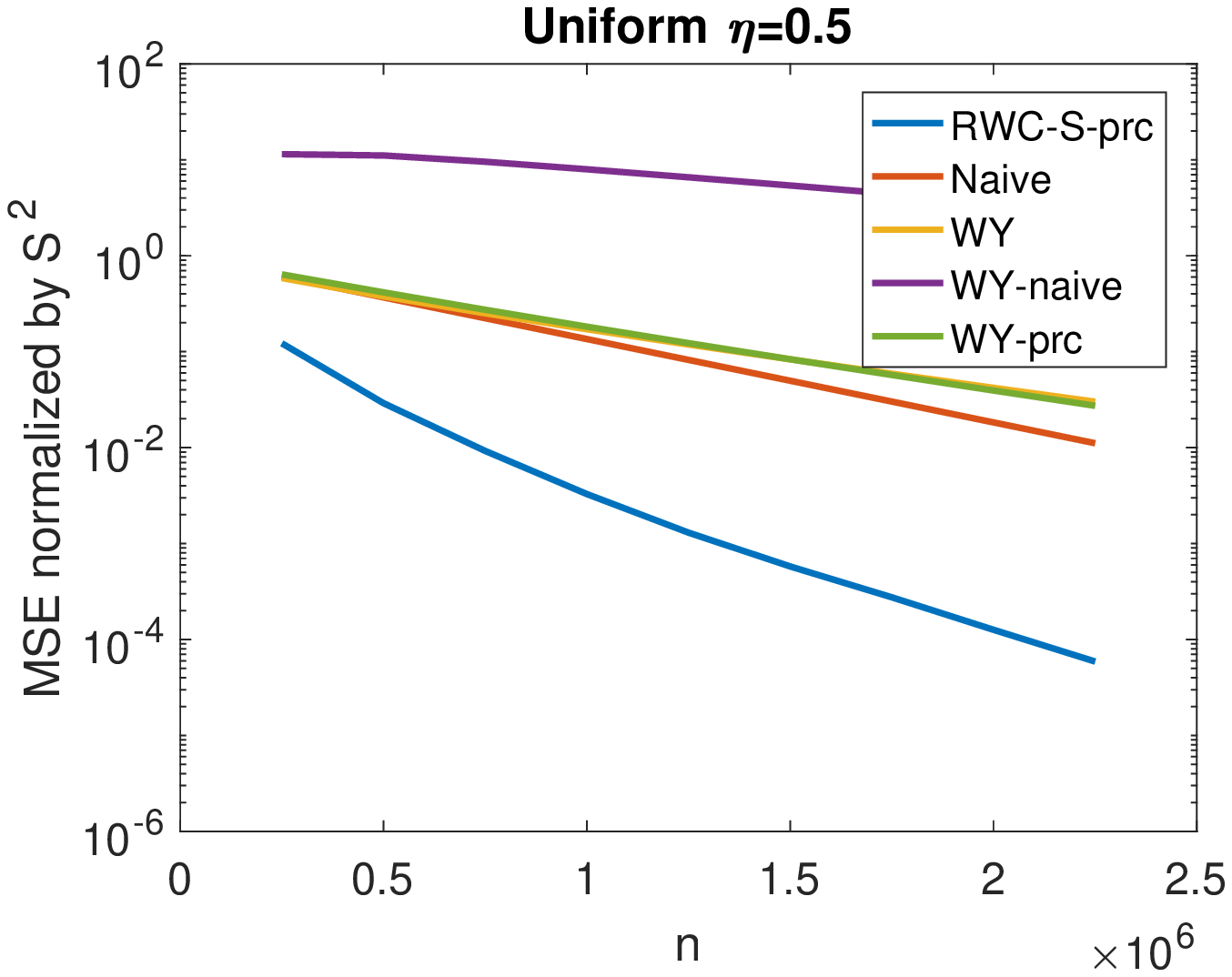}}
  \subfigure[Benford distribution.]{\includegraphics[width=0.32\linewidth]{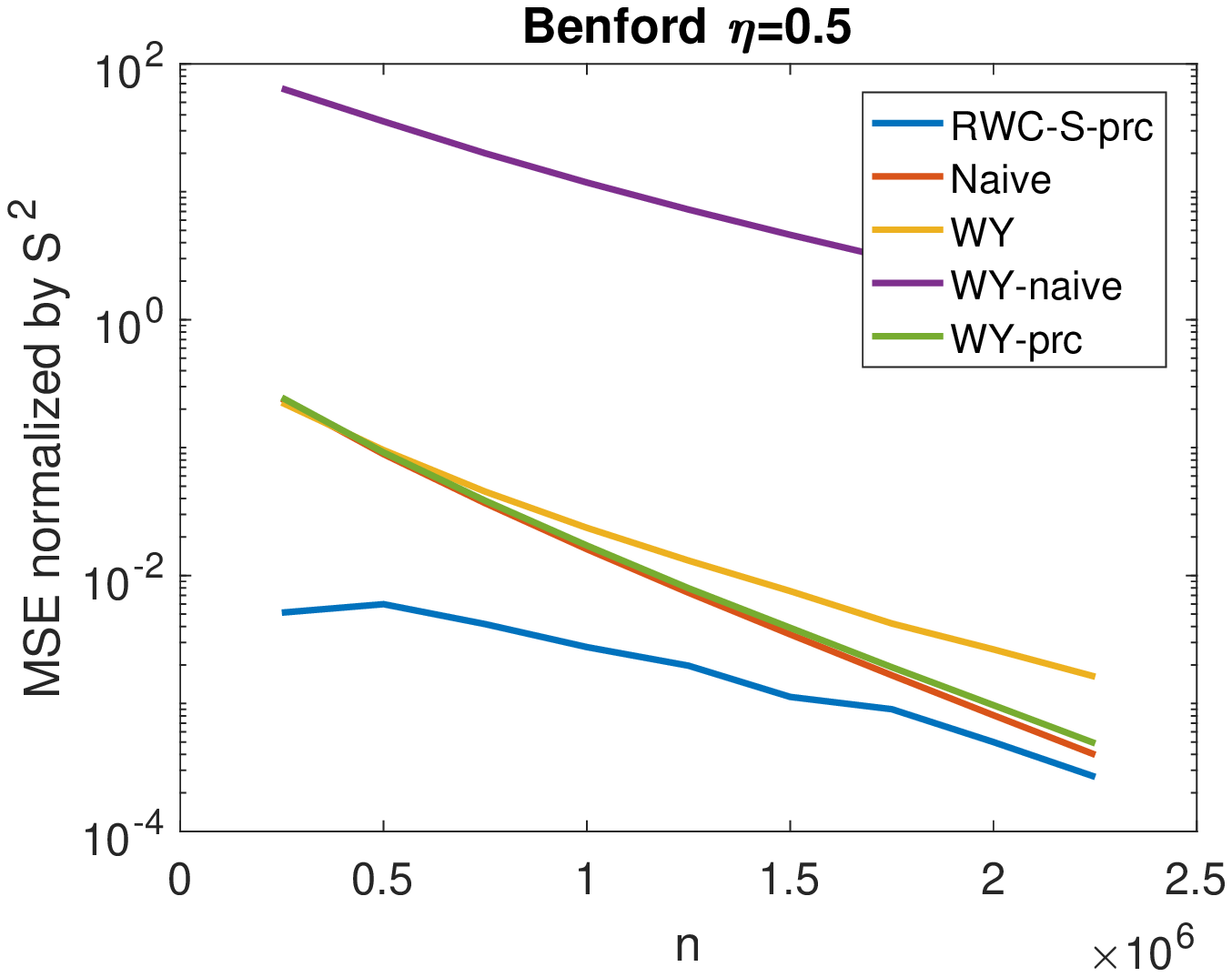}}
  \subfigure[Zipf$(1.5)$ distribution.]{\includegraphics[width=0.32\linewidth]{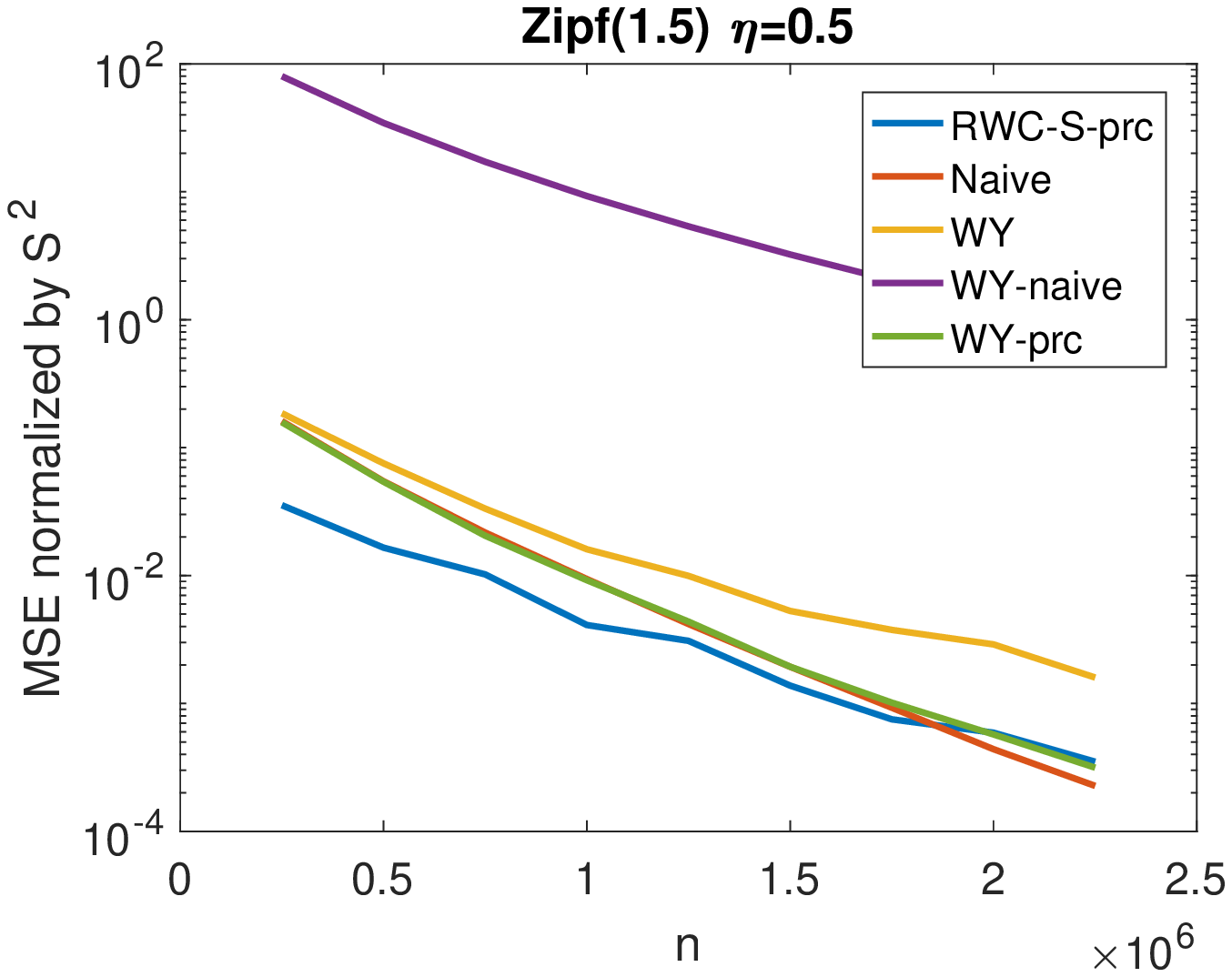}}
  \subfigure[Zipf$(1)$ distribution.]{\includegraphics[width=0.32\linewidth]{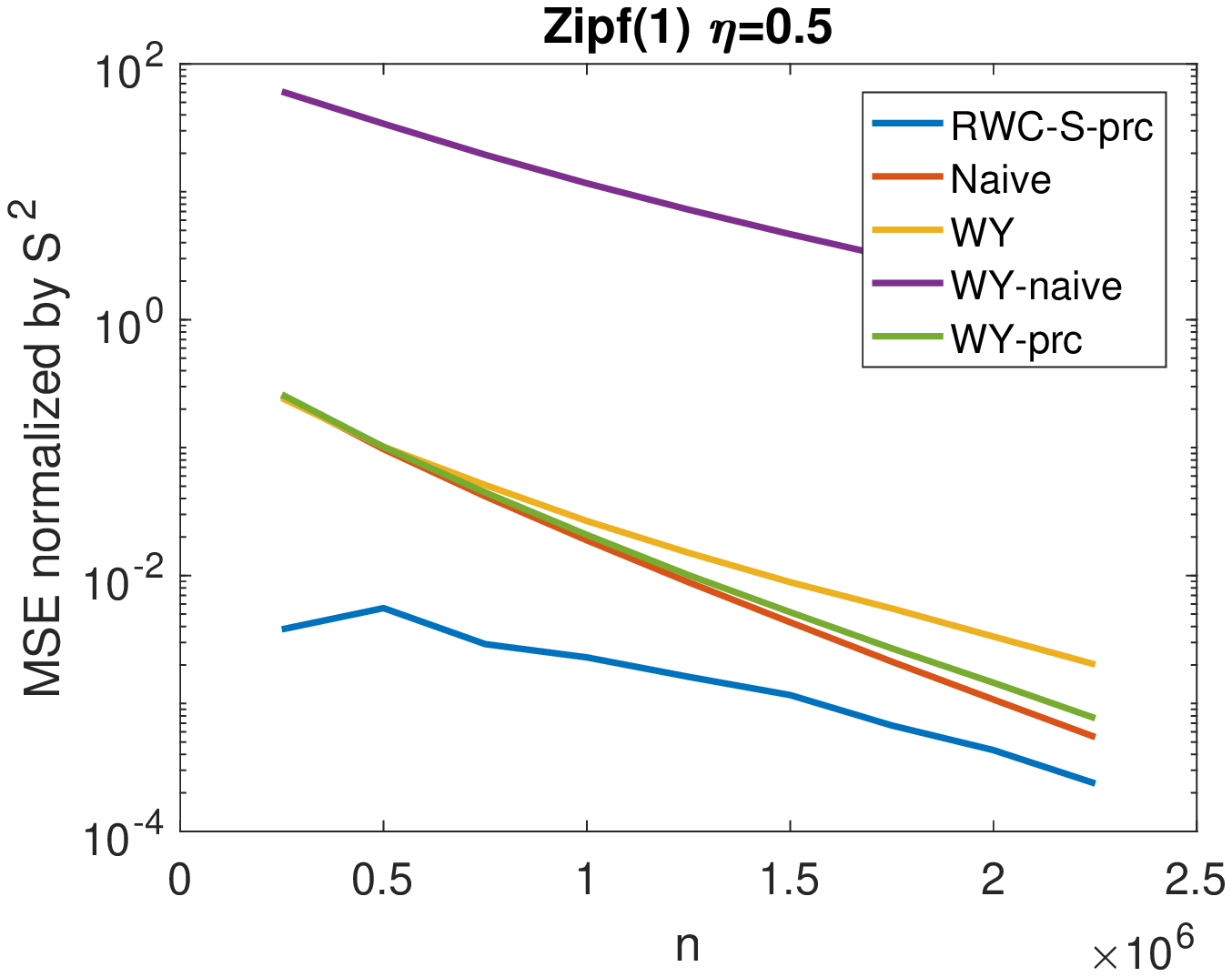}}
  \subfigure[Zipf$(0.5)$ distribution.]{\includegraphics[width=0.32\linewidth]{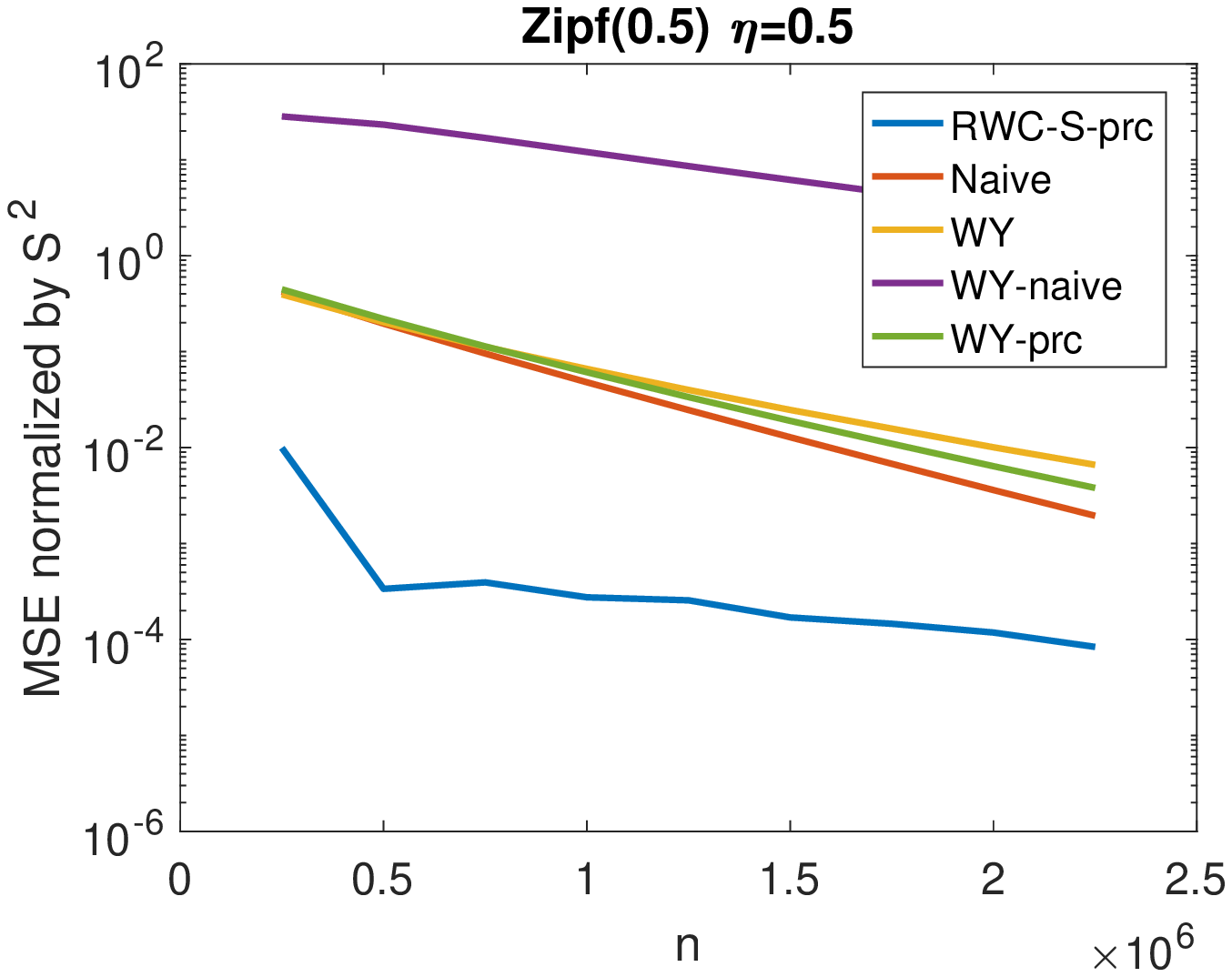}}
  \subfigure[Zipf$(0.25)$ distribution.]{\includegraphics[width=0.32\linewidth]{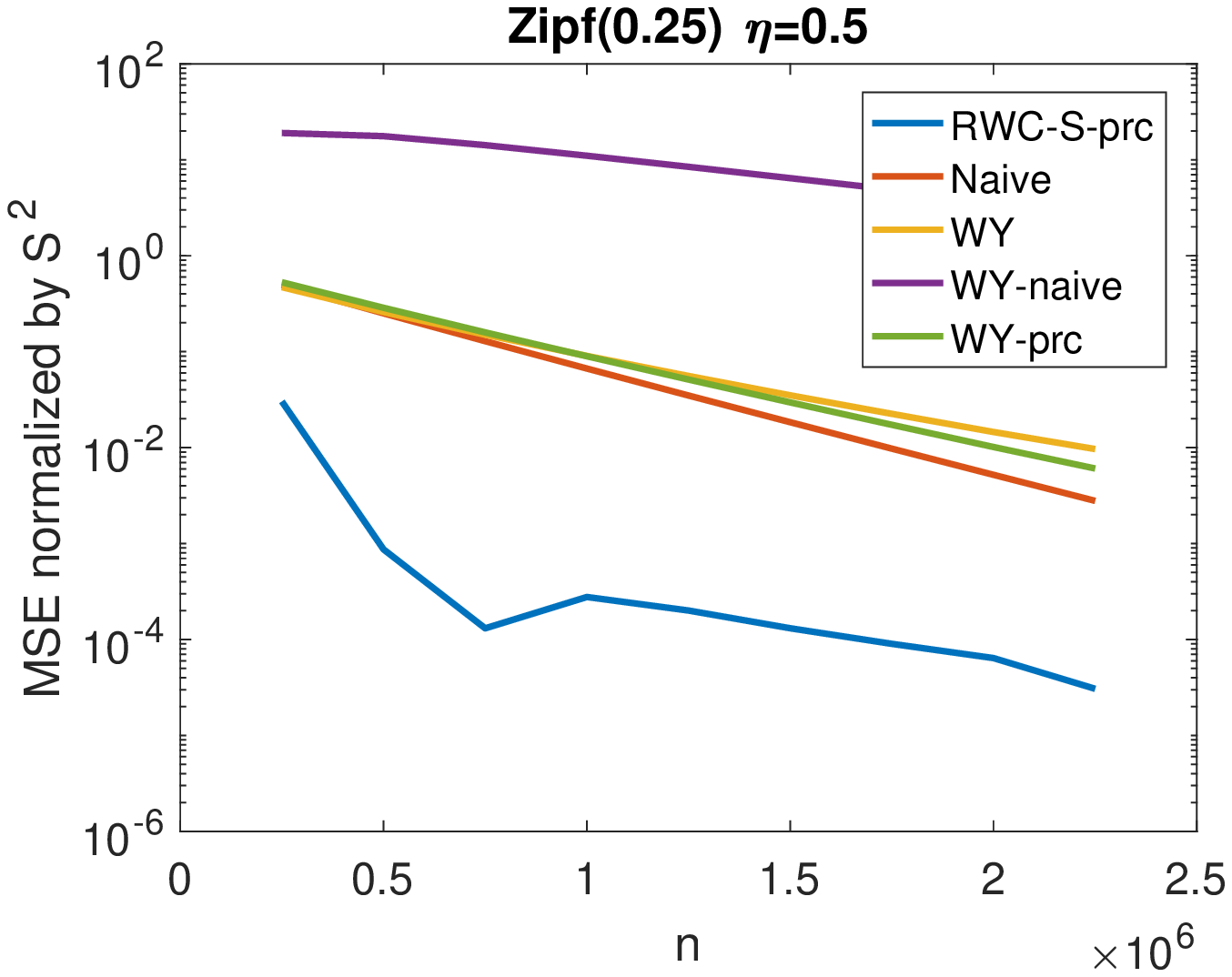}}
    \caption{The figures plot the MSE of all estimators considered on each tested distributions with Poisson repeats, $\eta = 0.5$. The $y$-axis is on the log scale.}
\end{figure}

\begin{figure}
    \centering
    \subfigure[Uniform distribution.]{\includegraphics[width=0.32\linewidth]{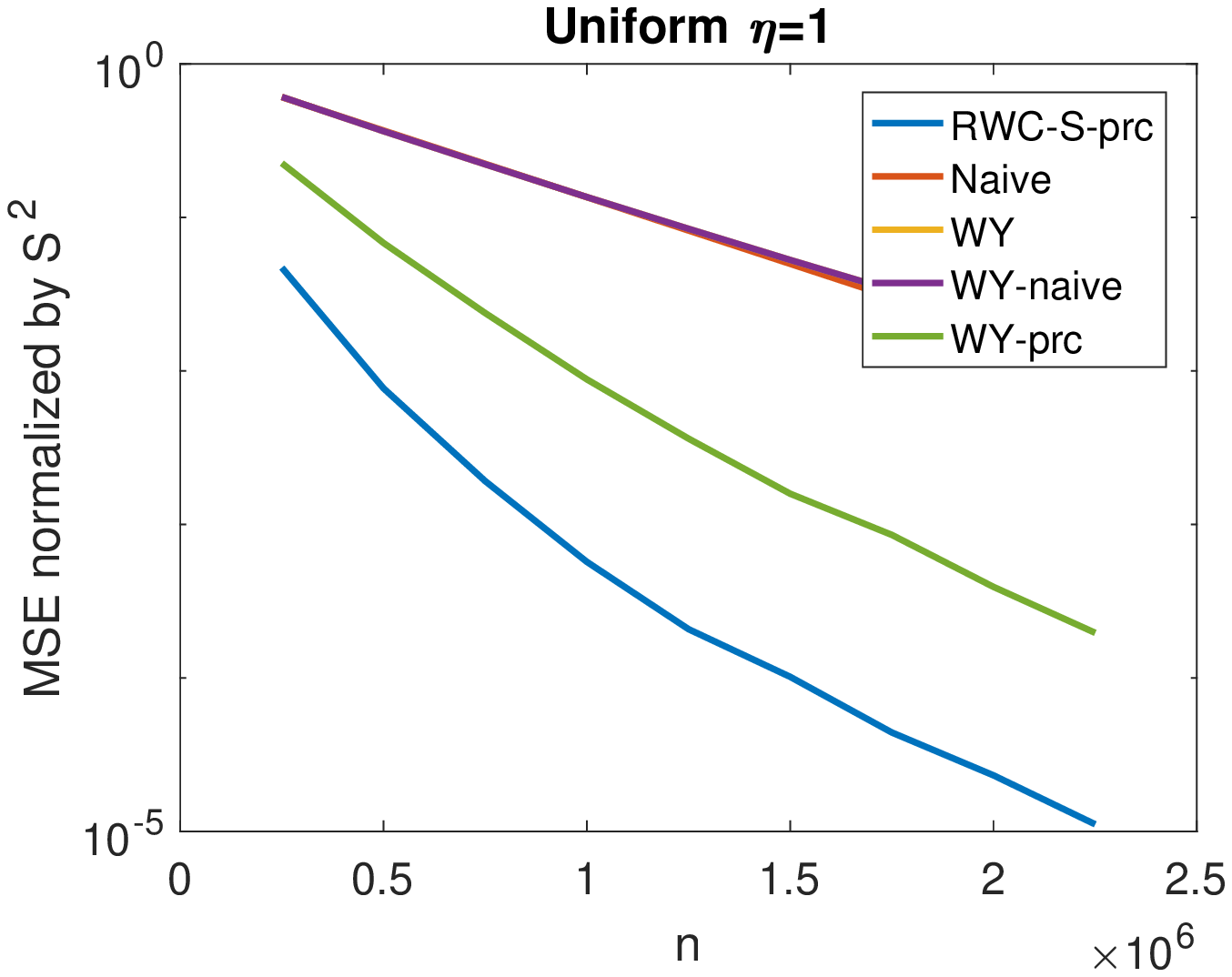}}
  \subfigure[Benford distribution.]{\includegraphics[width=0.32\linewidth]{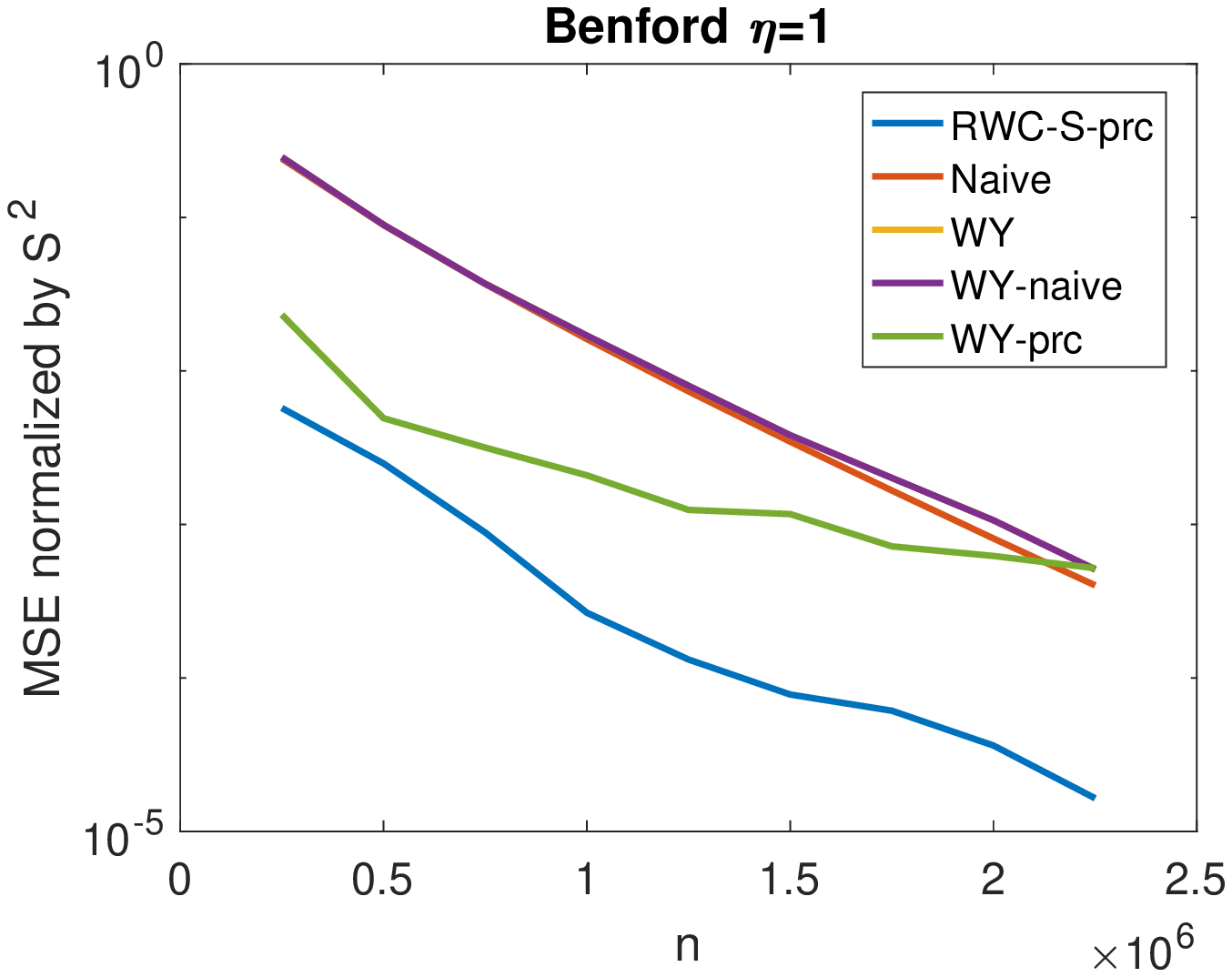}}
  \subfigure[Zipf$(1.5)$ distribution.]{\includegraphics[width=0.32\linewidth]{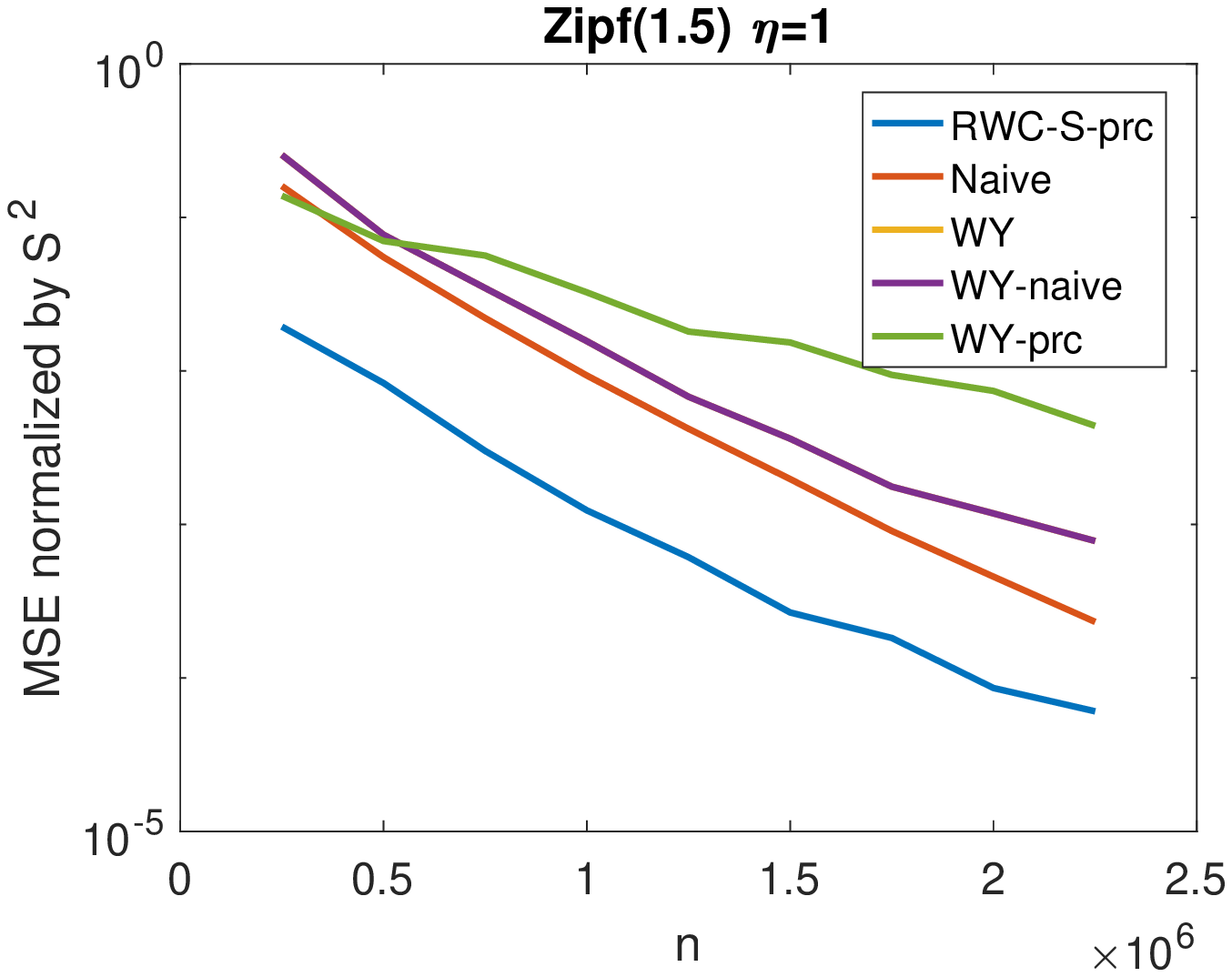}}
  \subfigure[Zipf$(1)$ distribution.]{\includegraphics[width=0.32\linewidth]{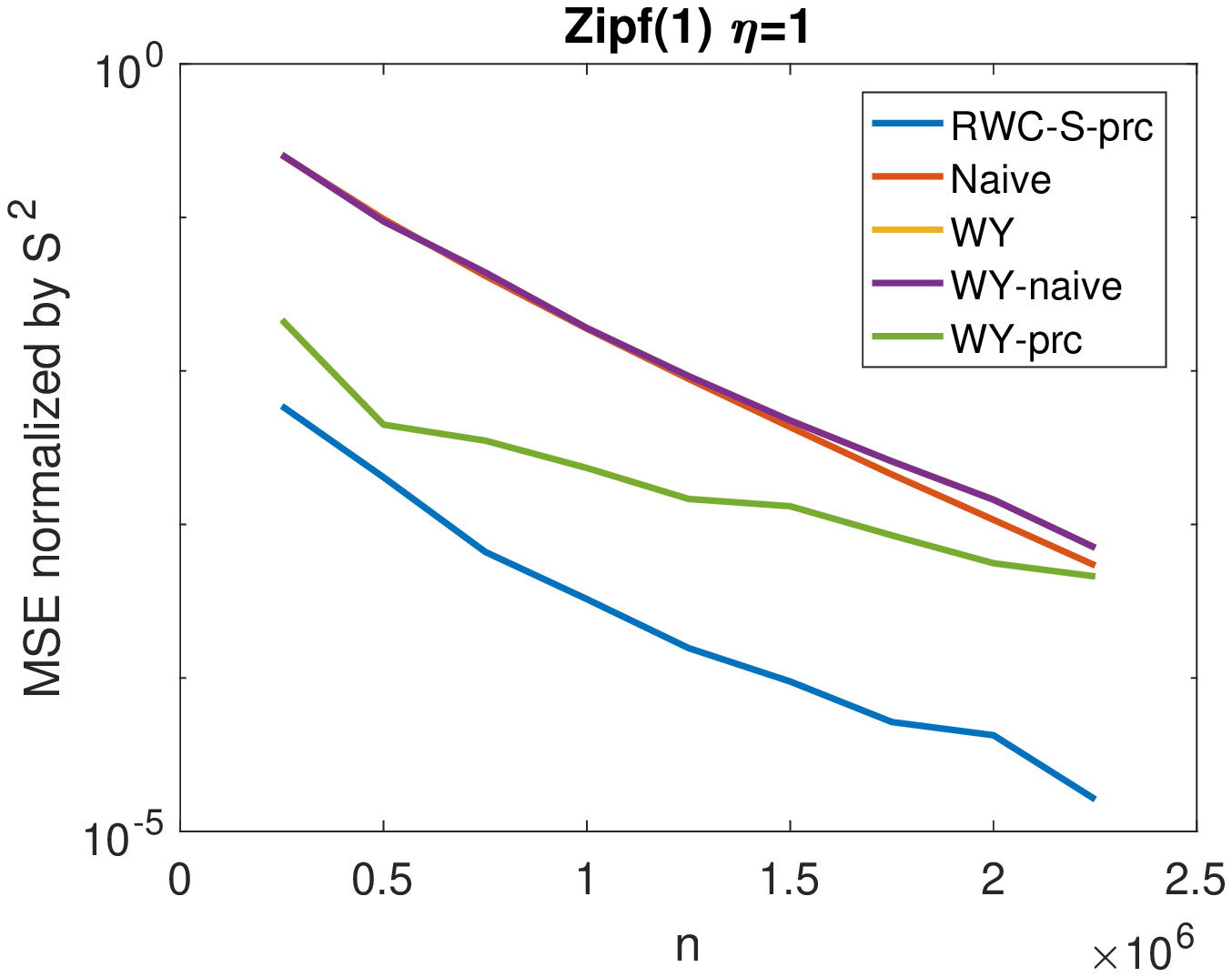}}
  \subfigure[Zipf$(0.5)$ distribution.]{\includegraphics[width=0.32\linewidth]{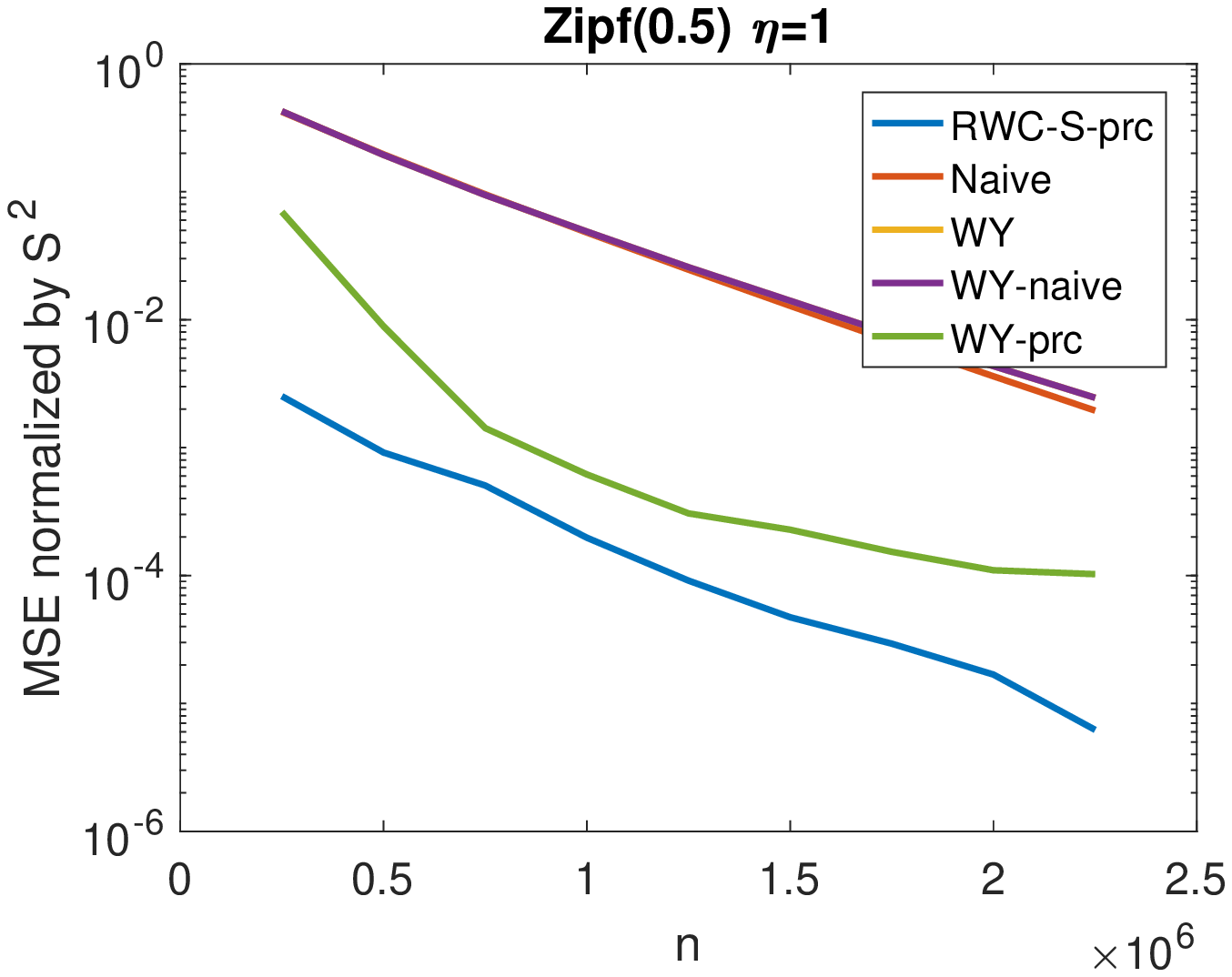}}
  \subfigure[Zipf$(0.25)$ distribution.]{\includegraphics[width=0.32\linewidth]{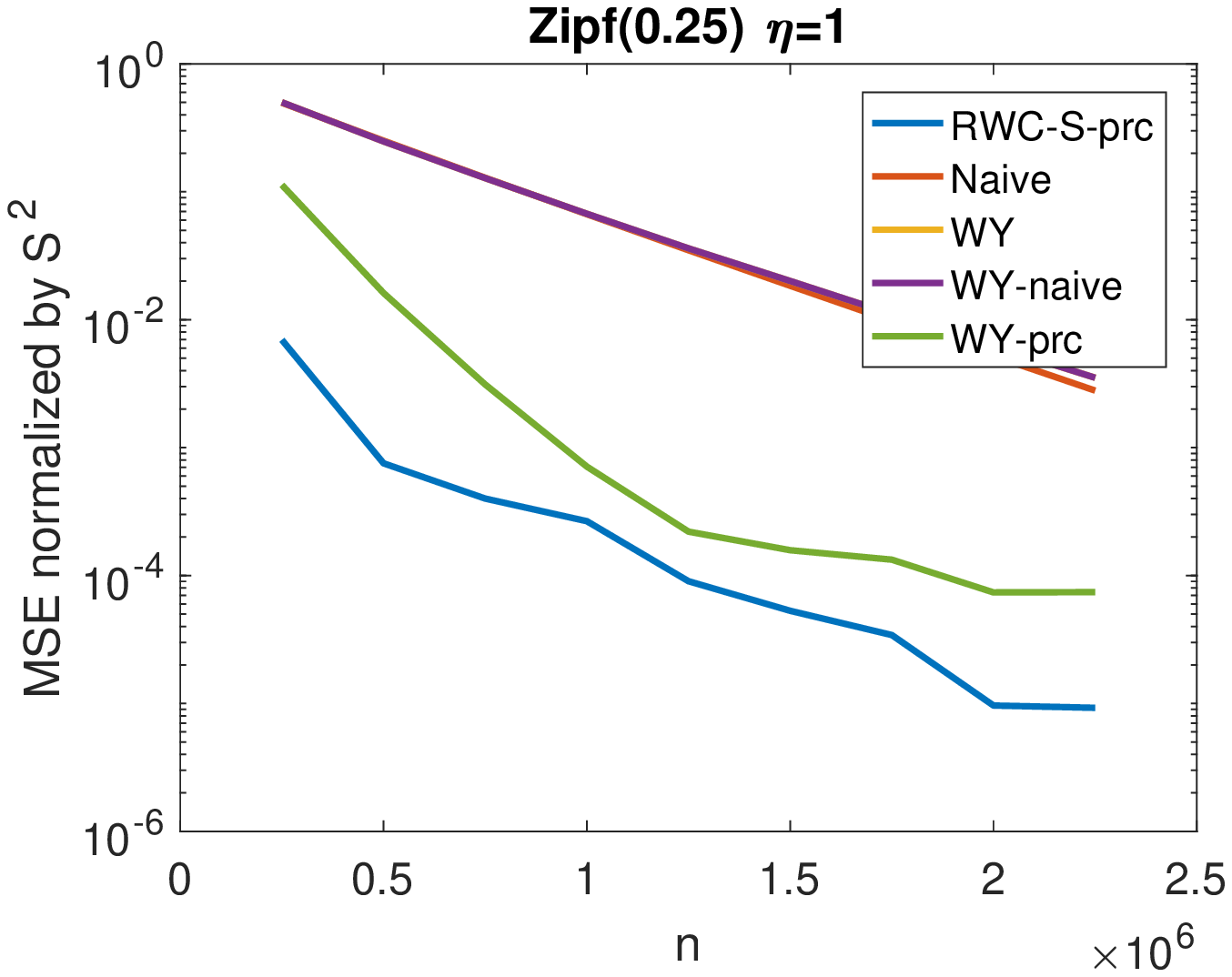}}
    \caption{The figures plot the MSE of all estimators considered on each tested distributions with Poisson repeats, $\eta = 1$. The $y$-axis is on the log scale.}
\end{figure}

\begin{figure}
    \centering
    \subfigure[Uniform distribution.]{\includegraphics[width=0.32\linewidth]{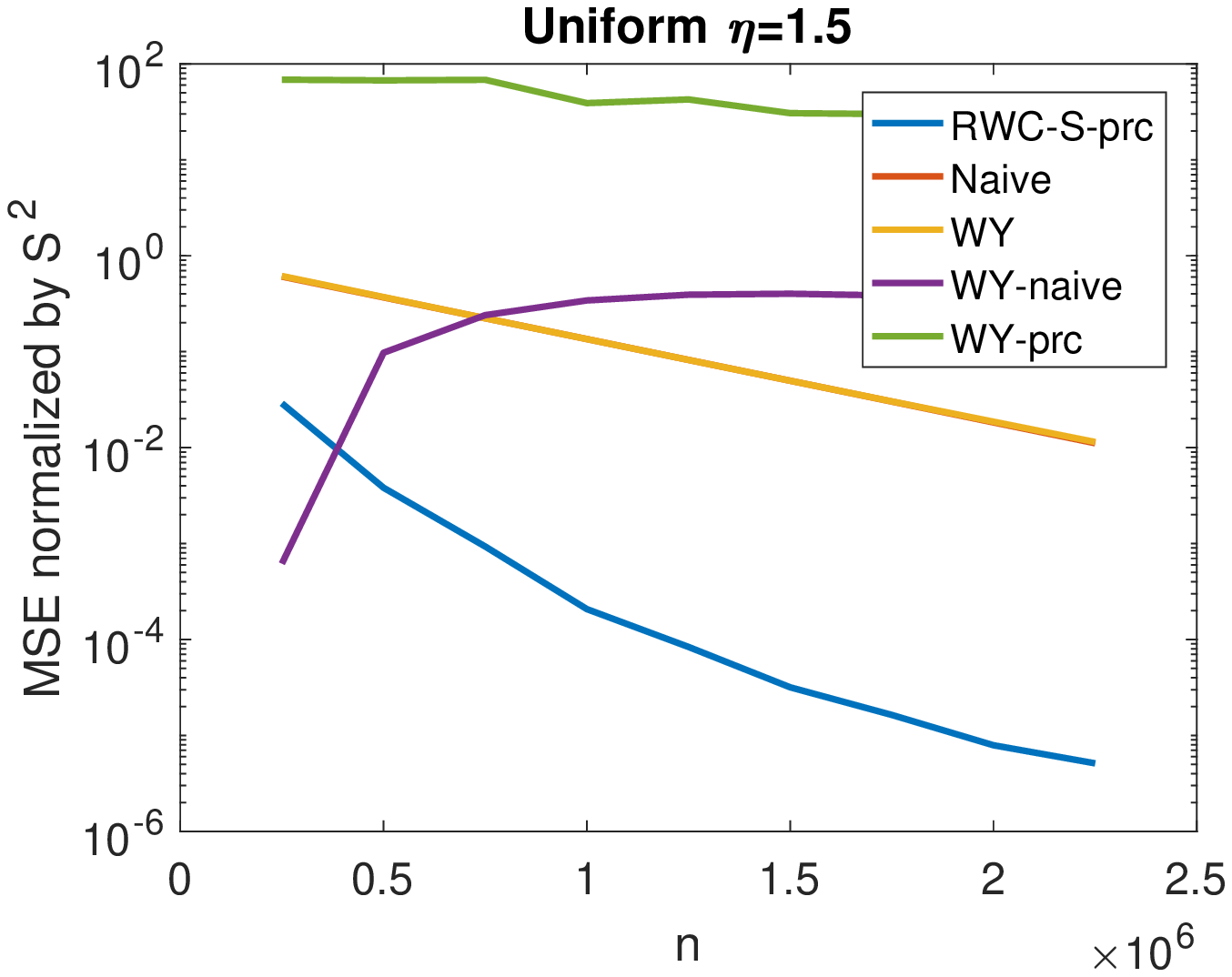}}
  \subfigure[Benford distribution.]{\includegraphics[width=0.32\linewidth]{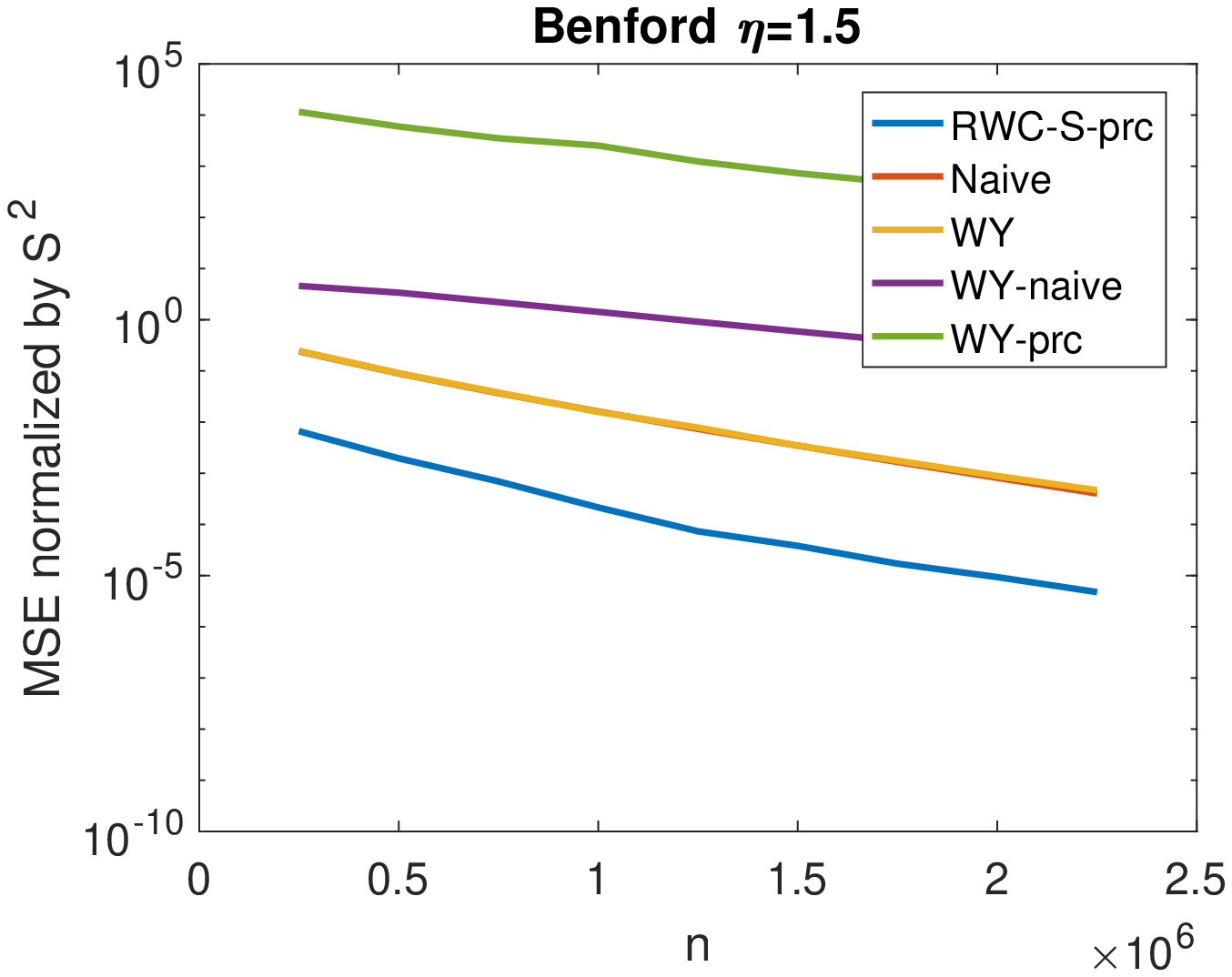}}
  \subfigure[Zipf$(1.5)$ distribution.]{\includegraphics[width=0.32\linewidth]{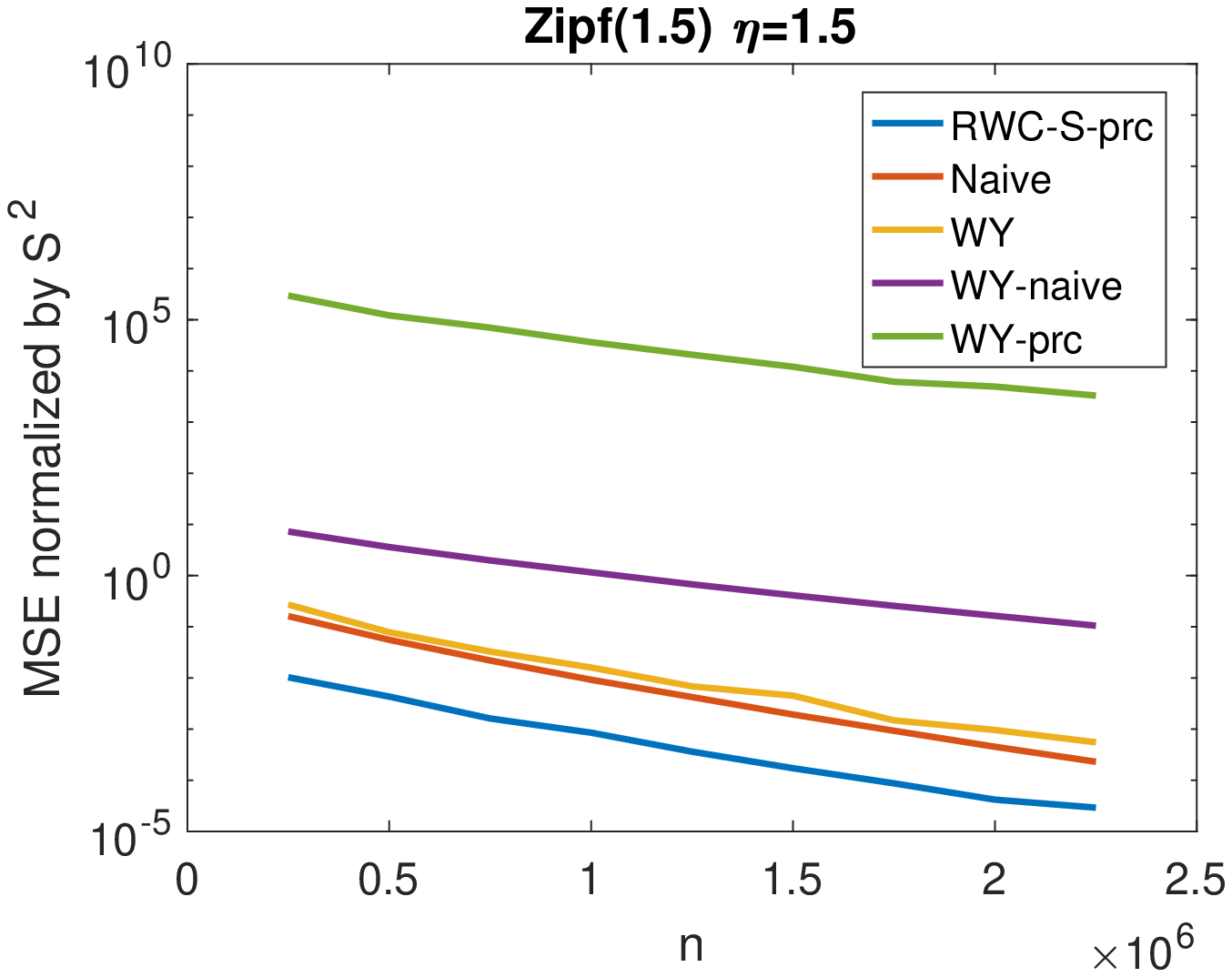}}
  \subfigure[Zipf$(1)$ distribution.]{\includegraphics[width=0.32\linewidth]{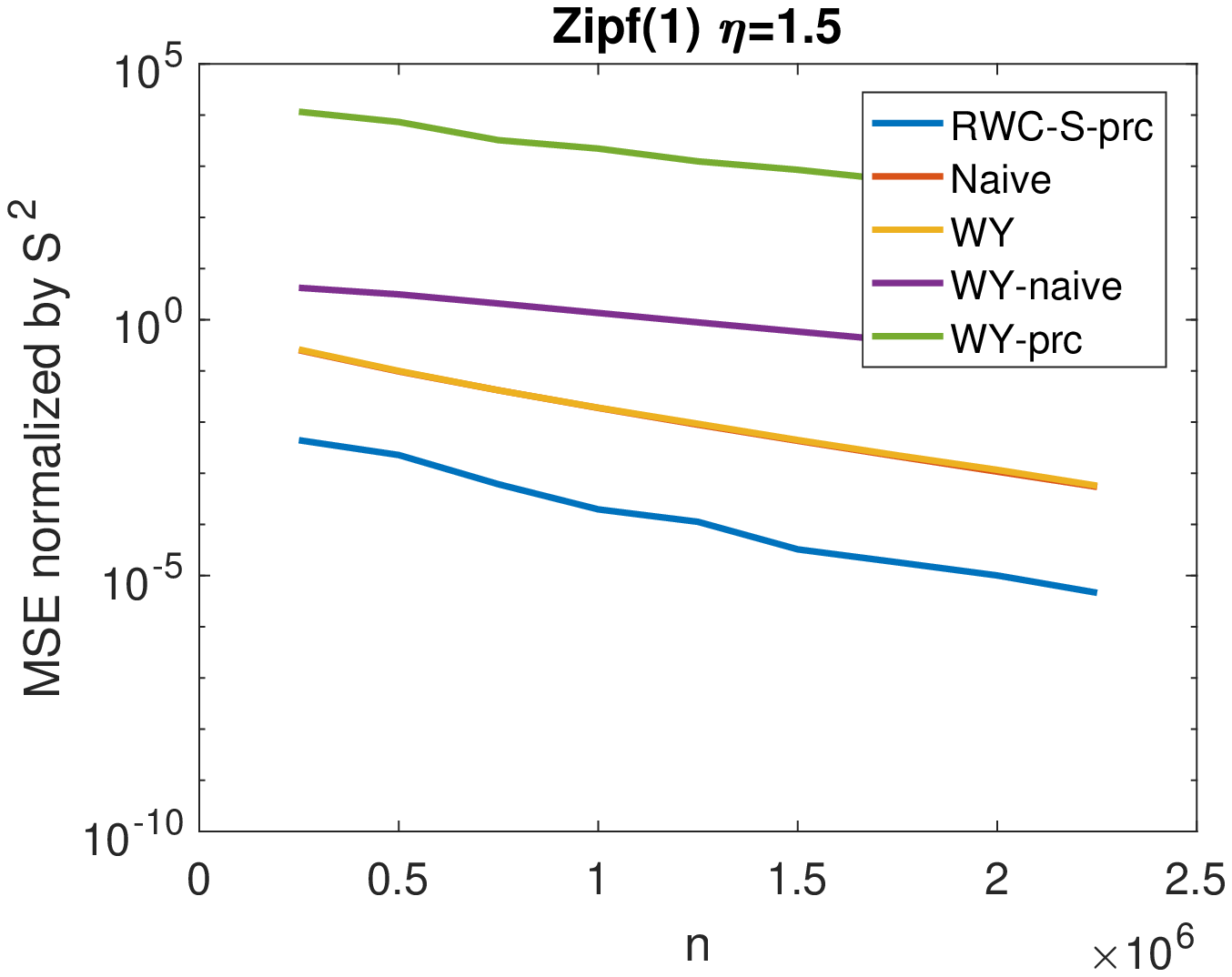}}
  \subfigure[Zipf$(0.5)$ distribution.]{\includegraphics[width=0.32\linewidth]{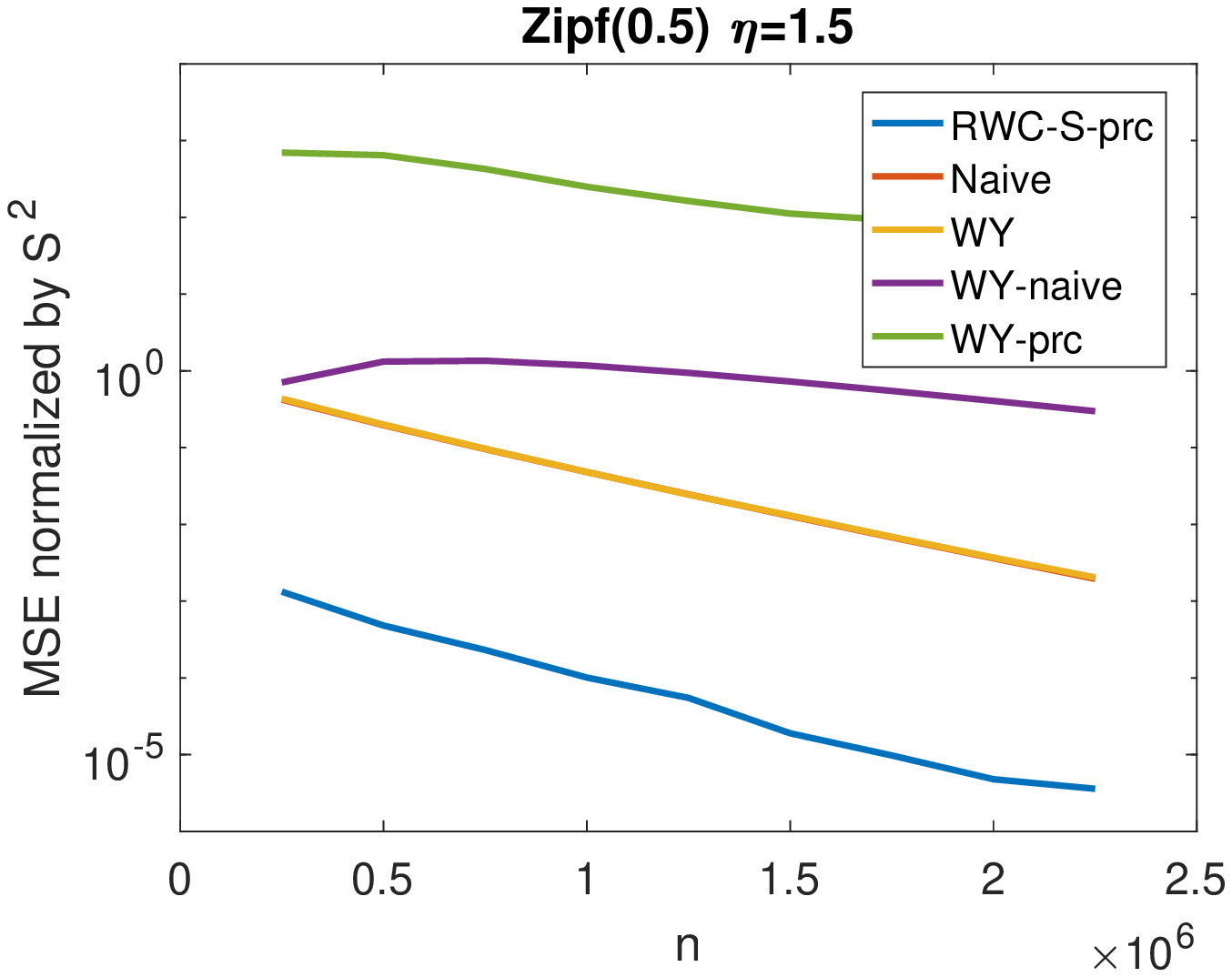}}
  \subfigure[Zipf$(0.25)$ distribution.]{\includegraphics[width=0.32\linewidth]{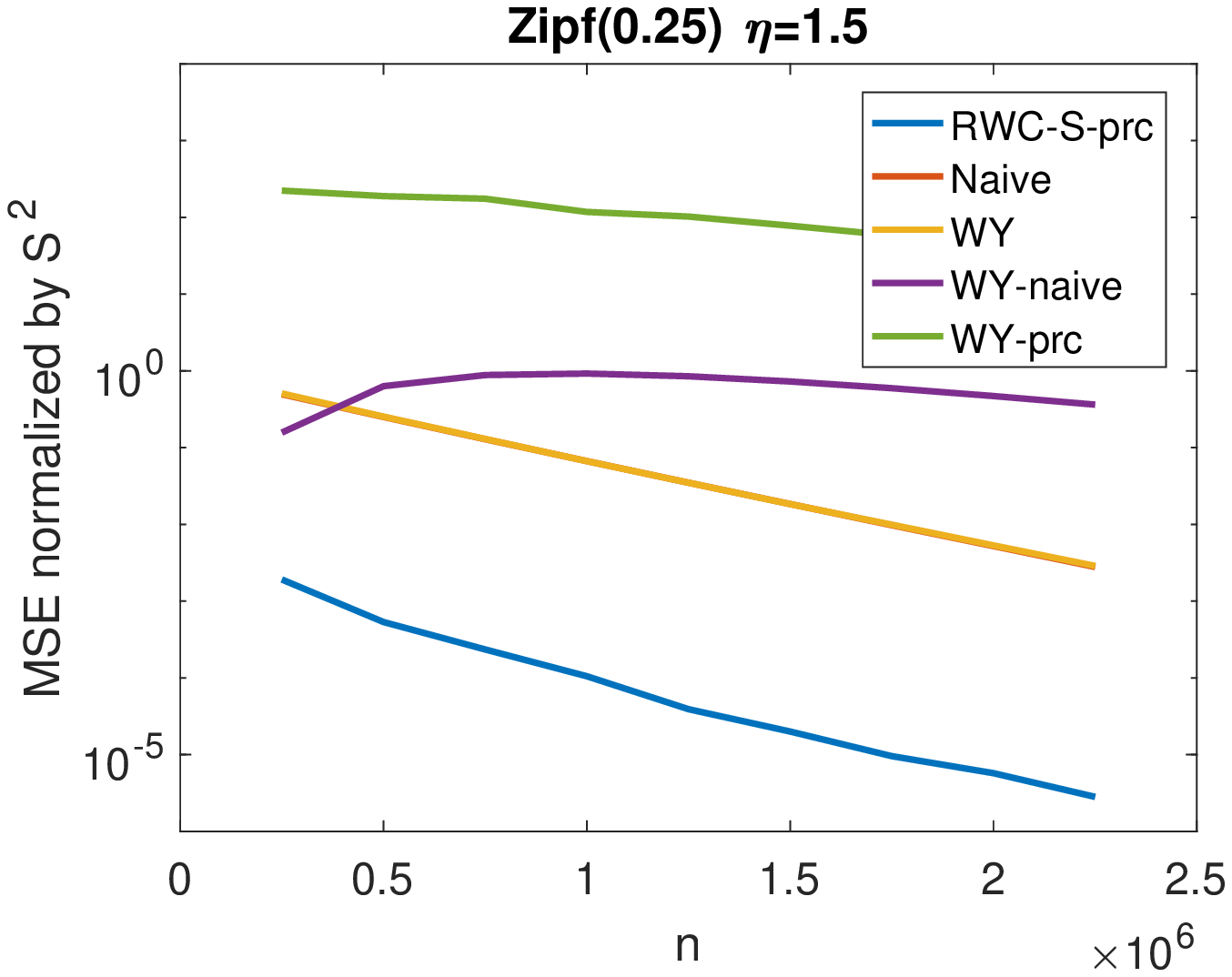}}
    \caption{The figures plot the MSE of all estimators considered on each tested distributions with Poisson repeats, $\eta = 1.5$. The $y$-axis is on the log scale.}
\end{figure}

\begin{figure}[!htb]
  \centering
  \subfigure[]{\includegraphics[width=0.245\linewidth]{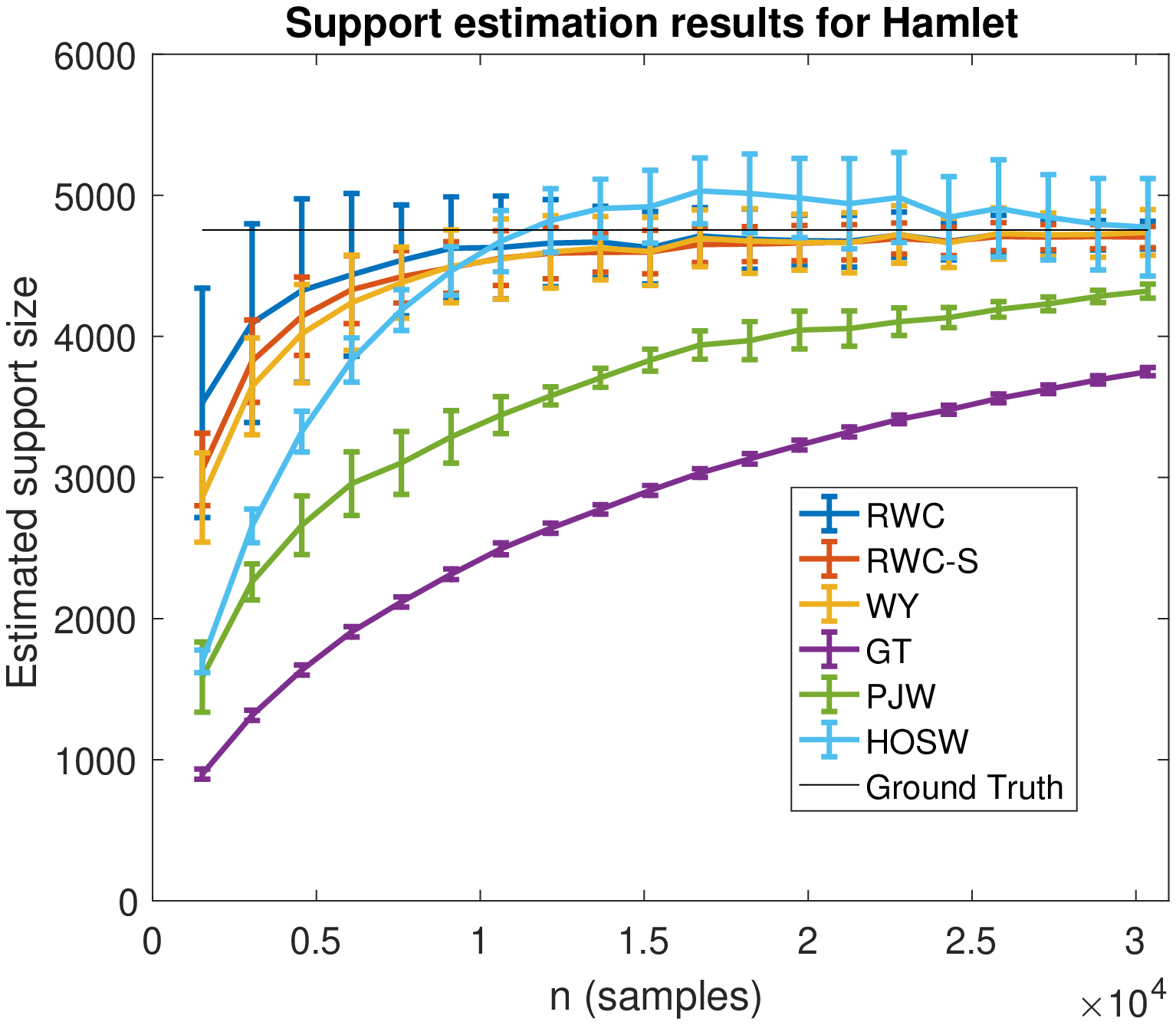}}
  \subfigure[]{\includegraphics[width=0.245\linewidth]{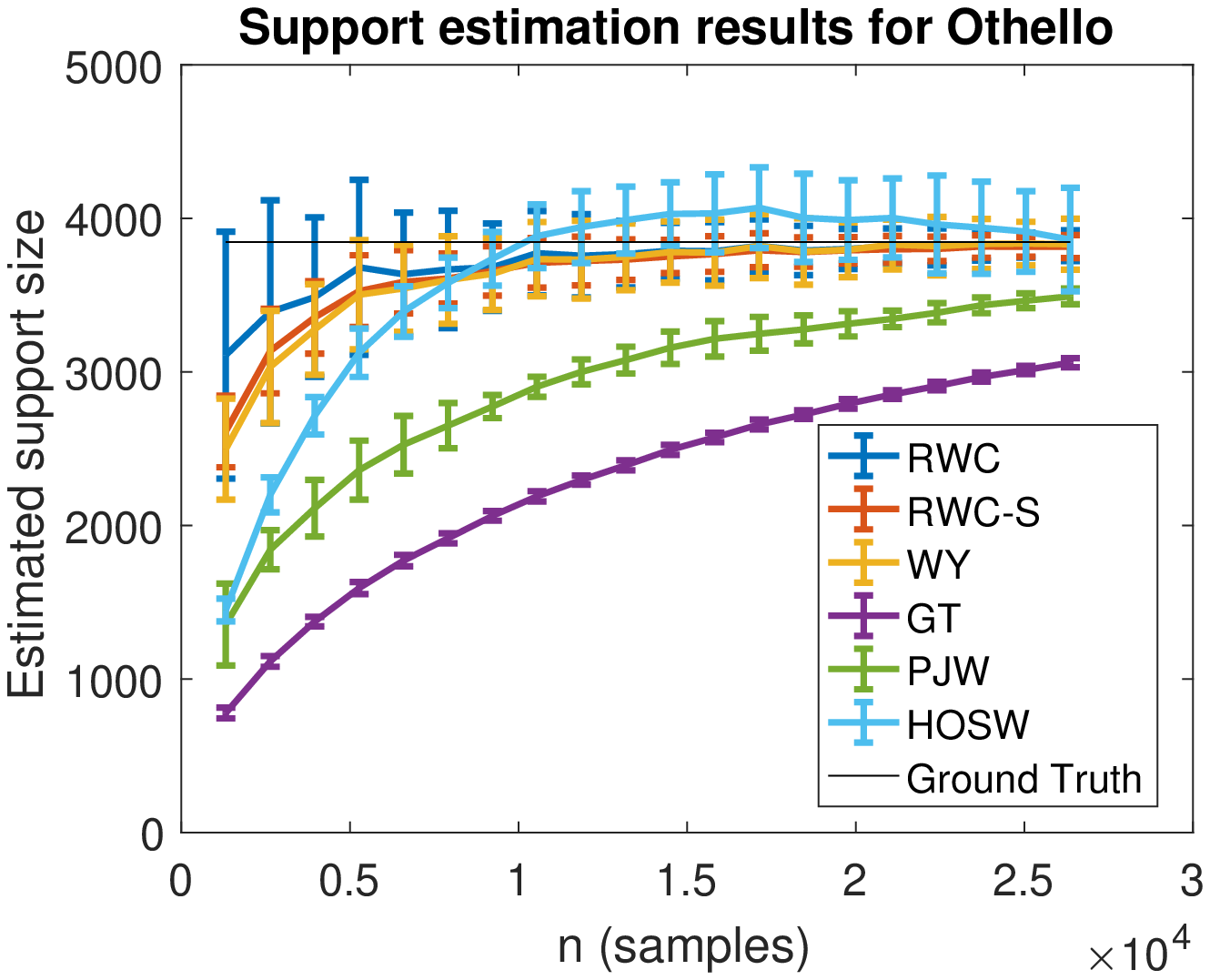}}
  \subfigure[]{\includegraphics[width=0.245\linewidth]{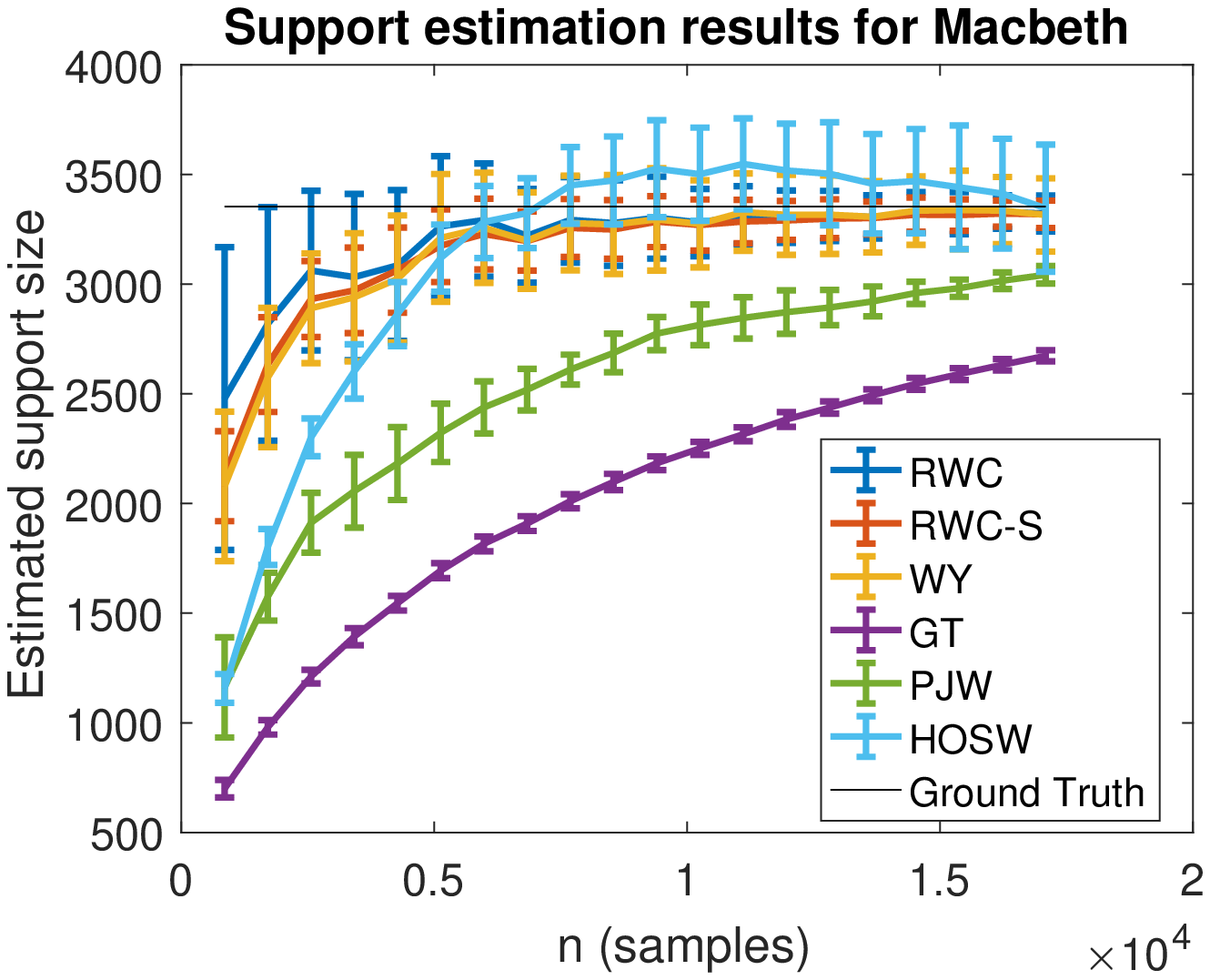}}
  \subfigure[]{\includegraphics[width=0.245\linewidth]{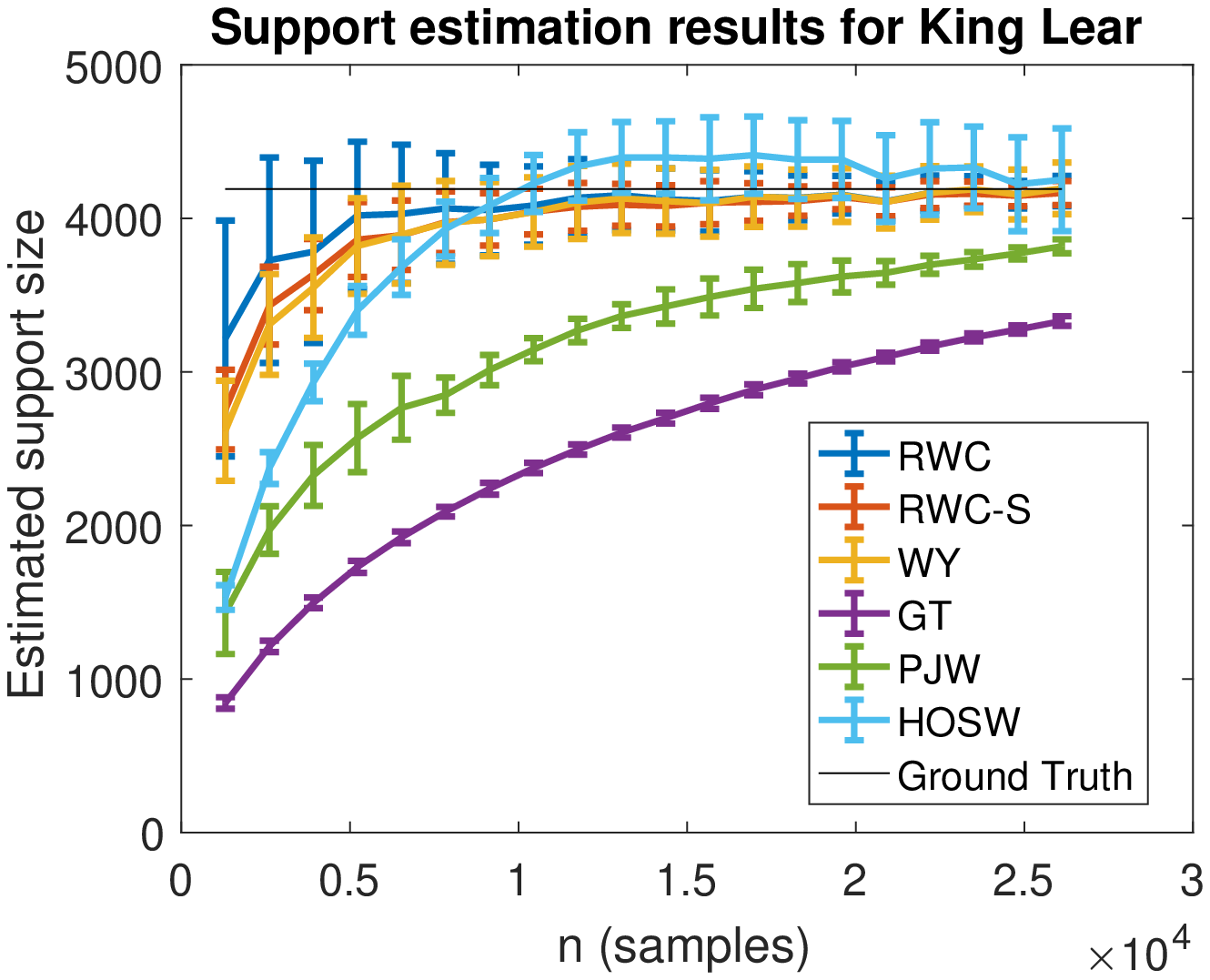}}
  \subfigure[]{\includegraphics[width=0.245\linewidth]{Hamlet_MSES2.eps}}
  \subfigure[]{\includegraphics[width=0.245\linewidth]{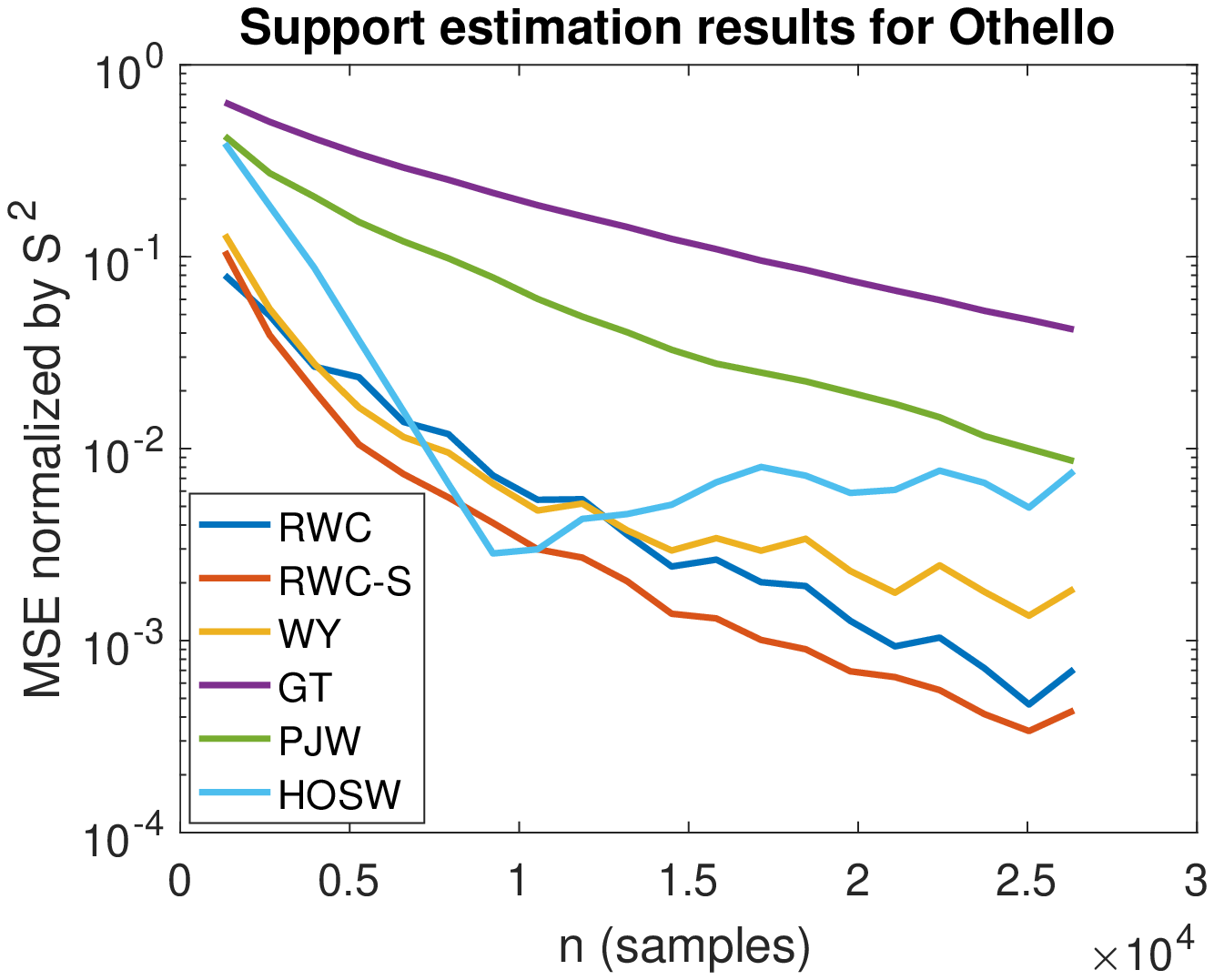}}
  \subfigure[]{\includegraphics[width=0.245\linewidth]{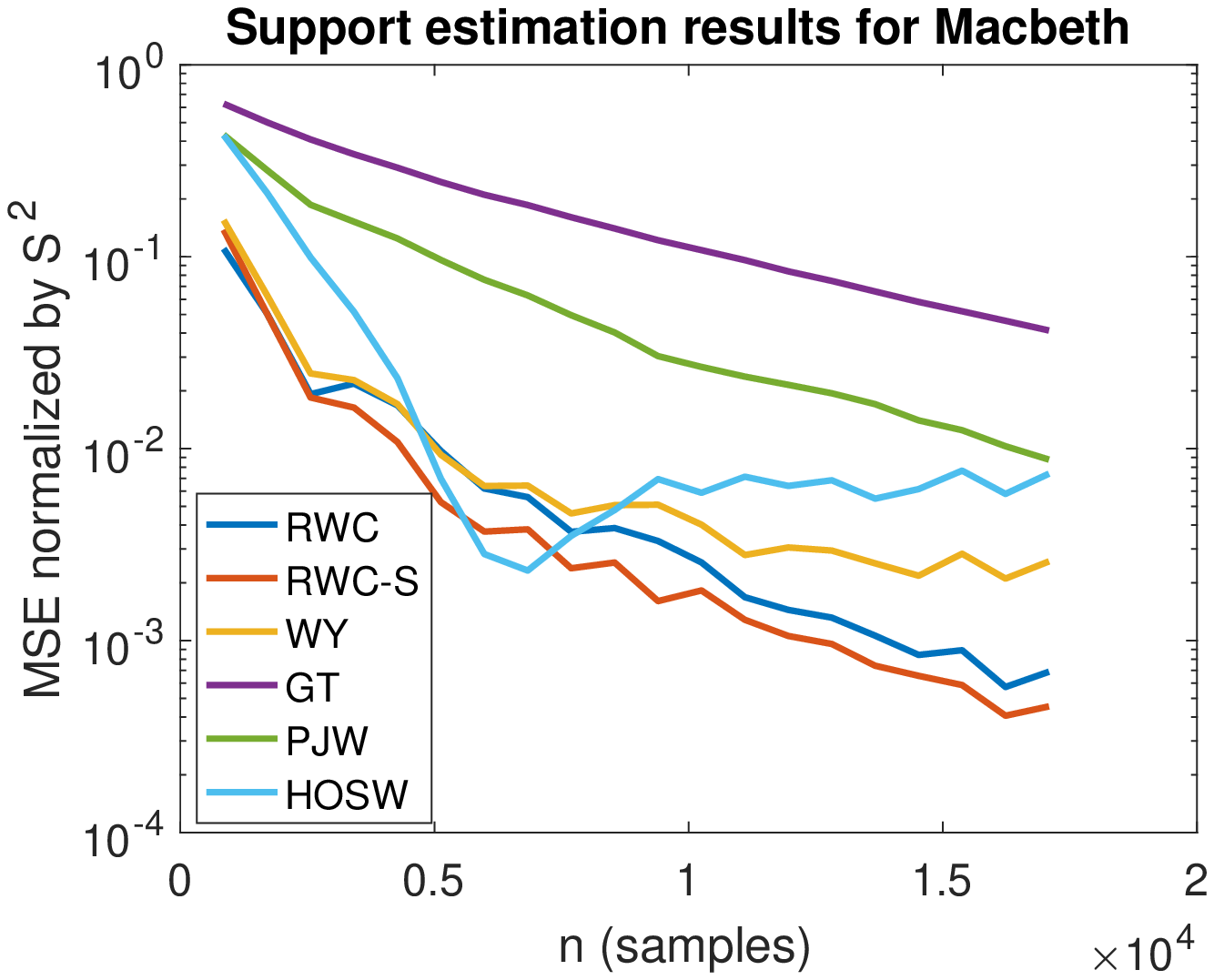}}
  \subfigure[]{\includegraphics[width=0.245\linewidth]{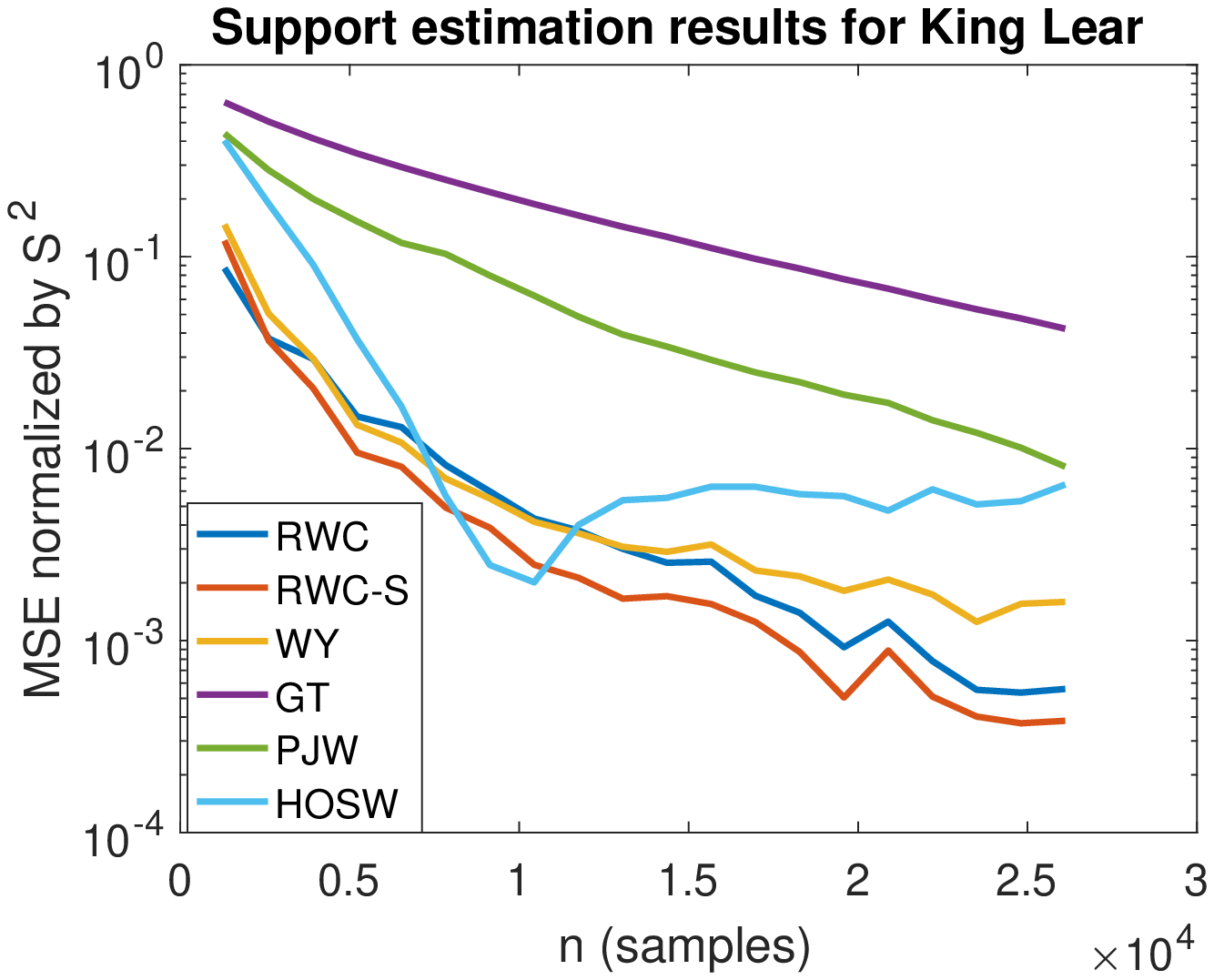}}
  \vspace{-0.2cm}
  \caption{The results obtained by aggregation over $100$ independent trials. The first row of the figures shows the mean and standard deviation of the estimators, while the second row of the figures shows the MSE normalized by $S^2$.}
\end{figure}

% ===============
\clearpage
% ===============

\begin{table}[ht!]
\setlength{\tabcolsep}{2pt}
\centering
\tiny
\begin{tabular}{@{}c|cccccc|c@{}}
\toprule
Genomic region & RWC-S & Naive & WY   & GT   & PJW  & HOSW & Maximum support \\ \midrule
ORF1a & 1746  & 911   & 2022 & 1082 & 1010 & 2770 & 13203           \\
ORF1b & 858   & 477   & 1046 & 549  & 598  & 1346 & 8087            \\
S     & 438   & 246   & 563  & 277  & 309  & 700  & 3822            \\
ORF3a & 166   & 99    & 204  & 108  & 108  & 272  & 828             \\
E     & 26    & 15    & 25   & 43   & 16   & 30   & 228             \\
M     & 81    & 51    & 85   & 63   & 64   & 103  & 669             \\
ORF6  & 90    & 52    & 105  & 158  & 61   & 118  & 186             \\
ORF7a & 116   & 66    & 131  & 157  & 76   & 156  & 366             \\
ORF8  & 49    & 32    & 53   & 42   & 43   & 60   & 366             \\
N     & 221   & 139   & 274  & 146  & 175  & 336  & 1260            \\
ORF10 & 39    & 30    & 21   & 35   & 40   & 21   & 117             \\
All   & 3830  & 2118  & 4529 & 2660 & 2500 & 5912 & 29132           \\ \bottomrule
\end{tabular}
\vspace{0.1in}
\caption{Results of the noiseless support estimation method on samples from Europe.}\label{table:noiseless}
\end{table}

\begin{table}[ht!]
\setlength{\tabcolsep}{2pt}
\centering
\tiny
\begin{tabular}{@{}c|cccc|cccc|cccc@{}}
\toprule
Genomic region      & \multicolumn{4}{|c}{$\eta = 0.5$}                     & \multicolumn{4}{|c|}{$\eta = 1$}               & \multicolumn{4}{c}{$\eta = 1.5$}                     \\ \midrule
      & RWC-S-prc      & Naïve        & WY-Naïve & WY-prc & RWC-S-prc      & Naïve & WY-Naïve & WY-prc & RWC-S-prc      & Naïve & WY-Naïve & WY-prc        \\
ORF1a & \textbf{-30.5} & -50.6 & -116.5 & -64.7 & \textbf{-21.3} & -28.9 & -54.7 & -38   & \textbf{-15.9} & -17.2 & -25.7 & -31.2 \\
ORF1b & \textbf{-28.4} & -50.9        & -120     & -63.6  & \textbf{-17.9} & -28.9 & -54.9    & -37.5  & \textbf{-12.2} & -17   & -22.9    & -17           \\
S     & \textbf{-30.8} & -51.6        & -122.9   & -64.7  & \textbf{-18.5} & -30.1 & -57.7    & -41    & \textbf{-12.6} & -17.5 & -25      & -16.2         \\
ORF3a & \textbf{-32.5} & -49.5 & -96.1  & -70.6 & \textbf{-18.1} & -27.3 & -51.5 & -39.2 & \textbf{-11.4} & -16.2 & -28.4 & -36.8 \\
E     & -76.9          & \textbf{-60} & -76      & -76    & \textbf{-26.9} & -33.3 & -44      & -48    & -15.4          & -20   & -28      & \textbf{-12}  \\
M     & \textbf{-25.9} & -49          & -82.4    & -61.2  & \textbf{-19.8} & -29.4 & -42.4    & -35.3  & -11.1          & -15.7 & -21.2    & \textbf{3.5}  \\
ORF6  & \textbf{-31.1} & -55.8        & -93.3    & -68.6  & \textbf{-22.2} & -30.8 & -50.5    & -41    & \textbf{-16.7} & -19.2 & -29.5    & -25.7         \\
ORF7a & \textbf{-33.6} & -54.5 & -90.1  & -68.7 & \textbf{-24.1} & -31.8 & -50.4 & -42.7 & \textbf{-17.2} & -18.2 & -29   & -24.4 \\
ORF8  & \textbf{-22.4} & -46.9        & -83      & -58.5  & \textbf{-16.3} & -28.1 & -39.6    & -32.1  & -10.2          & -15.6 & -18.9    & \textbf{-9.4} \\
N     & \textbf{-24}   & -45.3        & -121.9   & -56.6  & \textbf{-14.5} & -25.9 & -49.6    & -33.2  & \textbf{-10}   & -15.1 & -16.4    & -34.3         \\
ORF10 & \textbf{7.7}   & -40          & -61.9    & 14.3   & \textbf{7.7}   & -20   & 42.9     & 66.7   & \textbf{7.7}   & -10   & 85.7     & 71.4          \\
All   & \textbf{-29.6} & -50.5        & -114.7   & -64    & \textbf{-19.5} & -28.8 & -53.5    & -37.7  & \textbf{-13.8} & -16.9 & -24.1    & -24.7         \\ \bottomrule
\end{tabular}
\vspace{0.1in}
\caption{Support estimation with synthetic Poisson repeats on samples from Europe. We report the relative difference (in $\%$) of the results for the noisy and noiseless counterparts (the closer the value to $0,$ the better the performance).}
\end{table}

\begin{table}[ht!]
\setlength{\tabcolsep}{2pt}
\centering
\tiny
\begin{tabular}{@{}c|cccc|cccc|cccc|c@{}}
\toprule
Genomic region      & \multicolumn{4}{|c|}{$\eta = 0.5$}          & \multicolumn{4}{c}{$\eta = 1$}           & \multicolumn{4}{|c|}{$\eta = 1.5$}              & Maximum support \\ \midrule
      & RWC-S-prc & Naive & WY-Naive   & WY-prc & RWC-S-prc & Naive & WY-Naive & WY-prc & RWC-S-prc & Naive & WY-Naive & WY-prc       &                 \\
ORF1a & 2346      & 911   & \underline{465}  & 1262   & 3173      & 911   & 2022     & 3277   & 4030      & 911   & 3273     & \underline{105101} & 13203           \\
ORF1b & 1106      & 477   & \underline{230}  & 665    & 1465      & 477   & 1046     & 1718   & 1908      & 477   & 1649     & \underline{67043}  & 8087            \\
S     & 553 & 246 & \underline{104} & 353 & 720 & 246 & 563      & 967      & 949 & 246 & 888 & \underline{44332} & 3822 \\
ORF3a & 210 & 99  & \underline{70}  & 112 & 265 & 99  & 204      & 273      & 325 & 99  & 299 & \underline{5467}  & 828  \\
E     & 32  & 15  & 15        & 15  & 41  & 15  & 25       & 23       & 55  & 15  & 32  & 84          & 228  \\
M     & 94  & 51  & \underline{45}  & 59  & 114 & 51  & 85       & 96       & 135 & 51  & 133 & \underline{685}   & 669  \\
ORF6  & 106 & 52  & \underline{42}  & 64  & 132 & 52  & 105      & 124      & 153 & 52  & 153 & \underline{543}   & 186  \\
ORF7a & 139 & 66  & \underline{55}  & 80  & 173 & 66  & 131      & 151      & 202 & 66  & 187 & \underline{632}   & 366  \\
ORF8  & 56  & 32  & \underline{28}  & 37  & 67  & 32  & 53       & 61       & 79  & 32  & 82  & 196         & 366  \\
N     & 267 & 139 & \underline{74}  & 188 & 337 & 139 & 274      & 450      & 429 & 139 & 508 & \underline{16493} & 1260 \\
ORF10 & 46  & 30  & \underline{28}  & 32  & 54  & 30  & \underline{21} & \underline{29} & 69  & 30  & 91  & \underline{-225}  & 117  \\
All   & 4955      & 2118  & \underline{1156} & 2867   & 6541      & 2118  & 4529     & 7169   & 8334      & 2118  & 7295     & \underline{240351} & 29132           \\ \bottomrule
\end{tabular}
\vspace{0.1in}
\caption{Results of the noisy support estimation method directly applied to the real-world data on samples from Europe. We underline the results that are obviously false (i.e., those violating the maximum support constraint, taking negative values or values smaller than the results of the naive estimator).}
\end{table}

% \usepackage{booktabs}
\begin{table}[ht]
\tiny
\centering
\begin{tabular}{@{}cccccccc@{}}
\toprule
Genome region & RWC-S & Naive & WY   & GT   & PJW  & HOSW & Maximum support \\ \midrule
ORF1a & 1752  & 835   & 2139 & 1543 & 897  & 2752 & 13203           \\
ORF1b & 624   & 316   & 762  & 747  & 335  & 983  & 8087            \\
S     & 353   & 188   & 455  & 375  & 213  & 551  & 3822            \\
ORF3a & 171   & 93    & 185  & 186  & 107  & 242  & 828             \\
E     & 63    & 36    & 82   & 185  & 38   & 84   & 228             \\
M     & 51    & 31    & 65   & 55   & 31   & 72   & 669             \\
ORF6  & 3     & 3     & 3    & Inf  & 3    & 5    & 186             \\
ORF7a & 214   & 109   & 234  & 633  & 117  & 301  & 366             \\
ORF8  & 339   & 340   & 330  & 340  & 340  & 316  & 366             \\
N     & 93    & 60    & 115  & 70   & 74   & 132  & 1260            \\
ORF10 & 17    & 11    & 17   & 15   & 13   & 20   & 117             \\
All   & 3680  & 2022  & 4387 & 4149 & 2168 & 5458 & 29132           \\ \bottomrule
\end{tabular}
\vspace{0.1in}
\caption{Results of the noiseless support estimation method on samples from Asia.}
\end{table}

% \usepackage{booktabs}
\begin{table}[ht]
\tiny
\centering
\setlength{\tabcolsep}{2pt}
\begin{tabular}{@{}c|cccc|cccc|cccc@{}}
\toprule
Genome region &
  \multicolumn{4}{c|}{$\eta$ = 0.5} &
  \multicolumn{4}{c|}{$\eta$ = 1} &
  \multicolumn{4}{c}{$\eta$ = 1.5} \\ \midrule
 &
  RWC-S-prc &
  Naive &
  WY-Naive &
  WY-prc &
  RWC-S-prc &
  Naive &
  WY-Naive &
  WY-prc &
  RWC-S-prc &
  Naive &
  WY-Naive &
  WY-prc \\
ORF1a &
  \textbf{-35.3} &
  -55.7 &
  -122.7 &
  -69.6 &
  \textbf{-26.1} &
  -32.8 &
  -61.7 &
  -44.1 &
  \textbf{-18.8} &
  -19.8 &
  -32.3 &
  -19.9 \\
ORF1b &
  \textbf{-41.8} &
  -56.3 &
  -104.9 &
  -76.8 &
  \textbf{-22.9} &
  -33.9 &
  -59.7 &
  -47 &
  \textbf{-15.4} &
  -19.9 &
  -34.5 &
  -29.8 \\
S &
  \textbf{-39.1} &
  -55.9 &
  -105.9 &
  -76.5 &
  \textbf{-21.5} &
  -33 &
  -59.3 &
  -45.3 &
  \textbf{-14.7} &
  -19.7 &
  -34.9 &
  -33.8 \\
ORF3a &
  \textbf{-36.8} &
  -55.9 &
  -89.2 &
  -70.3 &
  \textbf{-26.3} &
  -32.3 &
  -50.8 &
  -44.3 &
  -16.4 &
  -19.4 &
  -30.8 &
  \textbf{-3.2} \\
E &
  \textbf{-38.1} &
  -61.1 &
  -98.8 &
  -73.2 &
  \textbf{-25.4} &
  -36.1 &
  -56.1 &
  -46.3 &
  \textbf{-19} &
  -19.4 &
  -35.4 &
  -28 \\
M &
  \textbf{-31.4} &
  -54.8 &
  -92.3 &
  -69.2 &
  \textbf{-23.5} &
  -32.3 &
  -53.8 &
  -46.2 &
  \textbf{-15.7} &
  -19.4 &
  -32.3 &
  -21.5 \\
ORF6 &
  \textbf{-66.7} &
  \textbf{-66.7} &
  \textbf{-66.7} &
  \textbf{-66.7} &
  \textbf{-33.3} &
  \textbf{-33.3} &
  \textbf{-33.3} &
  \textbf{-33.3} &
  \textbf{-33.3} &
  \textbf{-33.3} &
  \textbf{-33.3} &
  \textbf{-33.3} \\
ORF7a &
  \textbf{-42.1} &
  -57.8 &
  -91.5 &
  -73.1 &
  \textbf{-30.8} &
  -35.8 &
  -55.1 &
  -47.9 &
  \textbf{-19.2} &
  -21.1 &
  -33.8 &
  -23.5 \\
ORF8 &
  -5 &
  \textbf{-0.6} &
  -2.4 &
  3 &
  0.6 &
  \textbf{0} &
  2.7 &
  3 &
  0.3 &
  \textbf{0} &
  3.6 &
  -8.5 \\
N &
  \textbf{-24.7} &
  -45 &
  -99.1 &
  -65.2 &
  \textbf{-16.1} &
  -25 &
  -47.8 &
  -33.9 &
  \textbf{-9.7} &
  -15 &
  -22.6 &
  -19.1 \\
ORF10 &
  -70.6 &
  \textbf{-54.5} &
  -70.6 &
  -70.6 &
  \textbf{-17.6} &
  -27.3 &
  -35.3 &
  -41.2 &
  \textbf{-11.8} &
  -18.2 &
  -23.5 &
  \textbf{-11.8} \\
All &
  \textbf{-34.4} &
  -46.4 &
  -104 &
  -66.2 &
  \textbf{-22.6} &
  -27.4 &
  -54.8 &
  -41.2 &
  \textbf{-15.7} &
  -16.4 &
  -30 &
  -21.8 \\ \bottomrule
\end{tabular}
\vspace{0.1in}
\caption{Tests of the noisy support estimation methods against synthetic Poisson repeats introduced into samples from Asia. We report the relative difference (in $\%$) compared to the results of the noiseless counterparts (the closer the value to $0,$ the better the estimator).}
\end{table}

% \usepackage{booktabs}
\begin{table}[ht]
\tiny
\centering
\setlength{\tabcolsep}{2pt}
\begin{tabular}{@{}c|cccc|cccc|cccc|c@{}}
\toprule
Genome region & \multicolumn{4}{c|}{$\eta$ = 0.5}        & \multicolumn{4}{c|}{$\eta$ = 1}          & \multicolumn{4}{c|}{$\eta$ = 1.5}        & Maximum support \\ \midrule
              & RWC-S-prc & Naive & WY-Naive & WY-prc & RWC-S-prc & Naive & WY-Naive & WY-prc & RWC-S-prc & Naive & WY-Naive & WY-prc &                 \\
ORF1a & 2401      & 835  & {\ul 289}  & 1254 & 3278 & 835  & 2139      & 3864      & 4138 & 835  & 3508 & {\ul 183124} & 13203 \\
ORF1b & 832       & 316  & {\ul 192}  & 372  & 1144 & 316  & 762       & 1072      & 1578 & 316  & 1060 & {\ul 29256}  & 8087  \\
S     & 465       & 188  & {\ul 110}  & 223  & 621  & 188  & 455       & 654       & 809  & 188  & 627  & {\ul 4799}   & 3822  \\
ORF3a & 210       & 93   & {\ul 79}   & 112  & 266  & 93   & 185       & 207       & 314  & 93   & 257  & {\ul 2764}   & 828   \\
E     & 74        & 36   & {\ul 28}   & 47   & 90   & 36   & 82        & 100       & 105  & 36   & 114  & {\ul 560}    & 228   \\
M     & 58        & 31   & {\ul 25}   & 38   & 70   & 31   & 65        & 76        & 81   & 31   & 79   & 401          & 669   \\
ORF6  & 3         & 3    & 3          & 3    & 3    & 3    & 3         & 3         & 3    & 3    & 3    & 3            & 186   \\
ORF7a & 269       & 109  & {\ul 90}   & 135  & 346  & 109  & 234       & 265       & 409  & 109  & 314  & {\ul 4372}   & 366   \\
ORF8  & {\ul 339} & 340  & 340        & 340  & 341  & 340  & {\ul 330} & {\ul 324} & 346  & 340  & 352  & {\ul -11001} & 366   \\
N     & 113       & 60   & {\ul 41}   & 68   & 137  & 60   & 115       & 162       & 163  & 60   & 170  & 1015         & 1260  \\
ORF10 & 21        & 11   & 11         & 11   & 26   & 11   & 17        & 16        & 33   & 11   & 22   & 44           & 117   \\
All   & 4785      & 2022 & {\ul 1208} & 2603 & 6322 & 2022 & 4387      & 6743      & 7979 & 2022 & 6506 & {\ul 215337} & 29132 \\ \bottomrule
\end{tabular}
\vspace{0.1in}
\caption{Results of noisy support estimation methods applied directly on samples from Asia. We underline the results that are obviously false (i.e., those that violate the maximum support size constraint, those that produce negative results and estimates smaller than the results of the Naive estimator). }
\end{table}

% \usepackage{booktabs}
\begin{table}[ht]
\tiny
\centering
\begin{tabular}{@{}c|cccccc|c@{}}
\toprule
Genome region & RWC-S & Naive & WY   & GT   & PJW  & HOSW & Maximum support \\ \midrule
ORF1a         & 1509  & 804   & 1765 & 944  & 980  & 2374 & 13203           \\
ORF1b         & 727   & 403   & 912  & 437  & 442  & 1245 & 8087            \\
S             & 375   & 209   & 515  & 237  & 229  & 636  & 3822            \\
ORF3a         & 134   & 81    & 161  & 86   & 97   & 214  & 828             \\
E             & 25    & 15    & 24   & 33   & 15   & 29   & 228             \\
M             & 44    & 28    & 52   & 38   & 32   & 59   & 669             \\
ORF6          & 33    & 21    & 43   & 32   & 23   & 46   & 186             \\
ORF7a         & 269   & 135   & 285  & 587  & 142  & 384  & 366             \\
ORF8          & 44    & 29    & 45   & 30   & 36   & 51   & 366             \\
N             & 227   & 138   & 272  & 174  & 185  & 331  & 1260            \\
ORF10         & 16    & 10    & 16   & 20   & 14   & 16   & 117             \\
All           & 3403  & 1873  & 4090 & 2618 & 2195 & 5385 & 29132           \\ \bottomrule
\end{tabular}
\vspace{0.1in}
\caption{Results for the noiseless support estimation methods applied to samples from North America.}
\end{table}

% \usepackage{booktabs}
\begin{table}[ht]
\tiny
\centering
\setlength{\tabcolsep}{2pt}
\begin{tabular}{@{}c|cccc|cccc|cccc@{}}
\toprule
Genome region & \multicolumn{4}{c|}{$\eta$ = 0.5}                      & \multicolumn{4}{c|}{$\eta$ = 1}               & \multicolumn{4}{c}{$\eta$ = 1.5}                           \\ \midrule
              & RWC-S-prc      & Naive          & WY-Naive & WY-prc & RWC-S-prc      & Naive & WY-Naive & WY-prc & RWC-S-prc      & Naive & WY-Naive      & WY-prc         \\
ORF1a         & \textbf{-28.4} & -51            & -117.5   & -64.1  & \textbf{-20.1} & -29.2 & -54.7    & -38    & \textbf{-14.4} & -17.2 & -25           & -15.2          \\
ORF1b         & \textbf{-30.3} & -49.6          & -115.5   & -65    & \textbf{-19.5} & -28   & -56.1    & -40.2  & \textbf{-13.9} & -16.4 & -26.8         & -57            \\
S             & \textbf{-33.9} & -52.6          & -122.3   & -67.6  & \textbf{-18.4} & -30.1 & -60      & -42.3  & \textbf{-12.8} & -17.7 & -28.9         & -19.2          \\
ORF3a & \textbf{-30.6} & -50.6 & -96.9  & -69.6 & \textbf{-17.2} & -27.2 & -49.7 & -37.9 & \textbf{-11.2} & -16   & -25.5 & -30.4 \\
E             & -72            & \textbf{-53.3} & -70.8    & -70.8  & \textbf{-24}   & -33.3 & -41.7    & -45.8  & \textbf{-16}   & -20   & -25           & -20.8          \\
M             & \textbf{-27.3} & -50            & -86.5    & -63.5  & \textbf{-15.9} & -25   & -46.2    & -36.5  & \textbf{-11.4} & -17.9 & -26.9         & -15.4          \\
ORF6          & \textbf{-33.3} & -52.4          & -93      & -69.8  & \textbf{-18.2} & -33.3 & -51.2    & -39.5  & \textbf{-9.1}  & -19   & -30.2         & -20.9          \\
ORF7a         & \textbf{-43.1} & -57.8          & -90.2    & -73    & \textbf{-31.2} & -34.8 & -54.4    & -48.4  & -20.1          & -20   & -34           & \textbf{-18.2} \\
ORF8          & \textbf{-22.7} & -44.8          & -93.3    & -57.8  & \textbf{-15.9} & -24.1 & -35.6    & -22.2  & -9.1           & -13.8 & \textbf{-6.7} & 8.9            \\
N     & \textbf{-27.3} & -49.3 & -101.1 & -68.4 & \textbf{-15.4} & -27.5 & -50.4 & -37.1 & \textbf{-7.9}  & -15.9 & -22.8 & -16.9 \\
ORF10         & -68.8          & \textbf{-50}   & -68.8    & -68.8  & \textbf{-25}   & -30   & -37.5    & -37.5  & \textbf{-12.5} & -20   & -18.8         & -18.8          \\
All   & \textbf{-31.1} & -51.1 & -112.4 & -65.9 & \textbf{-20.2} & -29.2 & -54.7 & -39.6 & \textbf{-13.8} & -17.1 & -26.3 & -25.8 \\ \bottomrule
\end{tabular}
\vspace{0.1in}
\caption{Tests of the noisy support estimation methods against synthetic Poisson repeats introduced into samples from North America. We report the relative difference (in $\%$) compared to the results of the noiseless counterparts (the closer the value to $0,$ the better the estimator).}
\end{table}

% \usepackage{booktabs}
% \usepackage[normalem]{ulem}
\begin{table}[ht]
\tiny
\centering
\setlength{\tabcolsep}{2pt}
\begin{tabular}{@{}c|cccc|cccc|cccc|c@{}}
\toprule
Genome region & \multicolumn{4}{c|}{$\eta$ = 0.5}          & \multicolumn{4}{c|}{$\eta$ = 1}          & \multicolumn{4}{c|}{$\eta$ = 1.5}              & Maximum support \\ \midrule
              & RWC-S-prc & Naïve & WY-Naïve   & WY-prc & RWC-S-prc & Naïve & WY-Naïve & WY-prc & RWC-S-prc & Naïve & WY-Naïve & WY-prc       &                 \\
ORF1a         & 2005      & 804   & {\ul 407}  & 1114   & 2716      & 804   & 1765     & 2851   & 3502      & 804   & 2802     & {\ul 95895}  & 13203           \\
ORF1b & 924 & 403 & {\ul 206} & 558 & 1202      & 403 & 912 & 1474 & 1535      & 403 & 1391 & {\ul 55948} & 8087 \\
S     & 471 & 209 & {\ul 82}  & 305 & 608       & 209 & 515 & 901  & 799       & 209 & 793  & {\ul 50434} & 3822 \\
ORF3a & 167 & 81  & {\ul 58}  & 91  & 209       & 81  & 161 & 213  & 253       & 81  & 240  & {\ul 3656}  & 828  \\
E     & 31  & 15  & 15        & 15  & 39        & 15  & 24  & 22   & 50        & 15  & 30   & 72          & 228  \\
M     & 51  & 28  & {\ul 24}  & 34  & 60        & 28  & 52  & 60   & 71        & 28  & 77   & 235         & 669  \\
ORF6  & 39  & 21  & {\ul 17}  & 26  & 46        & 21  & 43  & 52   & 54        & 21  & 62   & {\ul 252}   & 186  \\
ORF7a & 345 & 135 & {\ul 113} & 165 & {\ul 448} & 135 & 285 & 317  & {\ul 535} & 135 & 381  & {\ul4729}   & 366  \\
ORF8  & 52  & 29  & {\ul 22}  & 32  & 61        & 29  & 45  & 61   & 70        & 29  & 93   & 168         & 366  \\
N     & 287 & 138 & {\ul 95}  & 157 & 369       & 138 & 272 & 372  & 469       & 138 & 408  & {\ul 8471}  & 1260 \\
ORF10 & 18  & 10  & 10        & 10  & 23        & 10  & 16  & 15   & 28        & 10  & 22   & 47          & 117  \\
All           & 4390      & 1873  & {\ul 1049} & 2507   & 5781      & 1873  & 4090     & 6338   & 7366      & 1873  & 6299     & {\ul 219907} & 29132           \\ \bottomrule
\end{tabular}
\caption{Results of noisy support estimation methods applying directly to North America samples. We underline the results that are obviously false (i.e. violating maximum support, being negative and smaller than the results of Naive estimator).  }
\end{table}

\end{document}